\documentclass{article}

\usepackage[preprint,nonatbib]{nips_2018}

\usepackage[utf8]{inputenc} 
\usepackage{hyperref}       
\usepackage{url}            
\usepackage{booktabs}       
\usepackage{amsfonts}       
\usepackage{nicefrac}       
\usepackage{microtype}      

\usepackage{textcomp}
\usepackage{xspace}
\usepackage{amsmath}
\usepackage[linesnumbered, inoutnumbered]{algorithm2e}
\usepackage{xcolor}
\usepackage{dsfont}
\usepackage{amssymb}
\usepackage{graphicx}
\usepackage{subfig}
\usepackage{float}
\usepackage{enumerate}

\newcommand{\RR}{\mathbb{R}} 
\newcommand{\half}[0]{\frac{1}{2}}

\newcommand{\expectv}{\mathbb{E}}

\newcommand\norm[1]{\left\lVert#1\right\rVert}
\newcommand{\dtX}{\pmb{\mathcal{X}}}
\newcommand{\dtY}{\pmb{\mathcal{Y}}}

\newcommand{\topp}[0]{\emph{top}\xspace}
\newcommand{\btm}[0]{\emph{bottom}\xspace}
\newcommand{\supp}[0]{\emph{supervised}\xspace}
\newcommand{\unsupp}[0]{\emph{unsupervised}\xspace}

\title{Neural Spectrum Alignment: Empirical Study}

\author{
  Dmitry Kopitkov\thanks{Dmitry Kopitkov is with the Technion Autonomous Systems Program (TASP), {\tt dimkak@technion.ac.il}.} 
  \And \\
  Technion - Israel Institute of Technology\\
  Haifa 32000, Israel
  \And 
  Vadim Indelman\thanks{Vadim Indelman is with the Department of Aerospace Engineering, {\tt vadim.indelman@technion.ac.il}.}
}

\begin{document}

\maketitle

\begin{abstract}
	
Expressiveness and generalization of deep models was recently addressed via the connection between neural networks (NNs) and kernel learning, where first-order dynamics of NN during a gradient-descent (GD) optimization were related to \emph{gradient similarity} kernel, also known as Neural Tangent Kernel (NTK) \cite{Jacot18nips}. In the majority of works this kernel is considered to be time-invariant \cite{Jacot18nips,Lee19arxiv}, with its properties being defined entirely by NN architecture and independent of the learning task at hand. In contrast, in this paper we empirically explore these properties along the optimization and show that in practical applications the NTK changes in a very dramatic and meaningful way, with its \topp eigenfunctions aligning toward the target function learned by NN. 
Moreover, these \topp eigenfunctions serve as basis functions for NN output - a function represented by NN is spanned almost completely by them for the entire optimization process. 
Further, since the learning along \topp eigenfunctions is typically fast, their alignment with the target function improves the overall optimization performance.
In addition, we study how the \emph{neural spectrum} is affected by learning rate decay, typically done by practitioners, showing various trends in the kernel behavior. We argue that the presented phenomena may lead to a more complete theoretical understanding behind NN learning.

\end{abstract}

\section{Introduction}
\label{sec:Intro}

Understanding expressiveness and generalization of deep models is essential for  robust performance of NNs. Recently, the optimization analysis for a general NN architecture was related to \emph{gradient similarity} kernel \cite{Jacot18nips}, whose properties govern NN expressivity level, generalization and convergence rate. Under various considered conditions \cite{Jacot18nips,Lee19arxiv}, this NN kernel converges to its steady state and is invariant along the entire optimization, which significantly facilitates the analyses of Deep Learning (DL) theory \cite{Jacot18nips,Lee19arxiv, Basri19arxiv,Bietti19arxiv,Hayou19arxiv,Arora19arxiv}.


Yet, in a typical realistic setting the \emph{gradient similarity} kernel is far from being constant, as we empirically demonstrate in this paper. Moreover, its spectrum undergoes a very specific change during  training, aligning itself towards the target function that is learned by NN. This kernel adaptation in its turn improves the optimization convergence rate, by decreasing a norm of the target function within the corresponding reproducing kernel Hilbert space (RKHS) \cite{Arora19arxiv}. Furthermore, these \emph{gradient similarity} dynamics can also explain the expressive superiority of deep NNs over more shallow models.
Hence, we argue that understanding the \emph{gradient similarity} of NNs beyond its time-invariant regime is a must for full comprehension of NN expressiveness power.

To encourage the onward theoretical research of the kernel, herein we report several strong empirical phenomena and trends of its dynamics. To the best of our knowledge, these trends neither were yet reported nor they can be explained by DL theory developed so far. We argue that accounting for the presented below phenomena can lead to a more complete learning theory of complex hierarchical models like modern NNs.

To this end, in this paper we perform an empirical investigation of fully-connected (FC) NN, its \emph{gradient similarity} kernel and the corresponding Gramian at training data points during the entire period of a typical learning process. Our main empirical contributions are:
\begin{enumerate}[(a)]
	\item We show that Gramian serves as a NN memory, with its \topp eigenvectors changing to align with the learned target function. This improves the optimization performance since the convergence rate along kernel \topp eigenvectors is typically higher.
	
	\item During the entire optimization NN output is located inside a sub-space spanned by these \topp eigenvectors, making the eigenvectors to be a basis functions of NN.

	\item Deeper NNs demonstrate a stronger alignment, which may explain their expressive superiority. In contrast, shallow wide NNs with a similar number of parameters achieve a significantly lower alignment level and a worse optimization performance.
	
	\item We show additional trends in kernel dynamics as a consequence of learning rate decay.	Specifically, after each decay the information about the target function, that is gathered inside \topp eigenvectors, is spread to a bigger number of \topp eigenvectors. Likewise, kernel eigenvalues grow after each learning rate drop, and an eigenvalue-learning-rate product is kept around the same value for the entire optimization.

	\item Experiments over various FC architectures and real-world datasets are performed. Likewise, several \supp and \unsupp learning algorithms and number of popular optimizers were evaluated. All experiments showed the mentioned above spectrum alignment.

\end{enumerate}

The paper is structured as follows. In Section \ref{sec:Nottn} we define necessary notations for first-order NN dynamics. In Section \ref{sec:FIMSec} we relate \emph{gradient similarity} with Fisher information matrix (FIM) of NN and in Section \ref{sec:L2Loss} we provide more insight about NN dynamics on L2 loss example. In Section \ref{sec:RelWork} the related work is described and in Section \ref{sec:Expr} we present our main empirical study. Later, conclusions are discussed in Section \ref{sec:Concl}. Further, additional derivations 
placed in the Appendix. Finally, more visual illustrations of NN spectrum and additional experiments are placed in the separate Supplementary Material (SM) \cite{Kopitkov19dm_Supplementary} due to the large size of involved graphics.

\section{Notations}
\label{sec:Nottn}

Consider a NN $f_{\theta}(X): \RR^d \rightarrow \RR$ with a parameter vector $\theta$, a typical sample loss $\ell$ and an empirical loss $L$, training samples $D = \left[ \dtX = \{ X^i \in \RR^d \}_{i = 1}^{N}, \dtY = \{ Y^i \in \RR \}_{i = 1}^{N} \right]$ and loss gradient $\nabla_{\theta} L$:
\begin{equation}
L(\theta, D)
=
\frac{1}{N}
\sum_{i = 1}^{N}
\ell
\left[
X^i,
Y^i,
f_{\theta}(X^i)
\right]
,
\quad
\nabla_{\theta} L(\theta, D)
=
\frac{1}{N}
\sum_{i = 1}^{N}
\ell'
\left[
X^i,
Y^i,
f_{\theta}(X^i)
\right]
\cdot
\nabla_{\theta} 
f_{\theta}(X^i)
.
\label{eq:GeneralLoss}
\end{equation}
%
The above formulation can be extended to include \unsupp learning methods in \cite{Kopitkov19arxiv} by eliminating labels $\dtY$ from the equations. Further, techniques with a model $f_{\theta}(X)$ returning multidimensional outputs are out of scope for this paper, to simplify the formulation.

Consider a GD optimization with learning rate $\delta$, where parameters change at each discrete optimization time $t$ as $d\theta_{t} \triangleq \theta_{t + 1} - \theta_{t} = - \delta \cdot \nabla_{\theta} L(\theta_{t}, D)$.
%
Further, a model output change at any $X$ according to first-order Taylor approximation is:
\begin{multline}
d f_{\theta_{t}}(X) 
\triangleq
f_{\theta_{t + 1}}(X)
-
f_{\theta_{t}}(X)
=
d\theta_{t}^T
\cdot
\int_{0}^{1}
\nabla_{\theta} 
f_{\theta_{s}}(X)
ds
\approx\\
\approx
\nabla_{\theta} 
f_{\theta_{t}}(X)^T \cdot d\theta_{t}
=
-
\frac{\delta}{N}
\sum_{i = 1}^{N}
g_{t}(X, X^i)
\cdot
\ell'
\left[
X^i,
Y^i,
f_{\theta_{t}}(X^i)
\right]
,
\label{eq:FDff}
\end{multline}
where $\theta_s \triangleq (1-s)\theta_{t} + s\theta_{t+1}$ and $\int_{0}^{1}
\nabla_{\theta} 
f_{\theta_{s}}(X)
ds$ is a gradient averaged over the straight line between $\theta_{t}$ and $\theta_{t + 1}$. Further,
$g_{t}(X, X') \triangleq \nabla_{\theta} 
f_{\theta_t}(X)^T \cdot \nabla_{\theta} 
f_{\theta_t}(X')$ is a \emph{gradient similarity} - the dot-product of gradients at two different input points also known as NTK \cite{Jacot18nips}, and where $\ell'
\left[
X^i,
Y^i,
f_{\theta_{t}}(X^i)
\right]
\triangleq
\nabla_{f_{\theta}} \ell
\left[
X^i,
Y^i,
f_{\theta_{t}}(X^i)
\right]
$.

In this paper we mainly focus on optimization dynamics of $f_{\theta}$ at training points. To this end, define a vector $\bar{f}_{t} \in \RR^{N}$ with $i$-th entry being $f_{\theta_{t}}(X^i)$. According to  Eq.~(\ref{eq:FDff}) the discrete-time evolution of $f_{\theta}$ at testing and training points follows:
\begin{equation}
d f_{\theta_{t}}(X) 
\approx
-
\frac{\delta}{N}
\cdot
g_{t}(X, \dtX)
\cdot
\bar{m}_{t}
,
\quad
d \bar{f}_{t}
\triangleq
\bar{f}_{t + 1}
-
\bar{f}_{t}
\approx
-
\frac{\delta}{N}
\cdot
G_{t}
\cdot
\bar{m}_{t}
,
\label{eq:FVecDff}
\end{equation}
where $G_{t} \triangleq g_{t}(\dtX, \dtX)$ is a $N \times N$ Gramian with entries $G_{t}(i,j) = g_{t}(X^i, X^j)$ and $\bar{m}_{t} \in \RR^{N}$ is a vector with the $i$-th entry being 
$\ell'
\left[
X^i,
Y^i,
f_{\theta_{t}}(X^i)
\right]
$
. Likewise, denote eigenvalues of $G_{t}$, sorted in decreasing order, by $\{ \lambda_{i}^t \}_{i = 1}^{N}$, with $\lambda_{max}^{t} \triangleq \lambda_{1}^{t}$ and $\lambda_{min}^{t} \triangleq \lambda_{N}^{t}$. Further, notate the associated orthonormal eigenvectors by $\{ \bar{\upsilon}_{i}^t \}_{i = 1}^{N}$. Note that $\{ \lambda_{i}^t \}_{i = 1}^{N}$ and $\{ \bar{\upsilon}_{i}^t \}_{i = 1}^{N}$ also represent estimations of eigenvalues and eigenfunctions of the kernel $g_{t}(X, X')$ (see Appendix \ref{sec:AppSp} for more details). Below we will refer to large and small eigenvalues and their associated eigenvectors by \topp and \btm terms respectively.

Eq.~(\ref{eq:FVecDff}) describes the first-order dynamics of GD learning, where $\bar{m}_{t}$ is a functional derivative of any considered loss $L$, and the global optimization convergence is typically associated with it becoming a zero vector, due to Euler-Lagrange equation of $L$.
Further, $G_t$ translates a movement in $\theta$-space into a movement in a space of functions defined on $\dtX$.

\section{Relation to Fisher Information Matrix}
\label{sec:FIMSec}

NN Gramian can be written as $G_t = A_t^T A_t$ where $A_t$ is $|\theta| \times N$ Jacobian matrix with $i$-th column being $\nabla_{\theta} f_{\theta_{t}}(X^i)$. Moreover, $F_t = A_t A_t^T$ is known as the empirical FIM of NN\footnote{In some papers \cite{Sagun17arxiv} FIM is also referred to as a Hessian of NN, due to the tight relation between $F_t$ and the Hessian of the loss. See Appendix \ref{sec:AppA} for more details} 
\cite{Park00nn,Ollivier15iai,Karakida18arxiv} that approximates the second moment of model gradients $\frac{1}{N} F_t \approx \expectv_X \left[\nabla_{\theta} f_{\theta_{t}}(X) \nabla_{\theta} f_{\theta_{t}}(X)^T \right]$. Since $F_t$ is dual of $G_t$, both matrices share same non-zero eigenvalues $\{ \lambda_i^t \neq 0 \}$. Furthermore, for each $\lambda_i^t$ the respectful eigenvector $\bar{\omega}_{i}^t$ of $F_t$ is associated with appropriate $\bar{\upsilon}_{i}^t$ - they are left and right singular vectors of $A_t$ respectively. Moreover, change of $\theta_t$ along the direction $\bar{\omega}_{i}^t$ causes a change to $\bar{f}_{t}$ along $\bar{\upsilon}_{i}^t$ (see Appendix \ref{sec:AppB} for the proof). Therefore, spectrums of $G_t$ and $F_t$ describe principal directions in function space and $\theta$-space respectively, according to which $\bar{f}_{t}$ and $\theta_t$ are changing during the optimization. Based on the above, in Section \ref{sec:RelWork} we relate some known properties of $F_t$ towards $G_t$.

\section{Analysis of L2 Loss For Constant Gramian}
\label{sec:L2Loss}

To get more insight into Eq.~(\ref{eq:FVecDff}),
we will consider L2 loss with $\ell
\left[
X^i,
Y^i,
f_{\theta}(X^i)
\right]
=
\half
\left[
f_{\theta}(X^i) - Y^i
\right]^2
$. In such a case we have $\bar{m}_{t} = \bar{f}_{t} - \bar{y}$, with $\bar{y}$ being a vector of labels. Assuming $G_t$ to be fixed along the optimization (see Section \ref{sec:RelWork} for justification), NN dynamics can be written as (see the Appendix \ref{sec:AppC} for a proper derivation):
\begin{equation}
\bar{f}_{t}
=
\bar{f}_{0}
-
\sum_{i = 1}^{N}
\left[
1 -
\left[
1 - 
\frac{\delta}{N}
\lambda_i
\right]^t
\right]
<\bar{\upsilon}_{i}, \bar{m}_{0}>
\bar{\upsilon}_{i}
,
\label{eq:L2DynamicsF}
\end{equation}
\begin{equation}
\bar{m}_{t}
=
\sum_{i = 1}^{N}
\left[
1 - 
\frac{\delta}{N}
\lambda_i
\right]^t
<\bar{\upsilon}_{i}, \bar{m}_{0}>
\bar{\upsilon}_{i}
.
\label{eq:L2DynamicsM}
\end{equation}
Further, dynamics of $f_{\theta_{t}}(X)$ at testing point $X$ appear in the Appendix \ref{sec:AppC_test} since they are not the main focus of this paper.
Under the stability condition $\delta < \frac{2N}{\lambda_{max}}$, the above equations can be viewed as a transmission of a signal from $\bar{m}_{0} = \bar{f}_{0} - \bar{y}$ into our model $\bar{f}_{t}$ - at each iteration $\bar{m}_{t}$ is decreased along each $\{\bar{\upsilon}_{i}: \lambda_i \neq 0\}$ since $\lim\limits_{t \rightarrow \infty} \left[
1 - 
\frac{\delta}{N}
\lambda_i
\right]^t = 0$. Furthermore, the same information decreased from $\bar{m}_{t}$ in Eq.~(\ref{eq:L2DynamicsM}) is appended to $\bar{f}_{t}$ in Eq.~(\ref{eq:L2DynamicsF}).

Hence, in case of L2 loss and for a constant Gramian matrix, conceptually GD transmits information packets from the residual $\bar{m}_{t}$ into our model $\bar{f}_{t}$ along each axis $\bar{\upsilon}_{i}$. Further, $s_i^t \triangleq 1 - |1 - 
\frac{\delta}{N}
\lambda_i|$ governs a speed of information flow along $\bar{\upsilon}_{i}$. Importantly, note that for a high learning rate (i.e. $\delta \approx \frac{2N}{\lambda_{max}}$) the information flow is slow for directions $\bar{\upsilon}_{i}$ with both very large and very small eigenvalues $\lambda_i$, since in former the term $1 - 
\frac{\delta}{N}
\lambda_i$ is close to $-1$ whereas in latter - to $1$. Yet, along with the learning rate decay, performed during a typical optimization, $s_i^t$ for very large $\lambda_i$ is increased. However, the speed along a direction with small $\lambda_i$ is further decreasing with the decay of $\delta$. 
As well, in case $\lambda_{min} > 0$, at the convergence $t \rightarrow \infty$ we will get from Eq.~(\ref{eq:L2DynamicsF})-Eq.~(\ref{eq:L2DynamicsM}) the global minima convergence: $\bar{f}_{\infty}
=
\bar{f}_{0}
- \bar{m}_{0} = \bar{y}$ and $\bar{m}_{\infty} = \bar{0}$.

Under the above setting, there are two important key observations.
First, due to the restriction over $\delta$ in practice the information flow along small $\lambda_i$ can be prohibitively slow in case a conditional number $\frac{\lambda_{max}}{\lambda_{min}}$ is very large. This implies that for a faster convergence it is desirable for NN to have many eigenvalues as close as possible to its $\lambda_{max}$ since this will increase a number of directions in the function space where information flow is fast.
Second, if $\bar{m}_{0}$ (or $\bar{y}$ if $\bar{f}_{0} \approx 0$) is contained entirely within \topp eigenvectors, small eigenvalues will not affect the convergence rate at all. Hence, the higher alignment between $\bar{m}_{0}$ (or $\bar{y}$) and \topp eigenvectors may dramatically improve overall convergence rate. The above conclusions and their extensions towards the testing loss are proved in formal manner in \cite{Arora19arxiv,Oymak19arxiv} for two-layer NNs. Further, the generalization is also shown to be dependent on the above alignment.
In Section \ref{sec:Expr} we support these conclusions experimentally.


\section{Related Work}
\label{sec:RelWork}

First-order NN dynamics can be understood by solving the system in Eq.~(\ref{eq:FVecDff}). However, its solution is highly challenging due to two main reasons - non-linearity of $\bar{m}_{t}$ w.r.t. $\bar{f}_{t}$ (except for the L2 loss) and intricate and yet not fully known time-dependence of Gramian $G_t$.
Although \emph{gradient similarity} $g_{t}(X, X')$ and corresponding $G_t$
achieved a lot of recent attention in DL community \cite{Jacot18nips,Lee19arxiv}, their properties are still investigated mostly only for limits under which $G_t$ becomes time-constant. The first work in this direction was done in \cite{Jacot18nips} where $g_{t}(X, X')$ was proven to converge to Neural Tangent Kernel (NTK) in infinite width limit. Similarly, in \cite{Lee19arxiv} $G_0$ was shown to accurately explain NN dynamics when $\theta_{t}$ is nearby $\theta_{0}$ during the entire optimization. The considered case of constant Gramian facilitates solution of Eq.~(\ref{eq:FVecDff}), as demonstrated in Section \ref{sec:L2Loss}, which otherwise remains intractable. Moreover, GD over NN with constant Gramian/kernel is identical to kernel methods where optimization is solved via kernel gradient descent \cite{Jacot18nips}, and hence theoretical insights from kernel learning can be extrapolated towards NNs.

Yet, in practical-sized NNs the spectrum of $G_t$ is neither constant nor it is similar to its initialization. Recent several studies explored its adaptive dynamics \cite{Woodworth19arxiv,Dou19arxiv,Williams19arxiv}, although most of the work was done for single or two layer NNs. Further, in \cite{Dyer19arxiv,Huang19arxiv} mathematical expressions for NTK dynamics were developed for a general NN architecture. Likewise, in the Appendix \ref{sec:AppD} we derive similar dynamics for the Gramian $G_t$.
Yet, the above derivations produce intricate equations and it is not straightforward to explain the actual behavior of $G_t$ along the optimization, revealed in this paper. Particularly, in Section \ref{sec:Expr} we empirically demonstrate that \topp spectrum of $G_t$ is dramatically affected by the learning task at hand, aligning itself with the target function. To the best of our knowledge, the presented NN kernel trends were not investigated in such detail before.

Further, many works explore properties of FIM $F_t$ both theoretically and empirically \cite{Sagun17arxiv,Gur18arxiv,Karakida18arxiv,Oymak19arxiv}. Specifically, most of these works come to conclusion that in typical NN an absolute majority of FIM eigenvalues are close to zero, with only small part of them being significantly strong. According to Section \ref{sec:FIMSec} the same is also true about eigenvalues of $G_t$.
Furthermore, in \cite{Arora19arxiv,Oymak19arxiv} authors showed for networks with a single hidden layer that NN learnability strongly depends on alignment between labels vector $\bar{y}$ and \topp eigenvectors of $G_t$. Intuitively, it can be explained by fast convergence rate along $\bar{\upsilon}_{i}$ with large $\lambda_i$ vs impractically slow one along directions with small $\lambda_i$, as was shortly described in Section \ref{sec:L2Loss}. Due to most of the eigenvalues being very small, the alignment between $\bar{y}$ and \topp eigenvectors of $G_t$ defines the optimization performance. Moreover, in \cite{Oymak19arxiv} authors also noted the increased aforementioned alignment comparing NN at start and end of the training. This observation was shortly made for ResNet convolutional NN architecture, and in Section \ref{sec:Expr} we empirically investigate this alignment for FC architecture, in comprehensive manner for various training tasks.

Furthermore, the picture of information flow from Section \ref{sec:L2Loss} also explains what target functions are more "easy" to learn. The \topp eigenvectors of $G_t$ typically contain low-frequency signal, which was discussed in \cite{Arora19arxiv} and proved in \cite{Basri19arxiv} for data uniformly distributed on a hypersphere. In its turn, this explains why low-frequency target functions are learned significantly faster as reported in \cite{Zhang16arxiv,Rahaman18arxiv,Arora19arxiv}. Combined with early stopping such behavior is used by DL community as a regularization to prevent fitting high-frequency signal affiliated with noise; this can also be considered as an instance of commonly known Landweber iteration algorithm \cite{Landweber51ajm}.
We support findings of \cite{Basri19arxiv} also in our experiments below, additionally revealing that for a general case the eigenvectors/eigenfunctions of the \emph{gradient similarity} are not spherical harmonics considered in \cite{Basri19arxiv}.

Finally, in context of kernel methods a lot of effort was done to learn the kernels themselves \cite{Varma09aicml,Gonen11jmlr,Wang15air,Wilson16ais}. The standard 2-stage procedure is to first learn the kernel and latter combine it with the original kernel algorithm, where the first stage can involve search for a kernel whose kernel matrix is strongly aligned with the label vector $\bar{y}$ \cite{Gonen11jmlr,Wang15air}, and the second is to solve a data fitting task (e.g. L2 regression problem) over RKHS defined by the new kernel. Such 2-stage adaptive-kernel methods demonstrated an improved accuracy and robustness compared to techniques with pre-defined kernel \cite{Varma09aicml,Gonen11jmlr,Wilson16ais}. In our experiments we show that NNs exhibit a similar alignment of $g_{t}(X, X')$ during the optimization, and hence can be viewed as an adaptive-kernel method where both kernel learning and data fitting proceed in parallel.

\section{Experiments}
\label{sec:Expr}

In this section we empirically study Gramian dynamics along the optimization process. Our main goal here is to illustrate the alignment nature of the \emph{gradient similarity} kernel and verify various deductions made in Section \ref{sec:L2Loss} under a constant-Gramian setting for a real learning case. To do so in detailed and intuitive manner, we focus our experiments on 2D dataset where visualization of kernel eigenfunctions is possible. 
We perform a simple regression optimization of FC network via GD, where a learning setup\footnote{Related code can be accessed via a repository \url{https://bit.ly/2kGVHhG}} is similar to common conventions applied by DL practitioners. \textbf{All} empirical conclusions are also validated for high-dimensional real-world data, which we present in SM \cite{Kopitkov19dm_Supplementary}.

\paragraph{Setup}

\begin{figure}
	\centering
	
	\begin{tabular}{ccccc}
		
		\subfloat[\label{fig:LearningDetails-a}]{\includegraphics[width=0.11\textwidth]{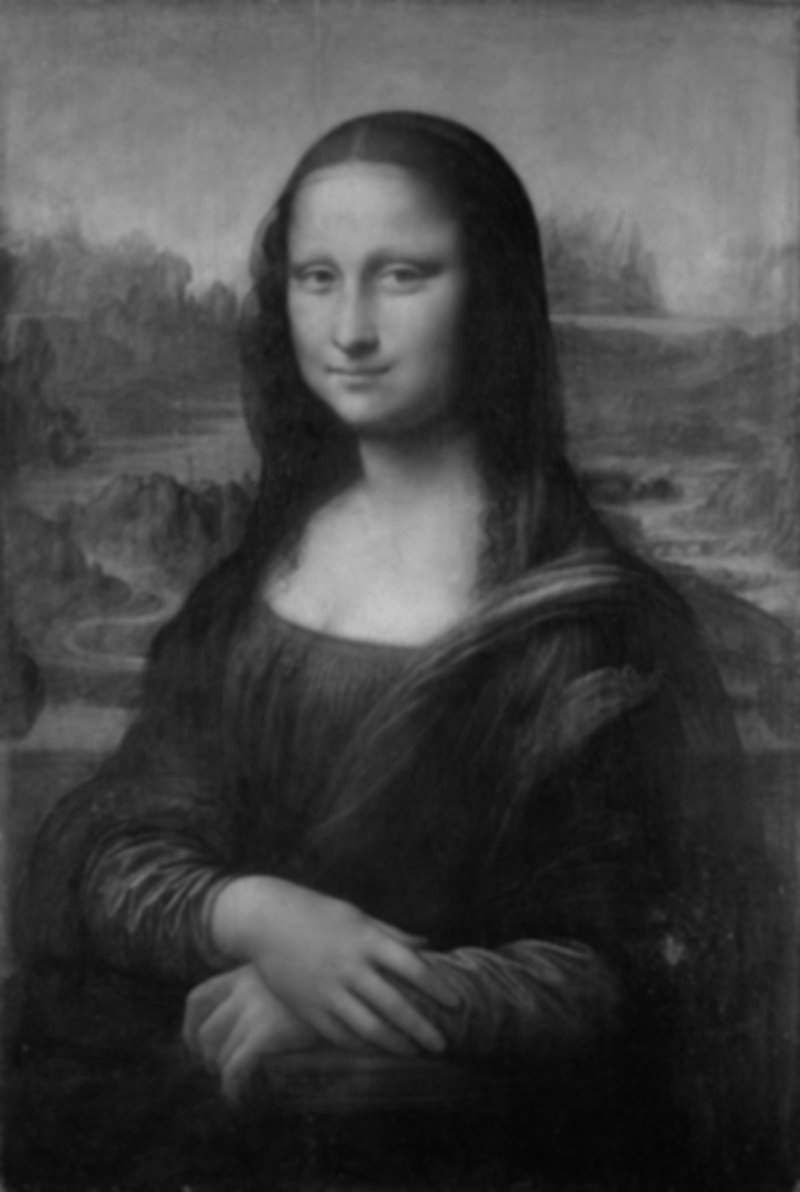}}
		&
		\subfloat[\label{fig:LearningDetails-b}]{\includegraphics[width=0.11\textwidth]{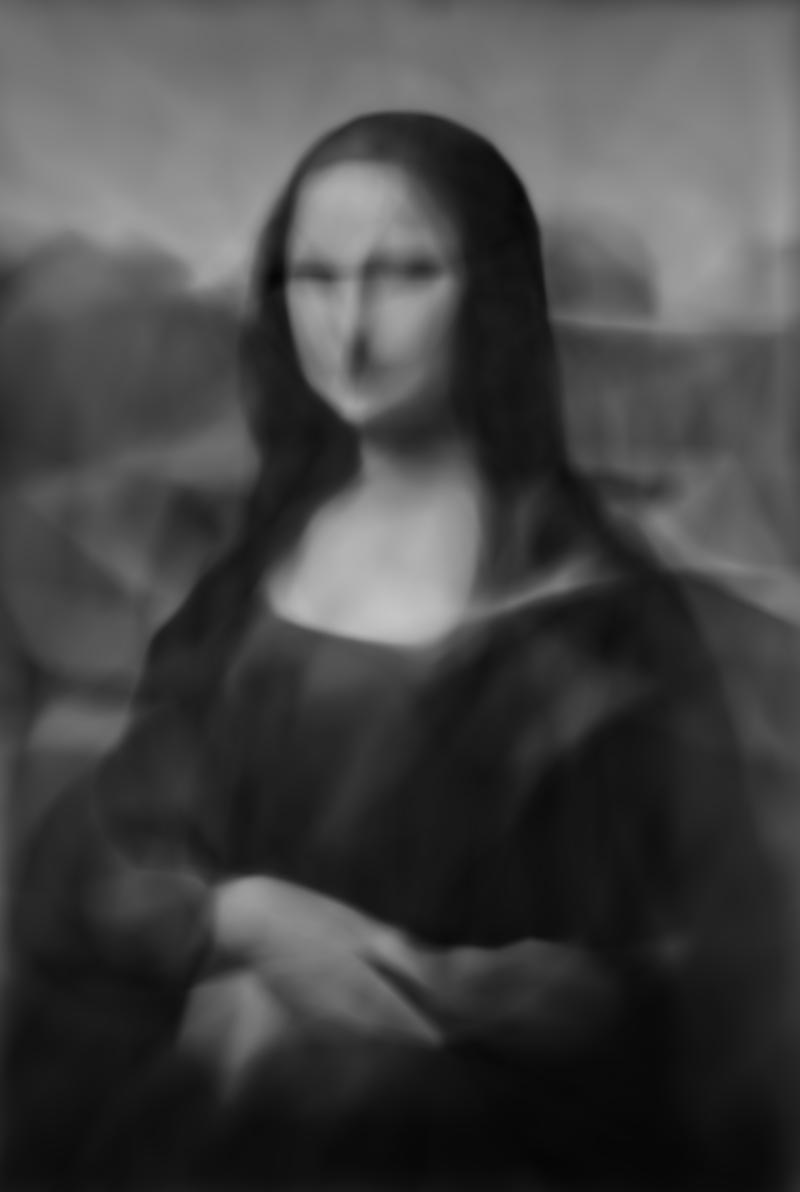}}
		&
		\subfloat[\label{fig:LearningDetails-c}]{\includegraphics[width=0.125\textwidth]{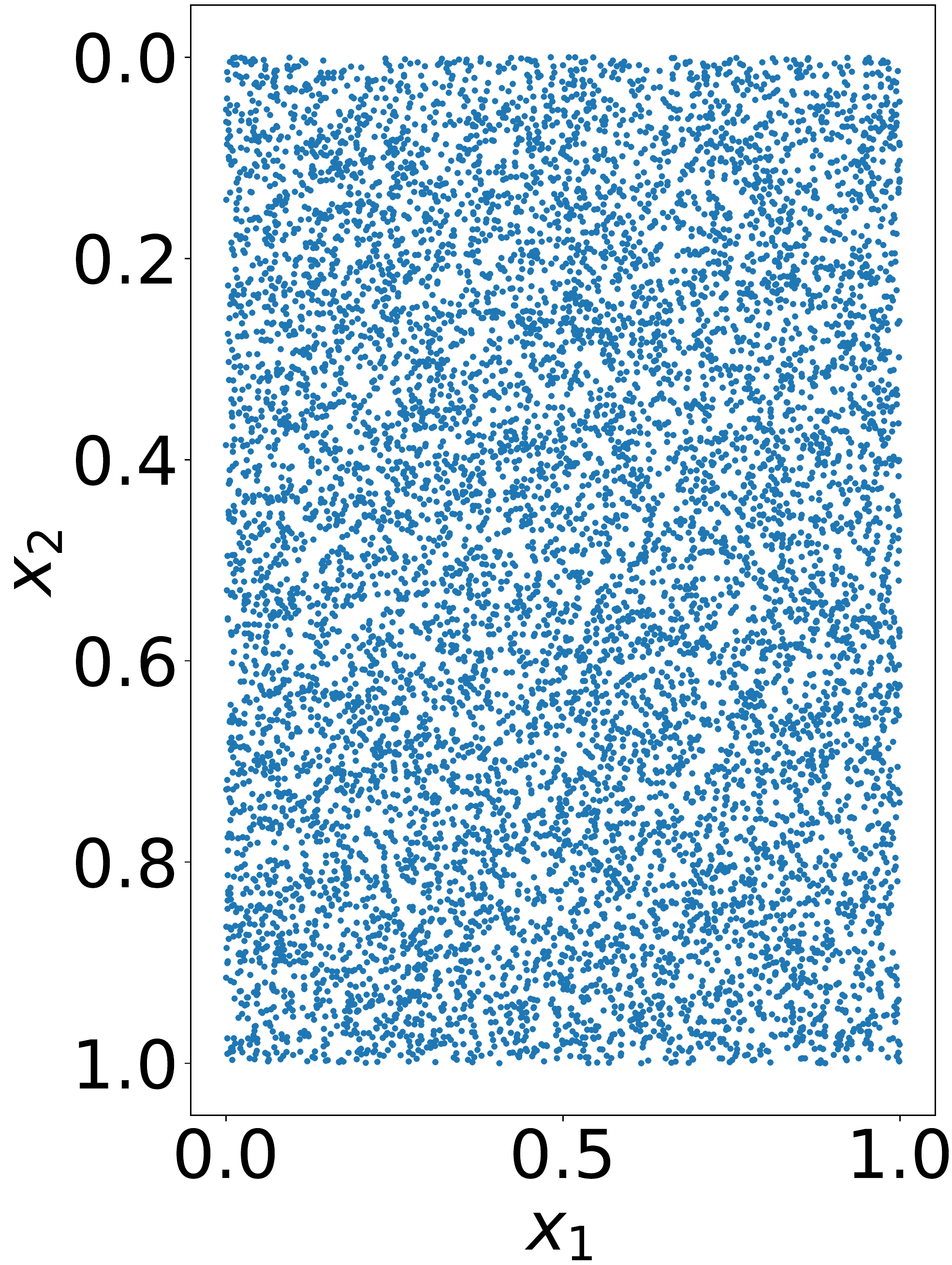}}
		&
		\subfloat[\label{fig:LearningDetails-d}]{\includegraphics[width=0.233\textwidth]{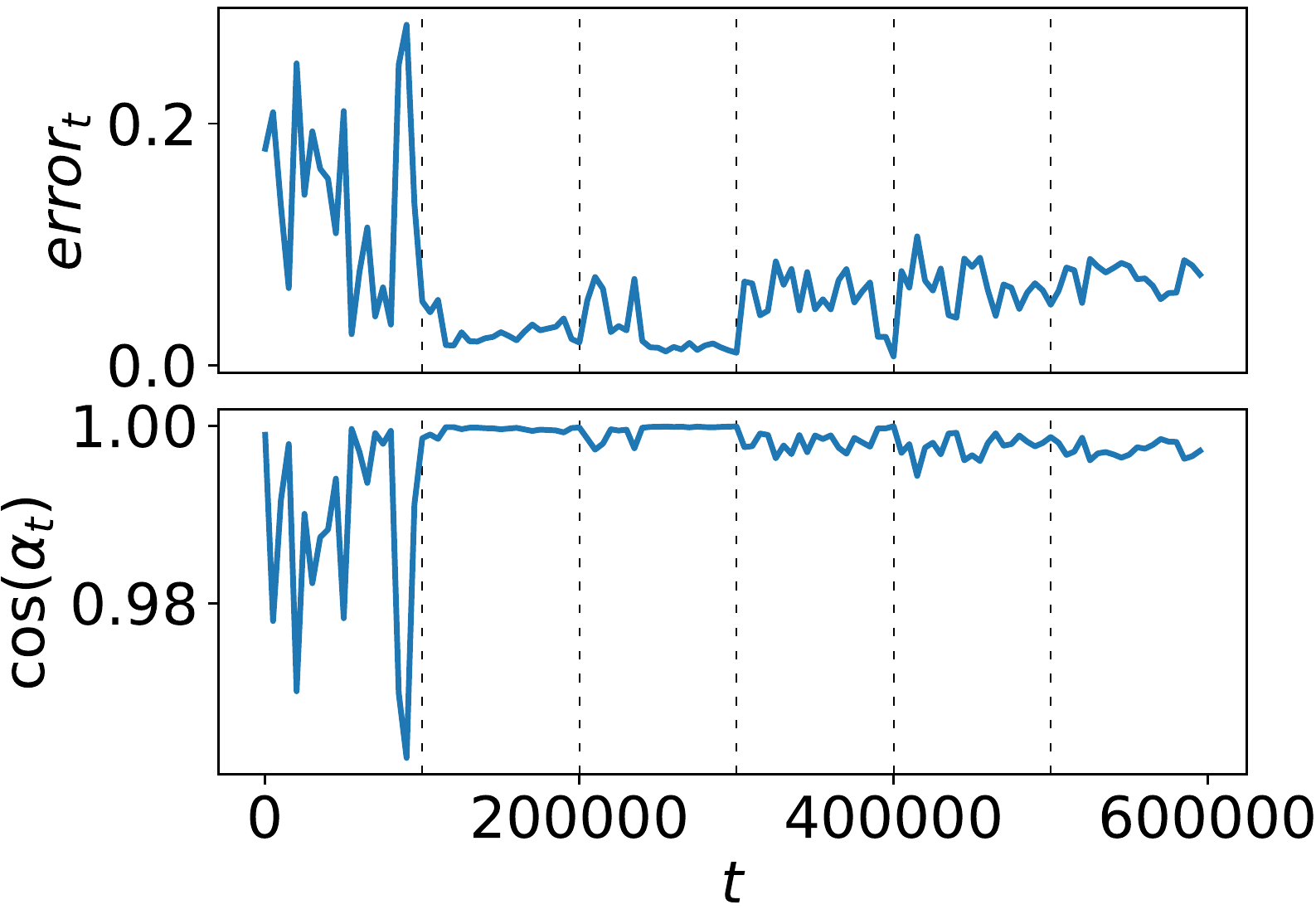}}
		&
		\subfloat[\label{fig:LearningDetails-e}]{\includegraphics[width=0.22\textwidth]{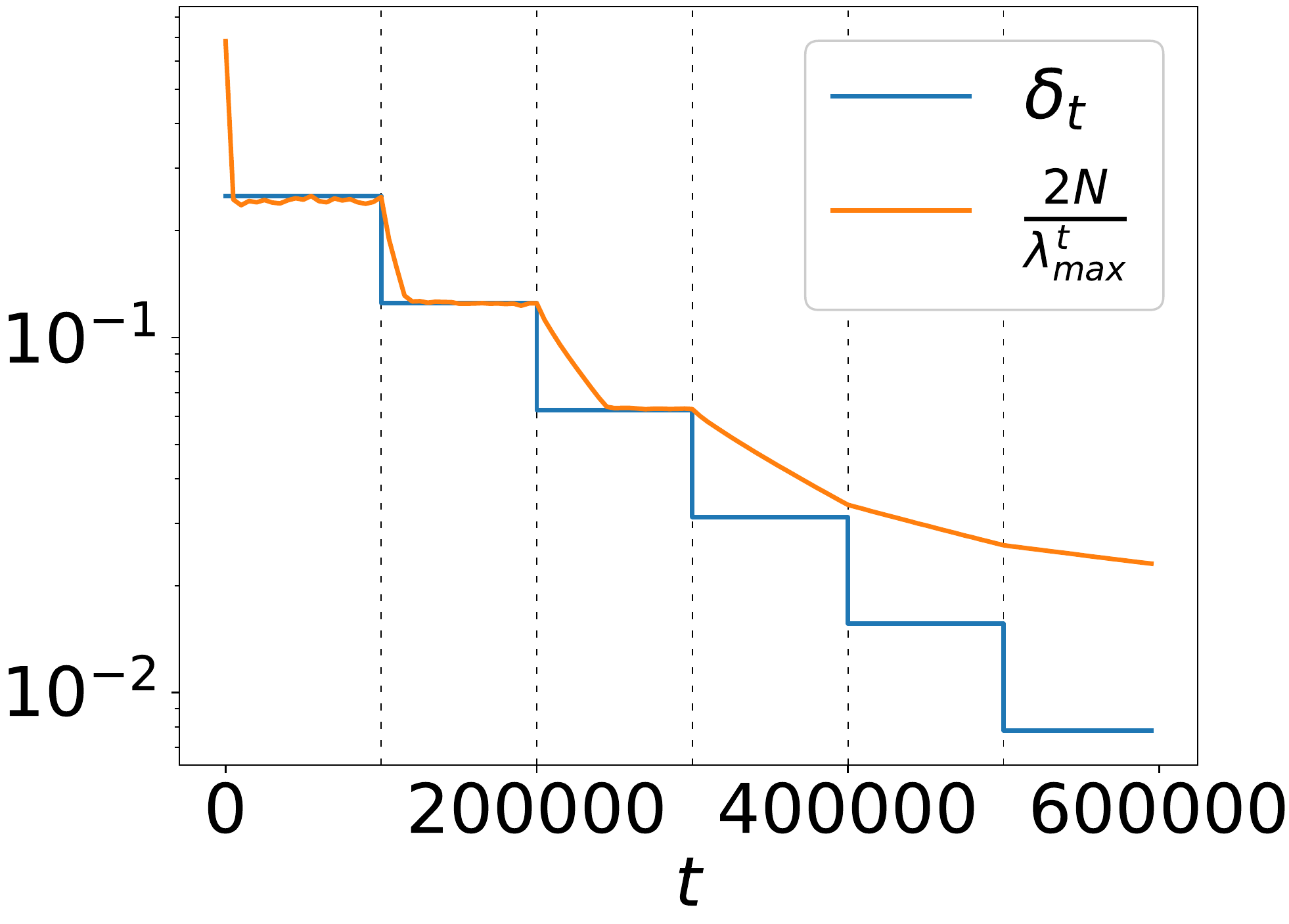}}
		
	\end{tabular}

	\protect
	\caption{(a) Mona Lisa target function for a regression task. (b) NN $f_{\theta}(X)$ at convergence. (c) $10^4$ sampled training points. 
	(d) Accuracy of first order dynamics in Eq.~(\ref{eq:FVecDff}). Depicted is $error_t = \frac{\norm{d \tilde{f}_{t} - d \bar{f}_{t}}}{\norm{d \tilde{f}_{t}}}$, where $d \bar{f}_{t} = -
		\frac{\delta_t}{N}
		\cdot
		G_{t}
		\cdot
		\bar{m}_{t}$ is the first-order approximation of a real differential $d  \tilde{f}_{t} \triangleq \bar{f}_{t + 1}
		-
		\bar{f}_{t}
		$; $\cos \left( \alpha_t \right)$ is cosine of an angle between $d \tilde{f}_{t}$ and $d \bar{f}_{t}$. 
		As observed, Eq.~(\ref{eq:FVecDff}) explains roughly $90 \%$ of NN change.
		(e) Learning rate $\delta_t$ and its upper stability boundary $\frac{2N}{\lambda_{max}^{t}}$ along the optimization. We empirically observe a relation $\lambda_{max}^{t} \propto \frac{1}{\delta_t}$. 
	}
	\label{fig:LearningDetails}
\end{figure}

To provide a better intuition, we specifically consider a regression  
of the target function $y(X)$ with $X \in [0, 1]^2 \subseteq \RR^2$ depicted in Figure \ref{fig:LearningDetails-a}. We approximate this function with Leaky-Relu FC network via L2 loss, using $N = 10000$ training points sampled uniformly from $[0, 1]^2$ (see Figure \ref{fig:LearningDetails-c}). Training dataset is normalized to an empirical mean 0 and a standard deviation 1. NN contains 6 layers with 256 neurons each, with $|\theta| = 264193$, that was initialized via Xavier initialization \cite{Glorot10aistats}. Such large NN size was chosen to specifically satisfy an over-parametrized regime $|\theta| \gg N$, typically met in DL community. Further, learning rate $\delta$ starts at $0.25$ and is decayed twice each $10^5$ iterations, with the total optimization duration being $6 \cdot 10^5$. At convergence $f_{\theta}(X)$ gets very close to its target, see Figure \ref{fig:LearningDetails-b}. Additionally, in Figure \ref{fig:LearningDetails-d} we show that first-order dynamics in Eq.~(\ref{eq:FVecDff}) describe around 90 percent of the change in NN output along the optimization, leaving another 10 for higher-order Taylor terms. Further, we compute $G_{t}$ and its spectrum along the optimization, and thoroughly analyze them below.

\paragraph{Eigenvalues}

\begin{figure}
	\centering
	
	\begin{tabular}{cccc}
		
		\subfloat[\label{fig:EigvalsEv-a}]{\includegraphics[width=0.179\textwidth]{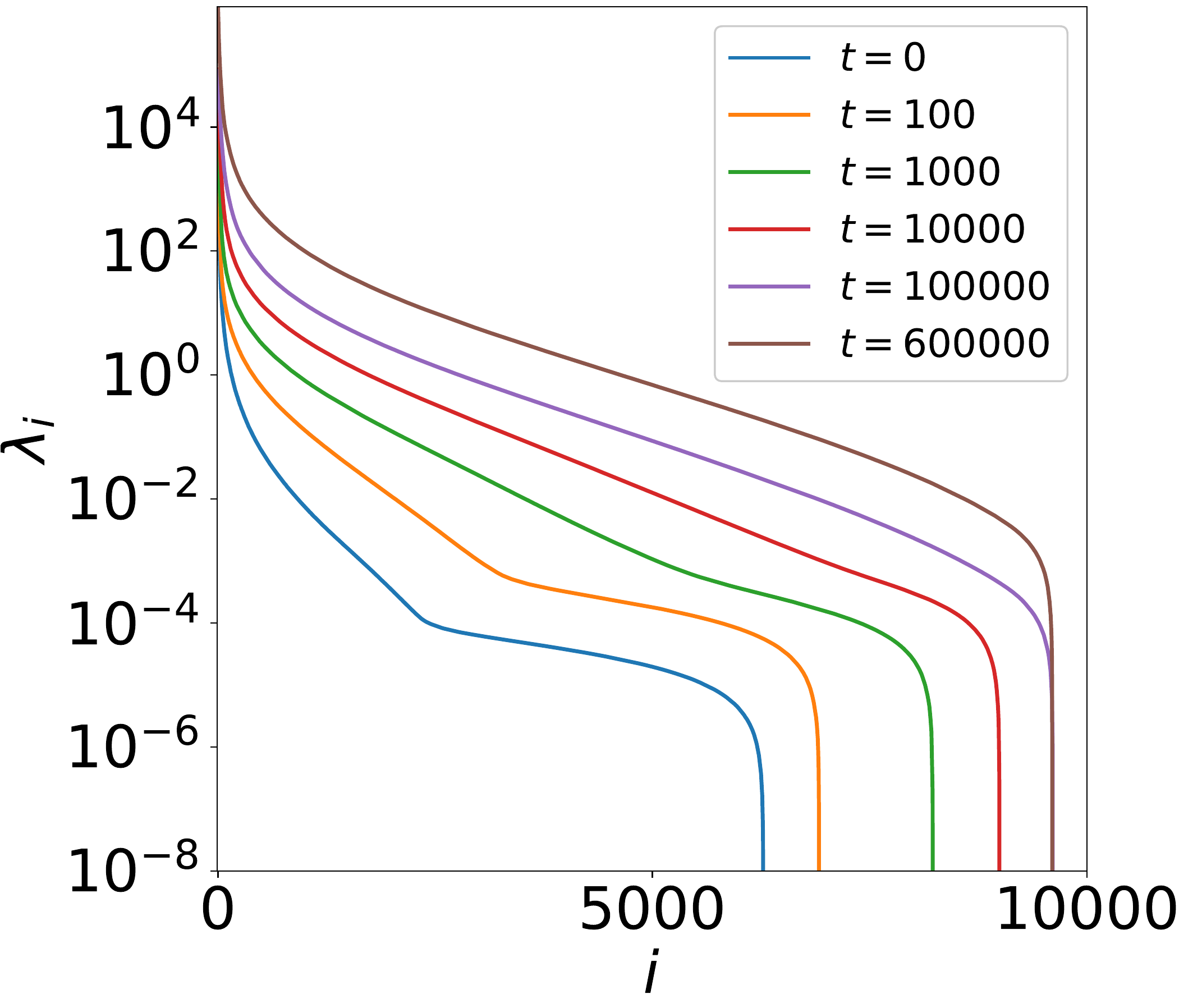}}
		
		&
		
		\subfloat[\label{fig:EigvalsEv-b}]{\includegraphics[width=0.179\textwidth]{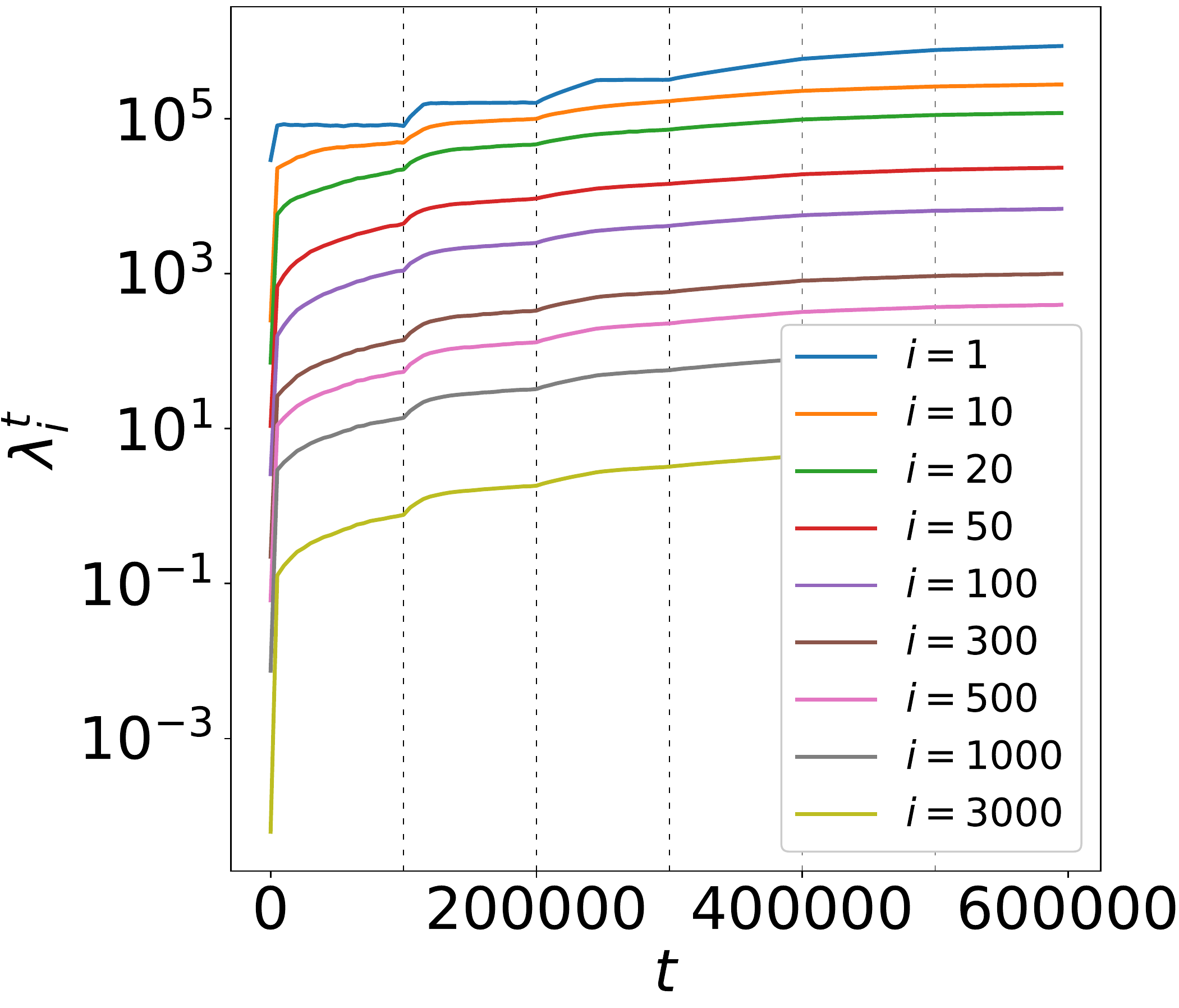}}
		
		&
		
		\subfloat[\label{fig:EigvalsEv-c}]{\includegraphics[width=0.179\textwidth]{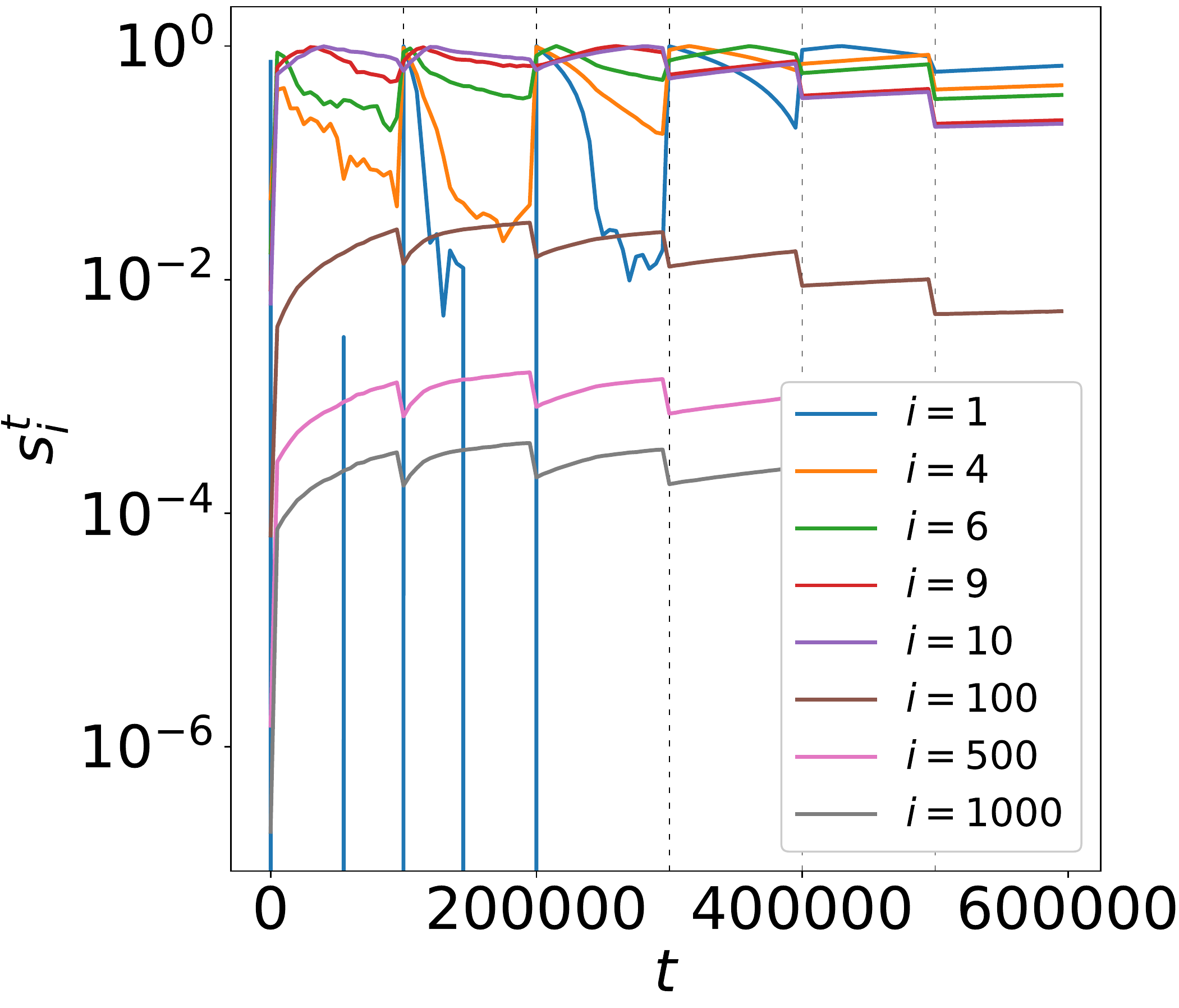}}
		
		&
		
		\subfloat[\label{fig:EigvalsEv-d}]{\includegraphics[width=0.18\textwidth]{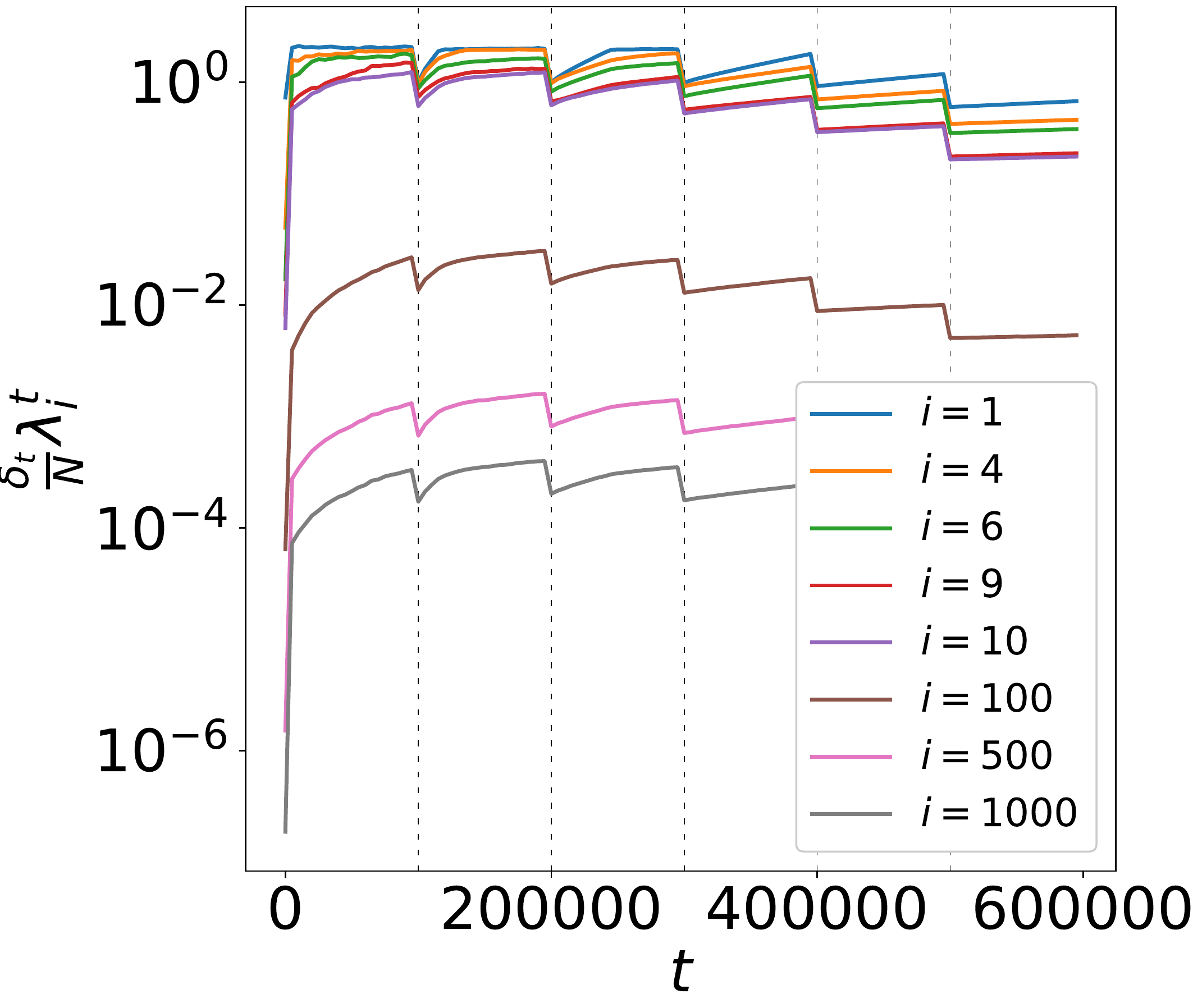}
			\includegraphics[width=0.176\textwidth]{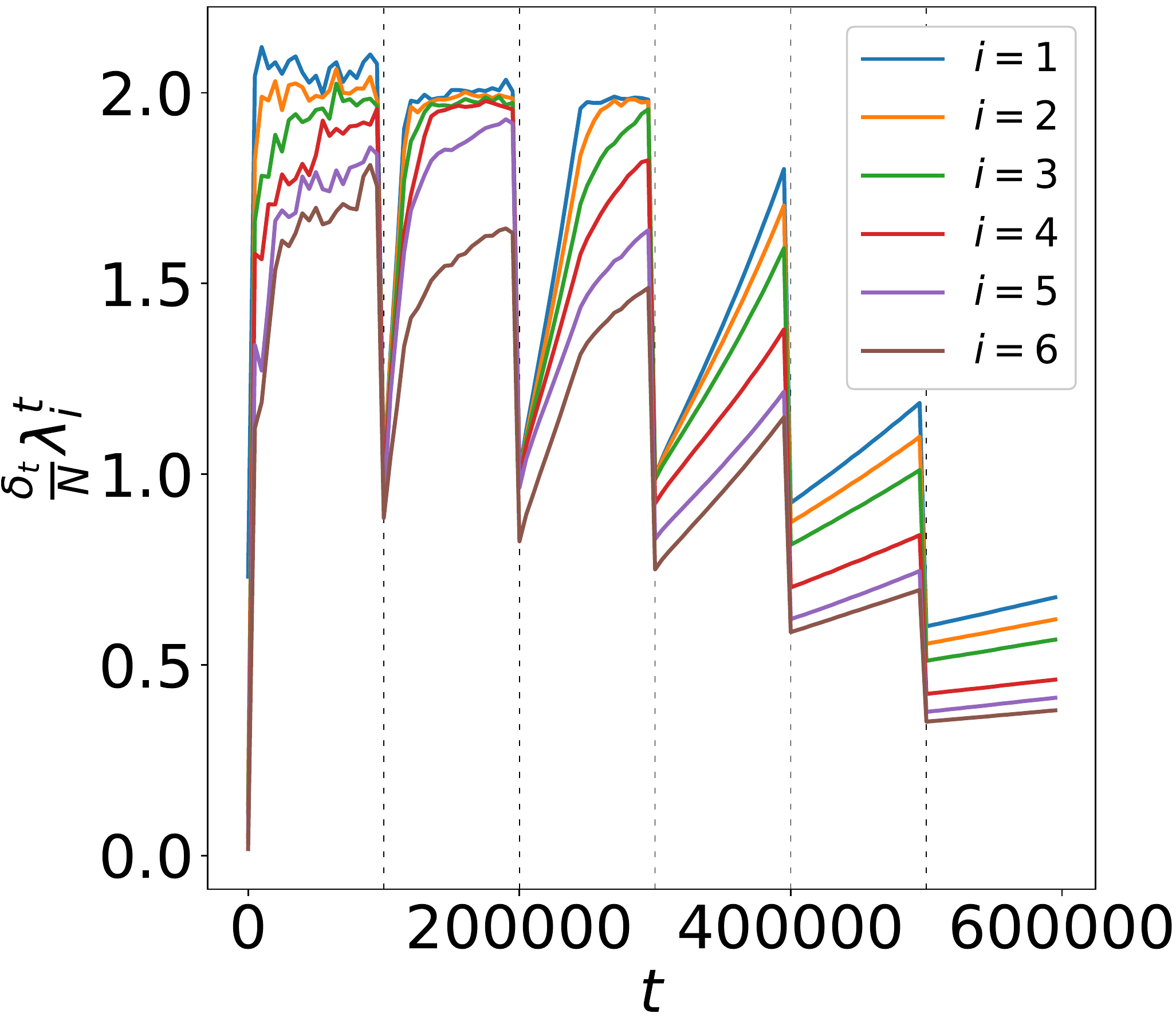}}
		
	\end{tabular}

	\protect
	\caption{(a) Eigenvalues $\{ \lambda_{i}^t \}_{i = 1}^{N}$ for different $t$. (b) Individual eigenvalues along $t$. As observed, eigenvalues monotonically grow along $t$, with growing boost at times of the learning rate drop.
		(c) The information flow speed $s_i^t$ discussed in Section \ref{sec:L2Loss} for several \topp eigenvectors. For first 8 eigenvectors, roughly, this speed is increased at learning rate drop. (d) $\frac{\delta_t}{N}
		\lambda_i^t$ along time $t$, for various $i$.
	}
	\label{fig:EigvalsEv}
\end{figure}

In Figures \ref{fig:EigvalsEv-a}-\ref{fig:EigvalsEv-b} it is shown that each eigenvalue is monotonically increasing along $t$. Moreover, at learning rate decay there is an especial boost in its growth. Since $\frac{\delta_t}{N}
\lambda_i^t$ also defines a speed of movement in $\theta$-space along one of FIM eigenvectors (see Section \ref{sec:FIMSec}), such behavior of eigenvalues suggests an existence of mechanism that keeps a roughly constant movement speed of $\theta$ within $\RR^{|\theta|}$. To do that, when $\delta_t$ is reduced, this mechanism is responsible for increase of $\{ \lambda_{i}^t \}_{i = 1}^{N}$ as a compensation. This is also supported by Figure \ref{fig:EigvalsEv-d} where
each $\frac{\delta_t}{N}
\lambda_i^t$ is balancing, roughly, around the same value along the entire optimization. Furthermore, in Figure \ref{fig:LearningDetails-e} it is clearly observed that an evolution of $\lambda_{max}^t$ stabilizes\footnote{Trend $\lambda_{max}^t \rightarrow \frac{2 N}{\delta_t}$ was consistent in FC NNs for a wide range of initial learning rates, number of layers and neurons, and various datasets (see SM \cite{Kopitkov19dm_Supplementary}), making it an interesting venue for a future theoretical investigation} only when it reaches value of $\frac{2 N}{\delta_t}$, further supporting the above hypothesis.

\paragraph{Neural Spectrum Alignment}

\begin{figure}
	\centering
	
	\newcommand{\width}[0] {0.23}
	
	\begin{tabular}{cc}

		\subfloat[\label{fig:NNSpectrum1.0-a}]{\includegraphics[width=\width\textwidth]{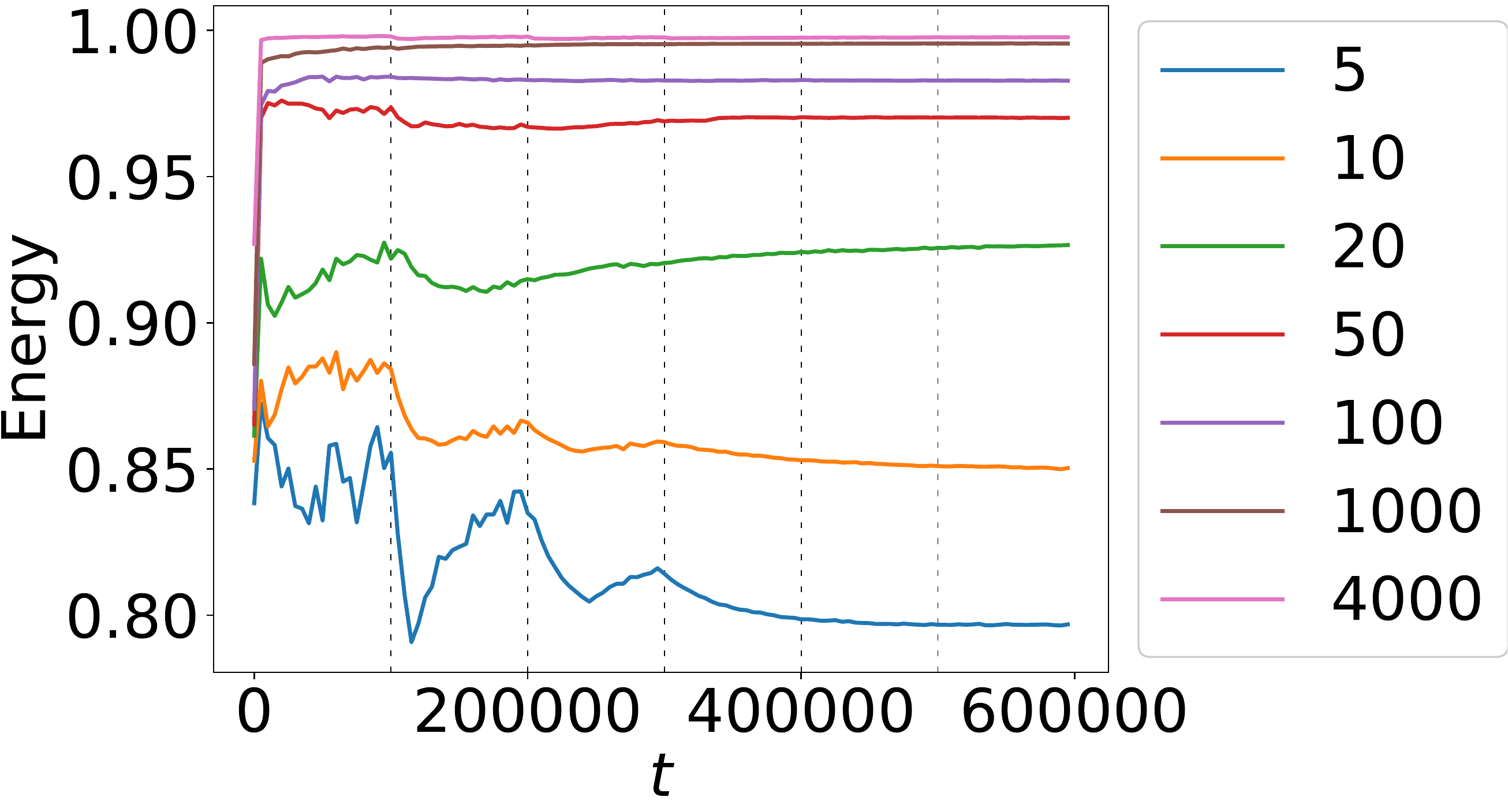}
			\includegraphics[width=\width\textwidth]{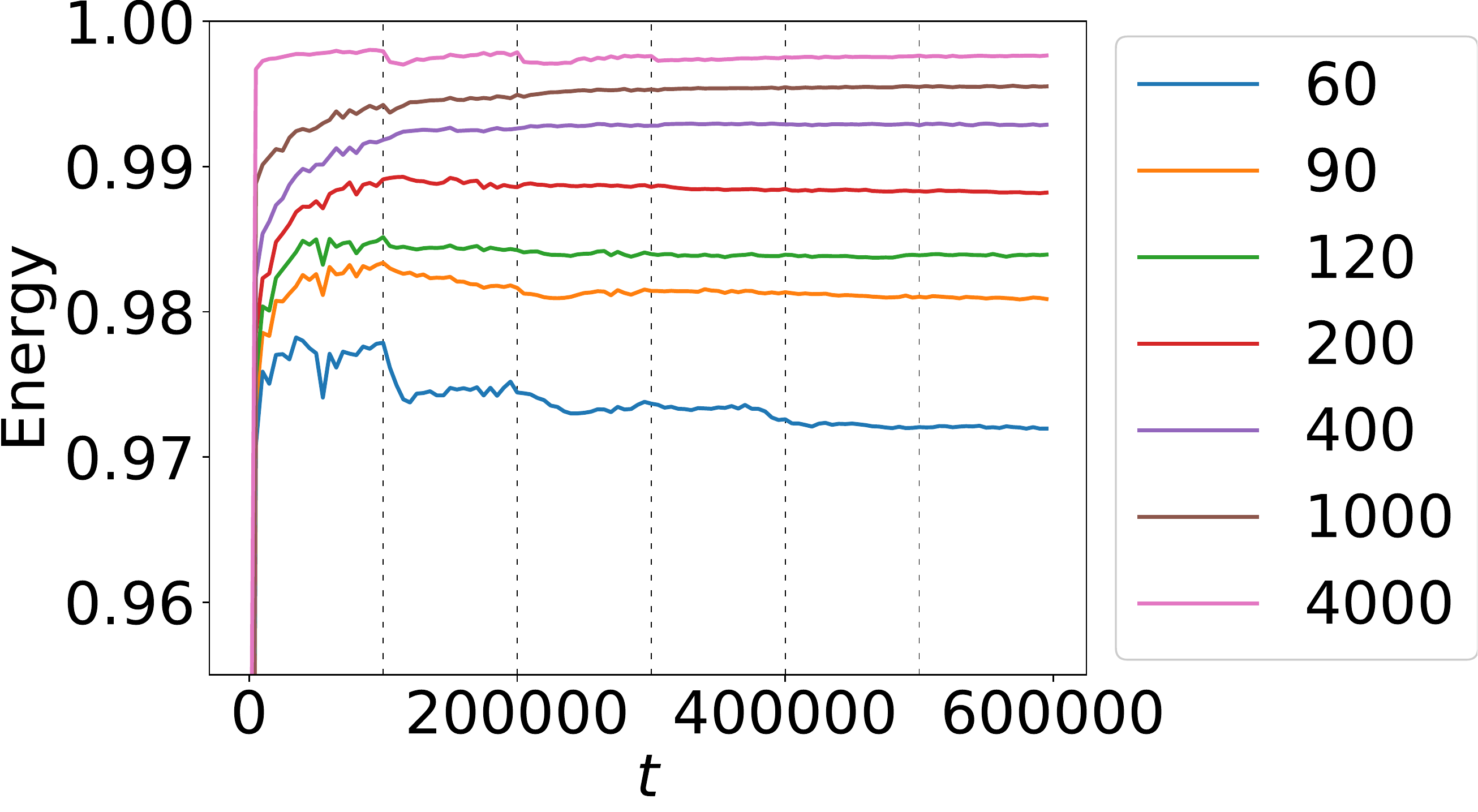}}
		
		&
		\subfloat[\label{fig:NNSpectrum1.0-b}]{\includegraphics[width=\width\textwidth]{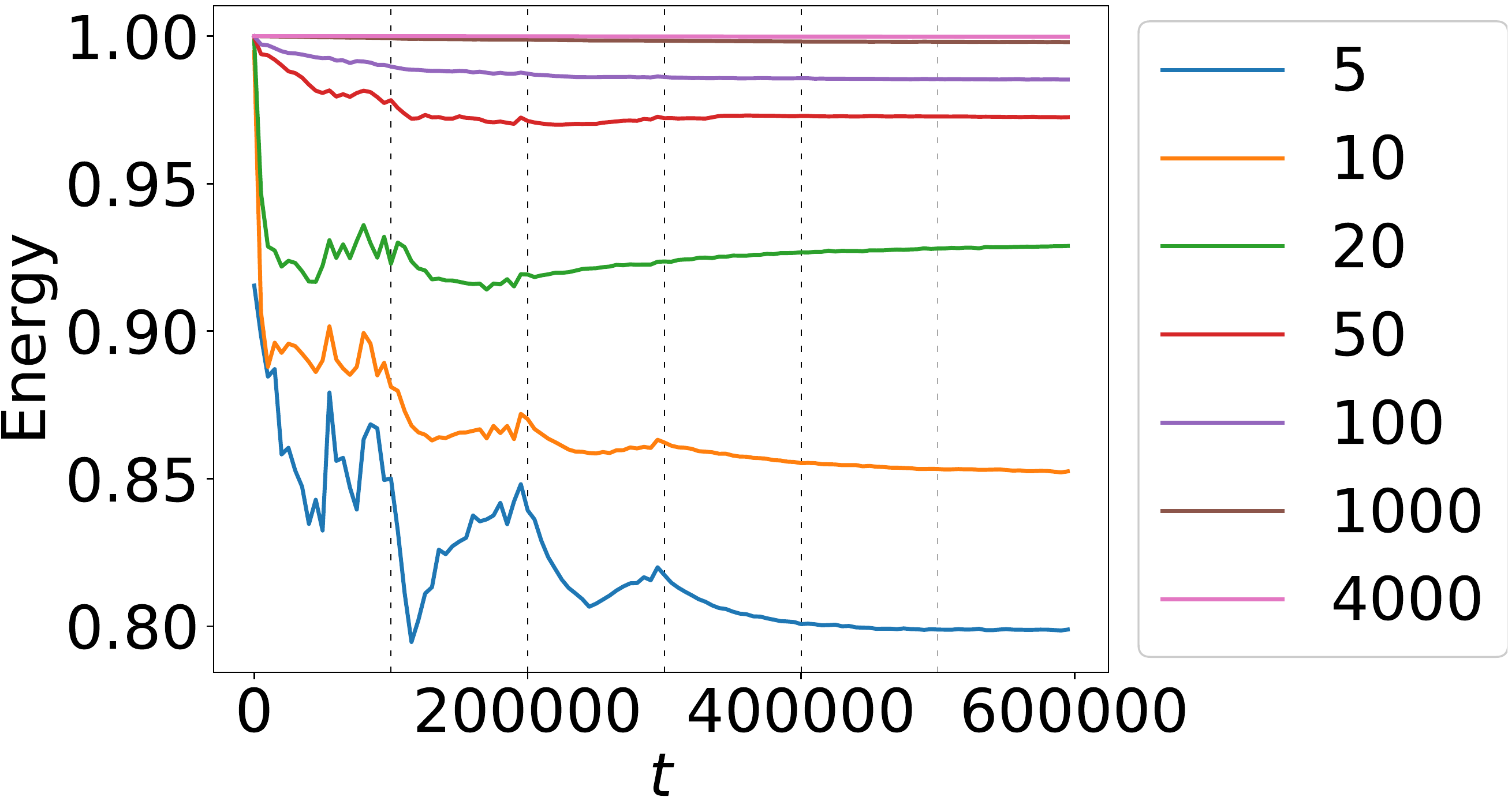}
			\includegraphics[width=\width\textwidth]{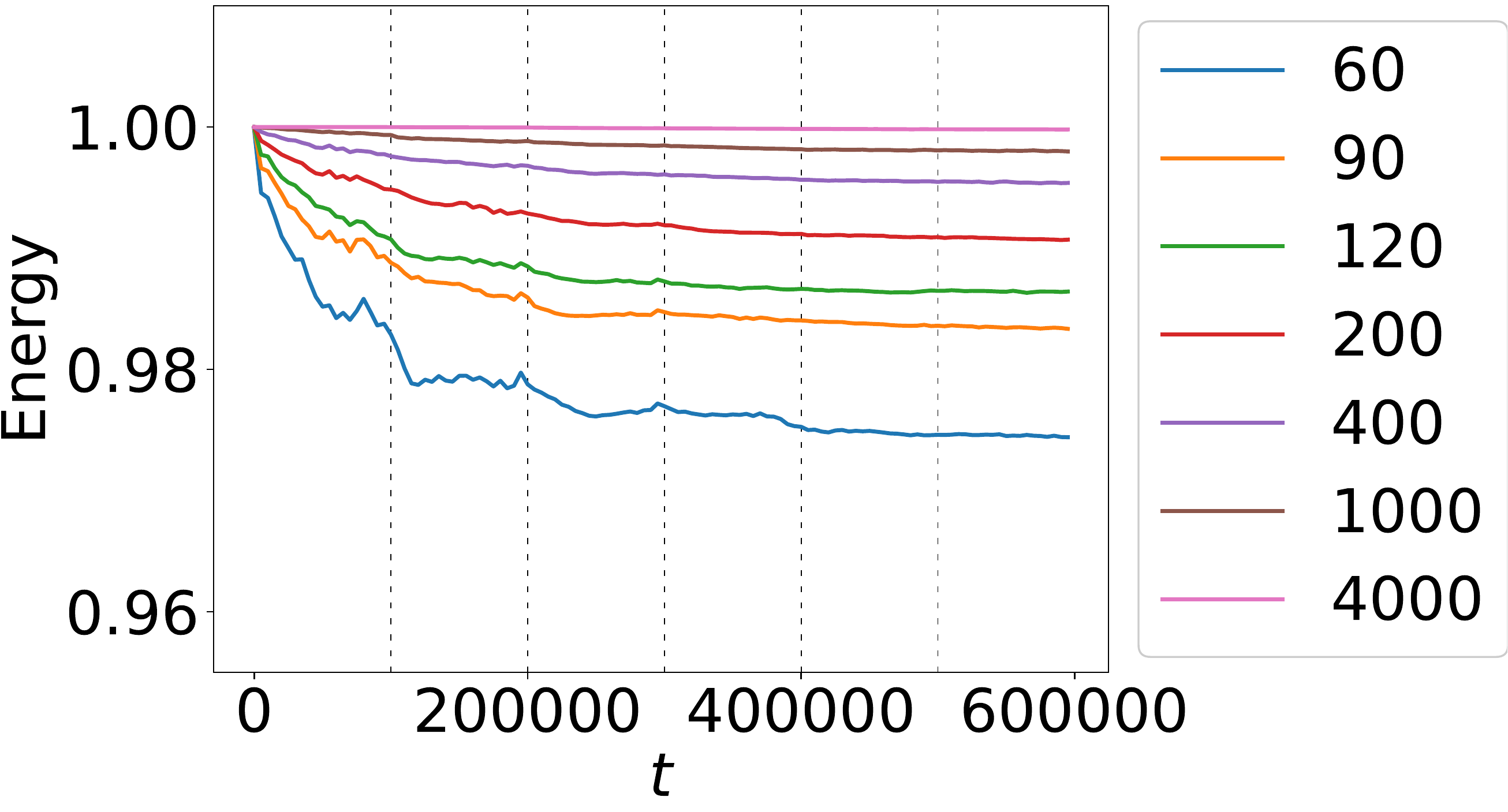}}
		
		\\
		\subfloat[\label{fig:NNSpectrum1.0-c}]{\includegraphics[width=\width\textwidth]{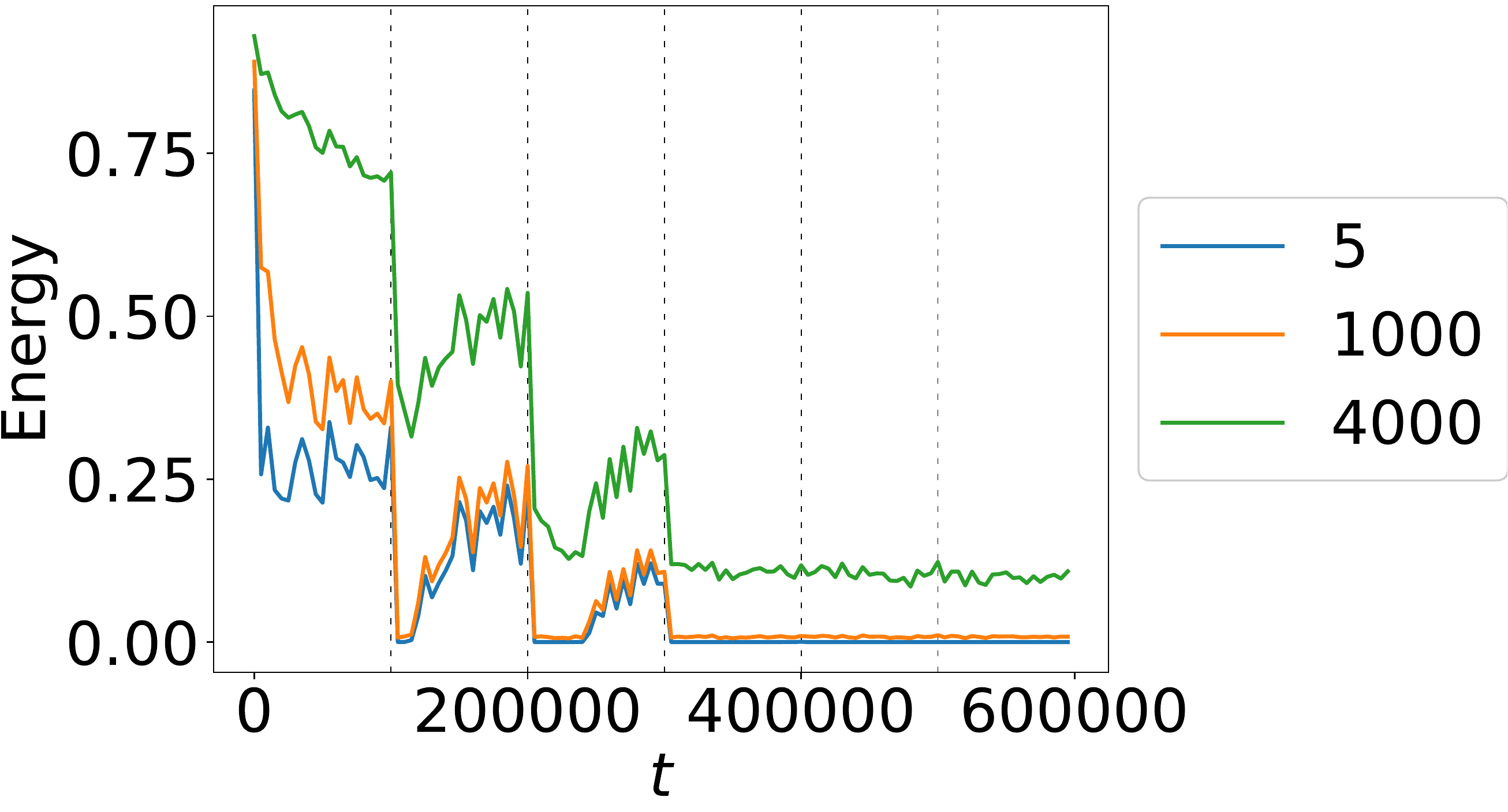}
			\includegraphics[width=\width\textwidth]{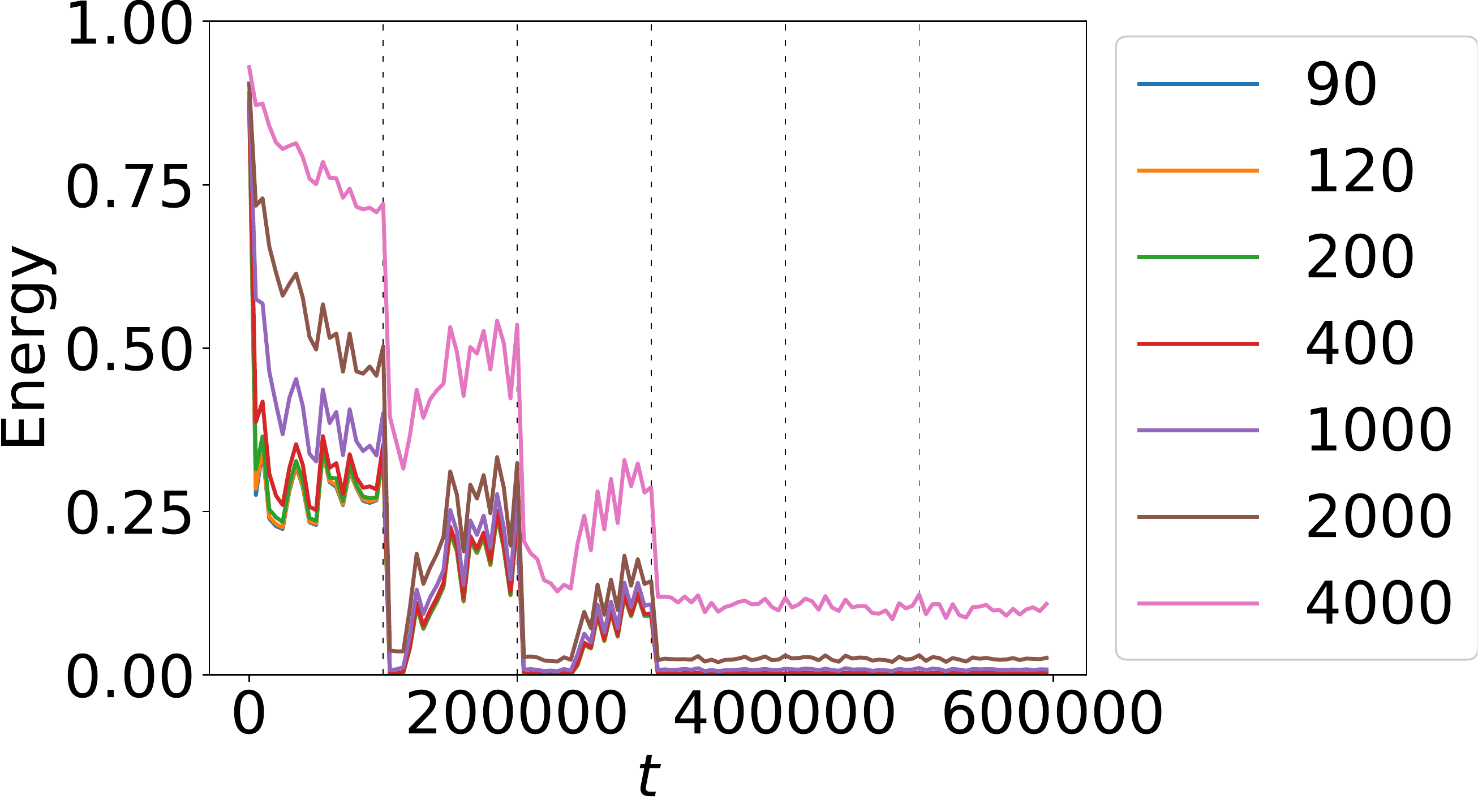}}
		
		&
		\subfloat[\label{fig:NNSpectrum1.0-d}]{\includegraphics[width=\width\textwidth]{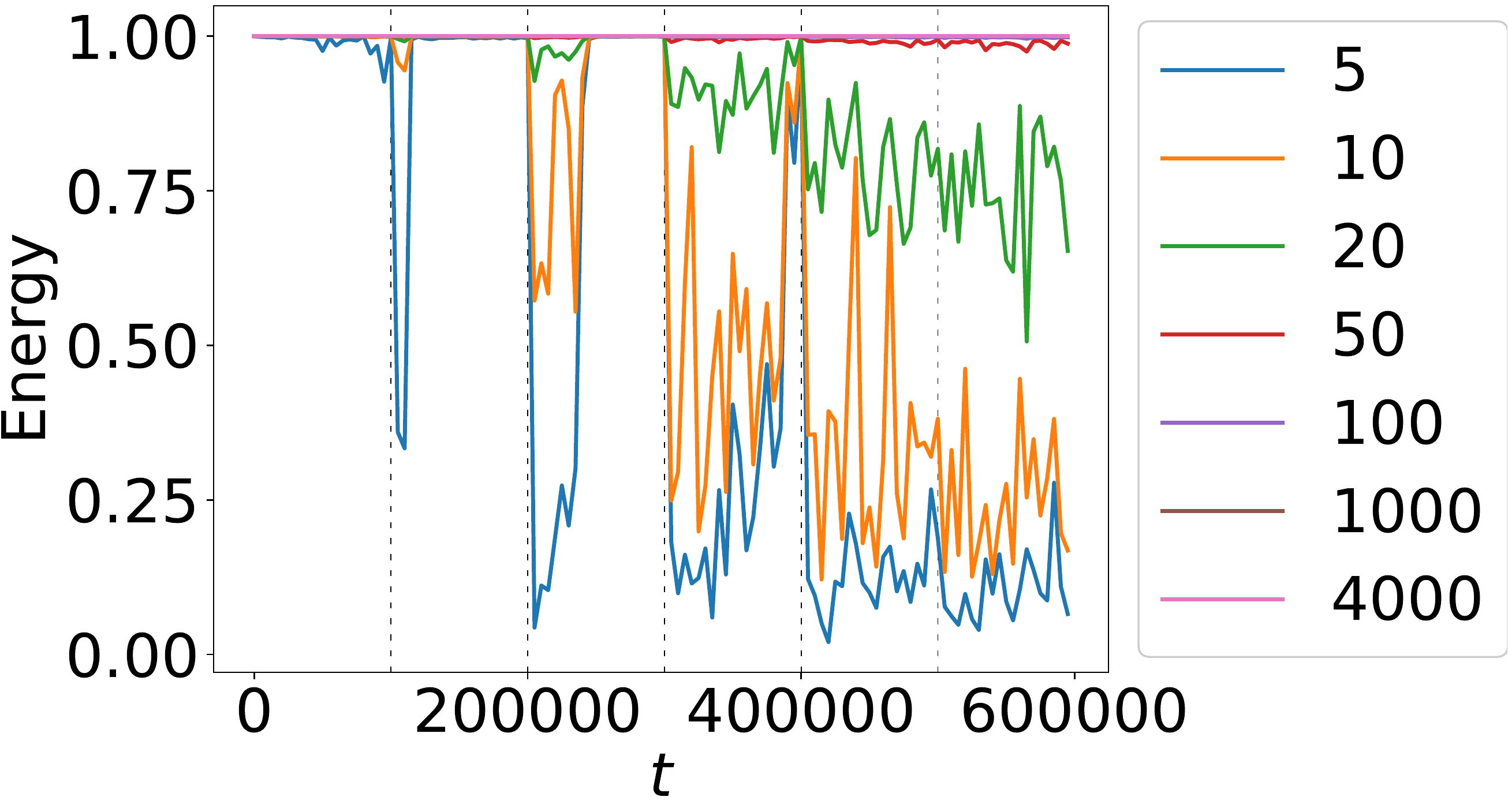}
			\includegraphics[width=\width\textwidth]{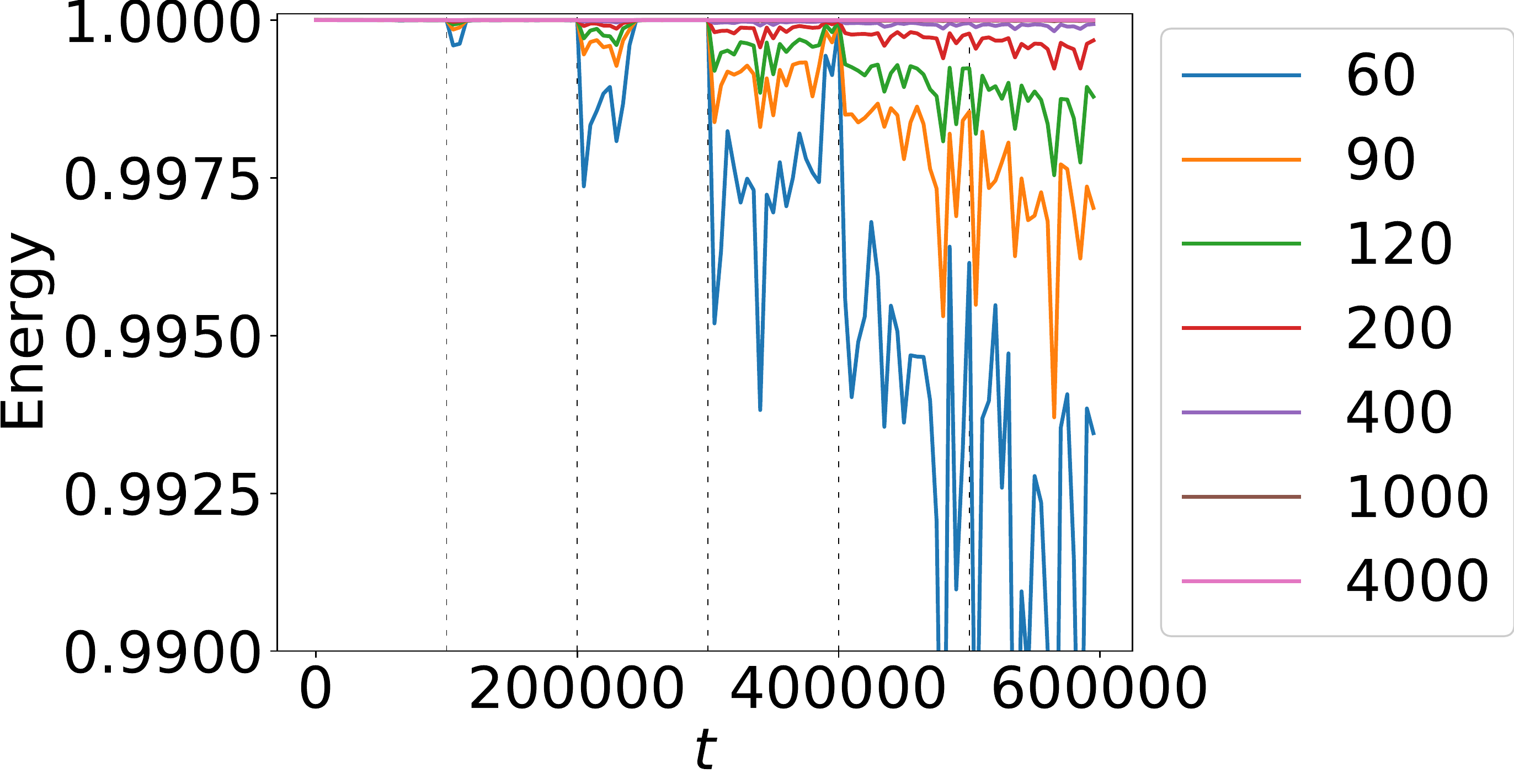}}
		
	\end{tabular}

	\begin{tabular}{c}
		\subfloat[\label{fig:NNSpectrum1.0-e}]{\includegraphics[width=\width\textwidth]{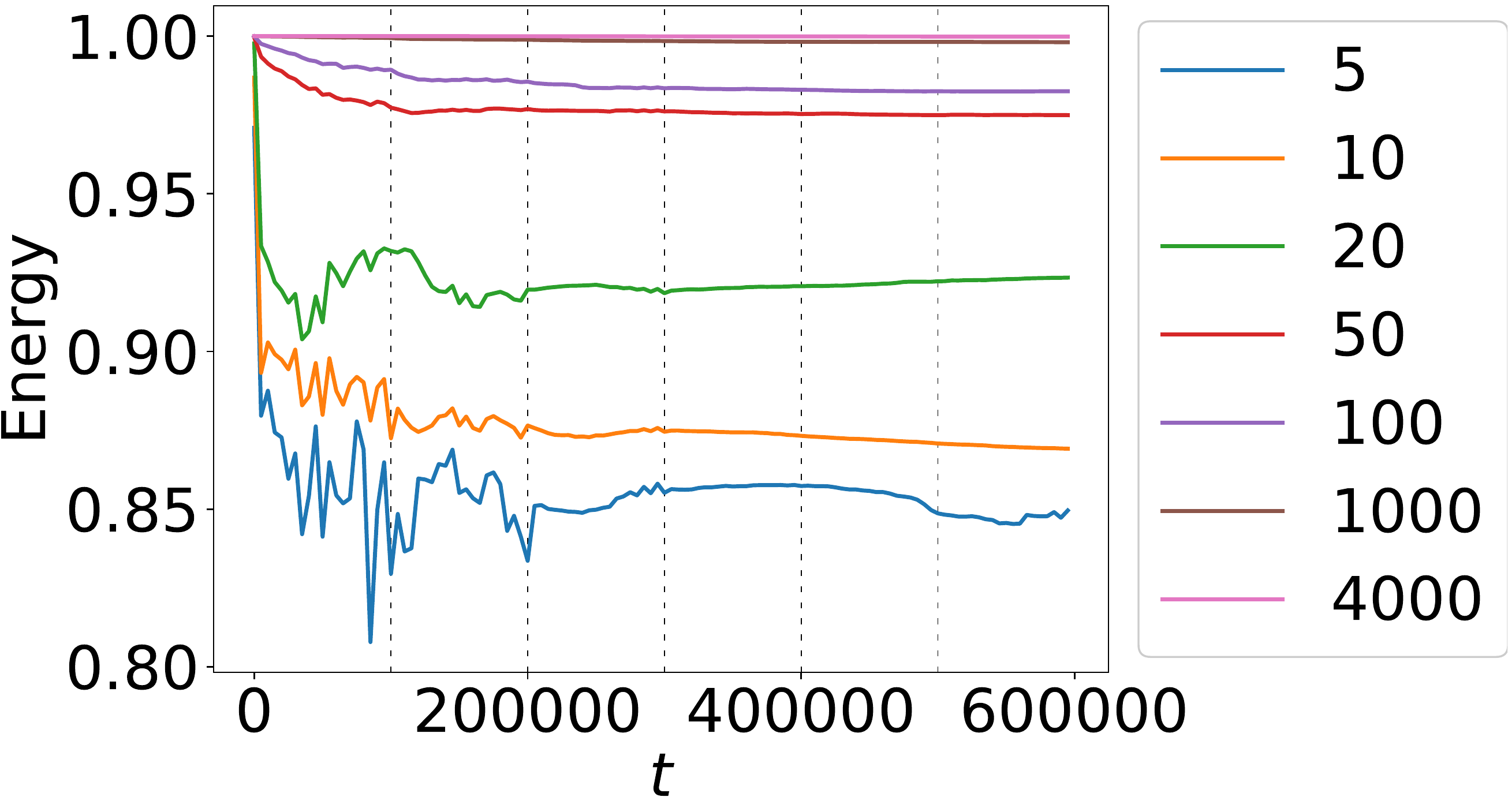}
			\includegraphics[width=\width\textwidth]{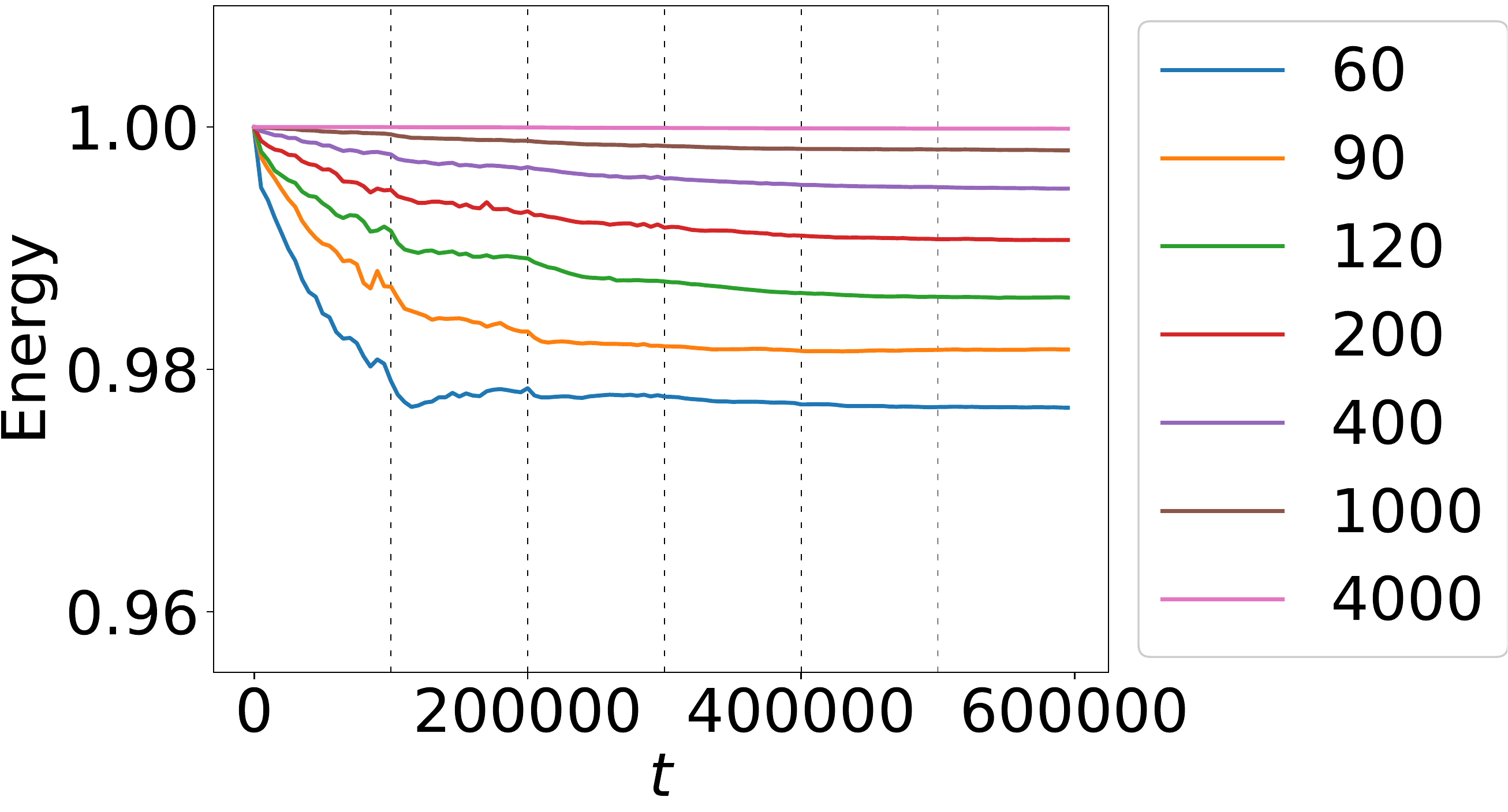}}
	\end{tabular}

	\protect
	\caption{(a) For different $k$, relative energy of the label vector $\bar{y}$ in \topp $k$ eigenvectors of $G_t$, $E_t(\bar{y}, k)$, along the optimization time $t$. (b) Relative energy of NN output, $E_t(\bar{f}_{t}, k)$. 
		(c) Relative energy of the residual, $E_t(\bar{m}_{t}, k)$.
		(d) Relative energy of the differential $d \bar{f}_{t} = -
		\frac{\delta_t}{N}
		\cdot
		G_{t}
		\cdot
		\bar{m}_{t}$, $E_t(d \bar{f}_{t}, k)$.
		(e) Relative energy of NN output, $E_t(\bar{f}_{t}^{test}, k)$, with both $G_t$ and $\bar{f}_{t}^{test}$ computed at $10^4$ testing points. 
		Dashed vertical lines depict time $t$ at which learning rate $\delta$ was decayed (see Figure \ref{fig:LearningDetails-e}).
	}
	\label{fig:NNSpectrum1.0}
\end{figure}

Notate by $\cos \left[ \alpha_t \left( \bar{\phi}, k \right) \right] \triangleq \sqrt{\frac{\sum_{i = 1}^{k}
		<\bar{\upsilon}_{i}^t, \bar{y}>^2}{\norm{\bar{\phi}}_2^2}}$ the cosine of an angle $\alpha_t \left( \bar{\phi}, k \right)$ between an arbitrary vector $\bar{\phi}$ and its projection to the sub-space of $\RR^N$ spanned by $\{ \bar{\upsilon}_{i}^t \}_{i = 1}^{k}$.
Further, $E_t(\bar{\phi}, k) \triangleq \cos^2 \left[ \alpha_t \left( \bar{\phi}, k \right) \right]$ can be considered as a \emph{relative energy} of $\bar{\phi}$, the percentage of its energy $\norm{\bar{\phi}}_2^2$ located inside $span \left( \{ \bar{\upsilon}_{i}^t \}_{i = 1}^{k} \right)$. In our experiments we will use $E_t(\bar{\phi}, k)$ as an alignment metric between $\bar{\phi}$ and $\{ \bar{\upsilon}_{i}^t \}_{i = 1}^{k}$.
Further, we evaluate alignment of $G_t$ with $\bar{y}$ instead of $\bar{m}_{0}$ since $\bar{f}_{0}$ is approximately zero in the considered FC networks.

In Figure \ref{fig:NNSpectrum1.0-a} we depict relative energy of the label vector $\bar{y}$ in \topp $k$ eigenvectors of $G_t$, $E_t(\bar{y}, k)$. 
As observed, 20 \topp eigenvectors of $G_t$ contain 90 percent of $\bar{y}$ for almost all $t$. Similarly, 200 \topp eigenvectors of $G_t$ contain roughly 98 percent of $\bar{y}$, with rest of eigenvectors being practically orthogonal w.r.t. $\bar{y}$. That is, $G_t$ aligns its \topp spectrum towards the ground truth target function $\bar{y}$ almost immediately after starting of training, which improves the convergence rate since the information flow is fast along \topp eigenvectors as discussed in Section \ref{sec:L2Loss} and proved in \cite{Arora19arxiv,Oymak19arxiv}.

Further, we can see that for $k < 400$ the relative energy $E_t(\bar{y}, k)$ is decreasing after each decay of $\delta$, yet for $k > 400$ it keeps growing along the entire optimization. Hence, the \topp eigenvectors of $G_t$ can be seen as NN memory that is learned/tuned toward representing the target $\bar{y}$, while after each learning rate drop the learned information is spread more evenly among a higher number of different \topp eigenvectors.

Likewise, in Figure \ref{fig:NNSpectrum1.0-b} we can see that NN outputs vector $\bar{f}_{t}$ is located entirely in a few hundreds of \topp eigenvectors. 
In case we consider $G_t$ to be constant, such behavior can be explained by Eq.~(\ref{eq:FVecDff}) since each increment of $\bar{f}_{t}$, $d \bar{f}_{t}$, is also located almost entirely within \topp 60 eigenvectors of $G_t$ (e.g. see $E_t(d \bar{f}_{t}, 60)$ in Figure \ref{fig:NNSpectrum1.0-d}). Yet, for a general NN with  a time-dependent kernel the theoretical justification for the above empirical observation is currently missing.
Further, similar relation is observed also at points outside of $\dtX$ (see Figure \ref{fig:NNSpectrum1.0-e}), 
leading to the empirical conclusion that \topp eigenfunctions of \emph{gradient similarity} $g_{t}(X, X')$ are the basis functions of NN $f_{\theta}(X)$.

\paragraph{Residual Dynamics}

\begin{figure}
	\centering
	\newcommand{\imagepath}[0] {Figures/FC_NN_6L/time_00595000}

	\newcommand{\width}[0] {0.091}
	
	\begin{tabular}{cccc}

		\subfloat[\label{fig:NNSpectrum3-a}]{\includegraphics[width=0.19\textwidth]{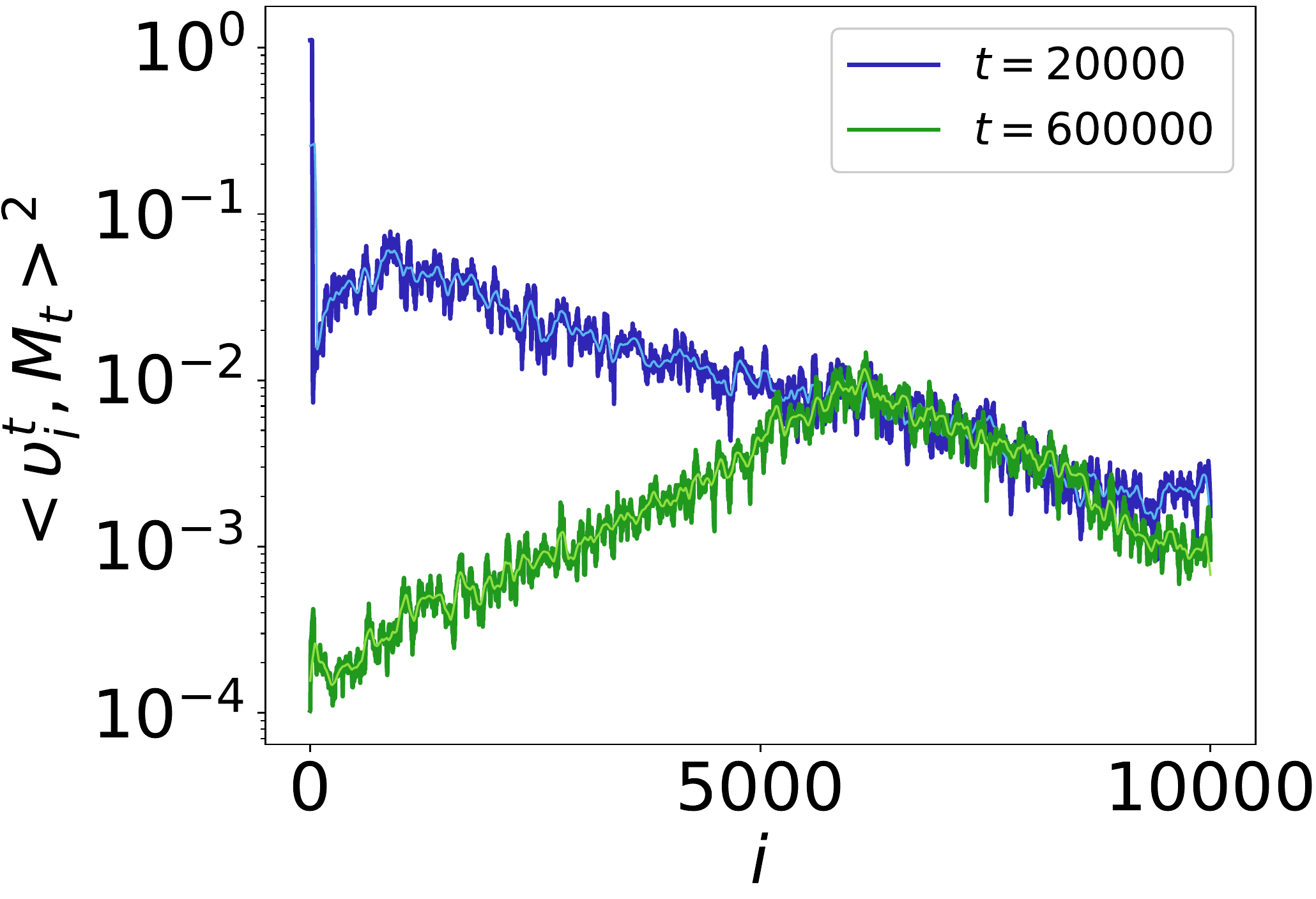}}
		
		&
		
		\subfloat[\label{fig:NNSpectrum3-b}]{\includegraphics[width=0.17\textwidth]{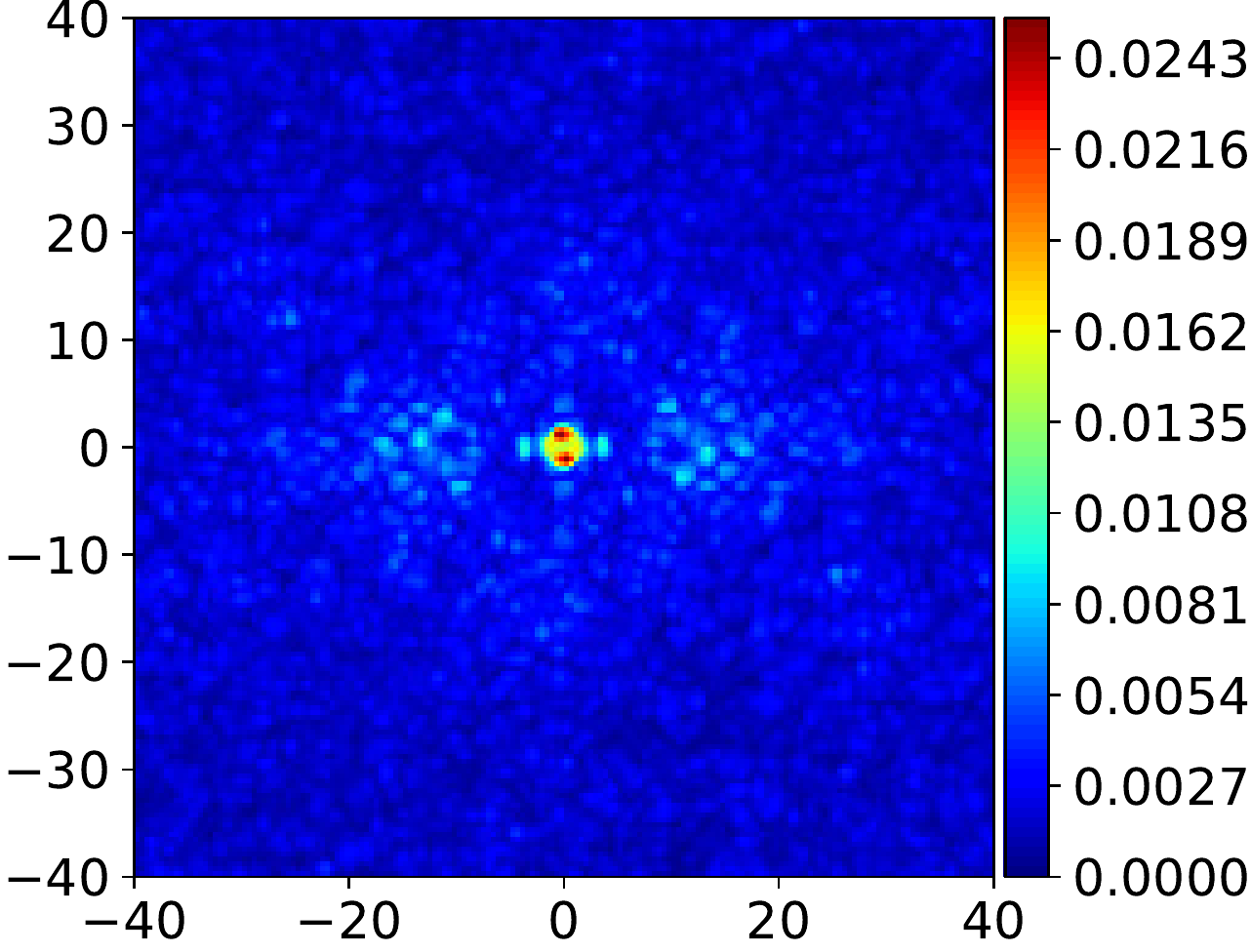}}
		
		&

		\subfloat[\label{fig:NNSpectrum3-c}]{\includegraphics[width=0.15\textwidth]{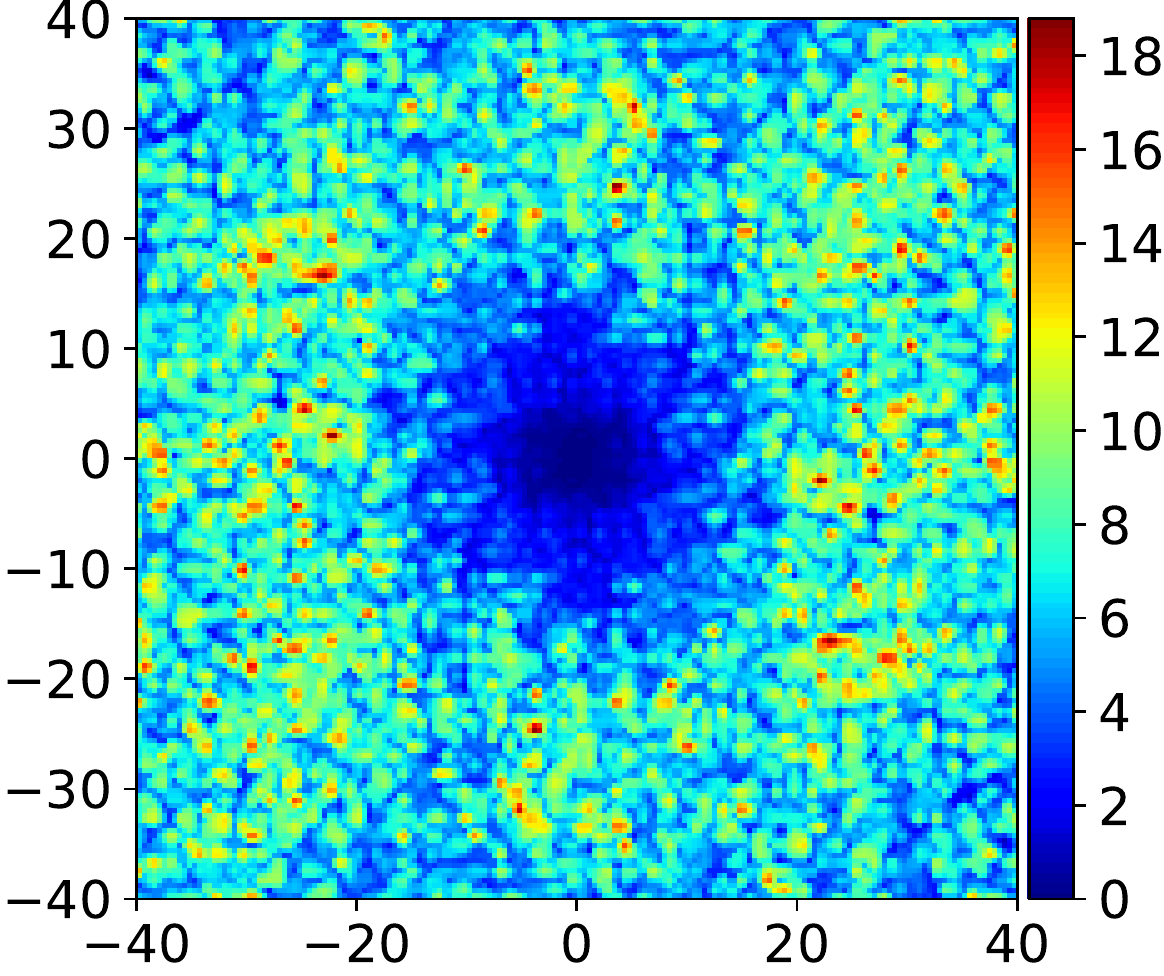}}

		&
		
		\subfloat[\label{fig:NNSpectrum3-d}]{\includegraphics[width=\width\textwidth]{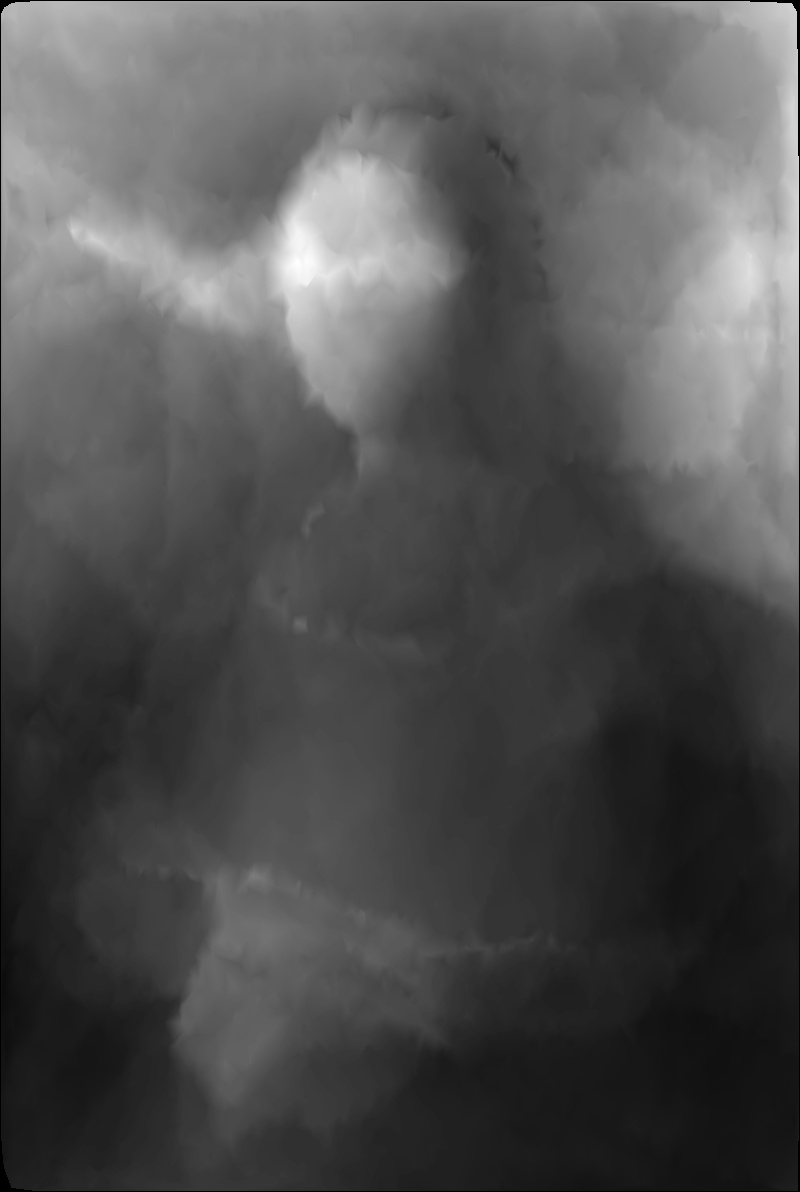}
			\includegraphics[width=\width\textwidth]{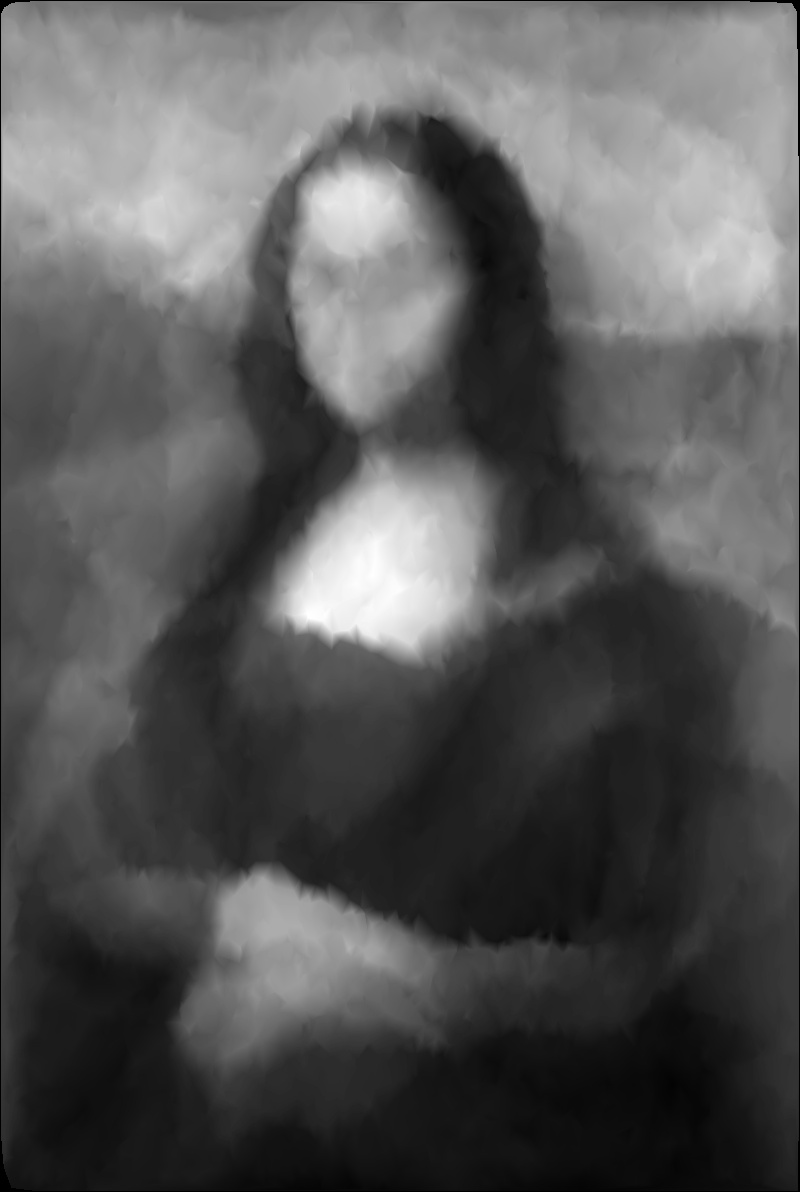}
			\includegraphics[width=\width\textwidth]{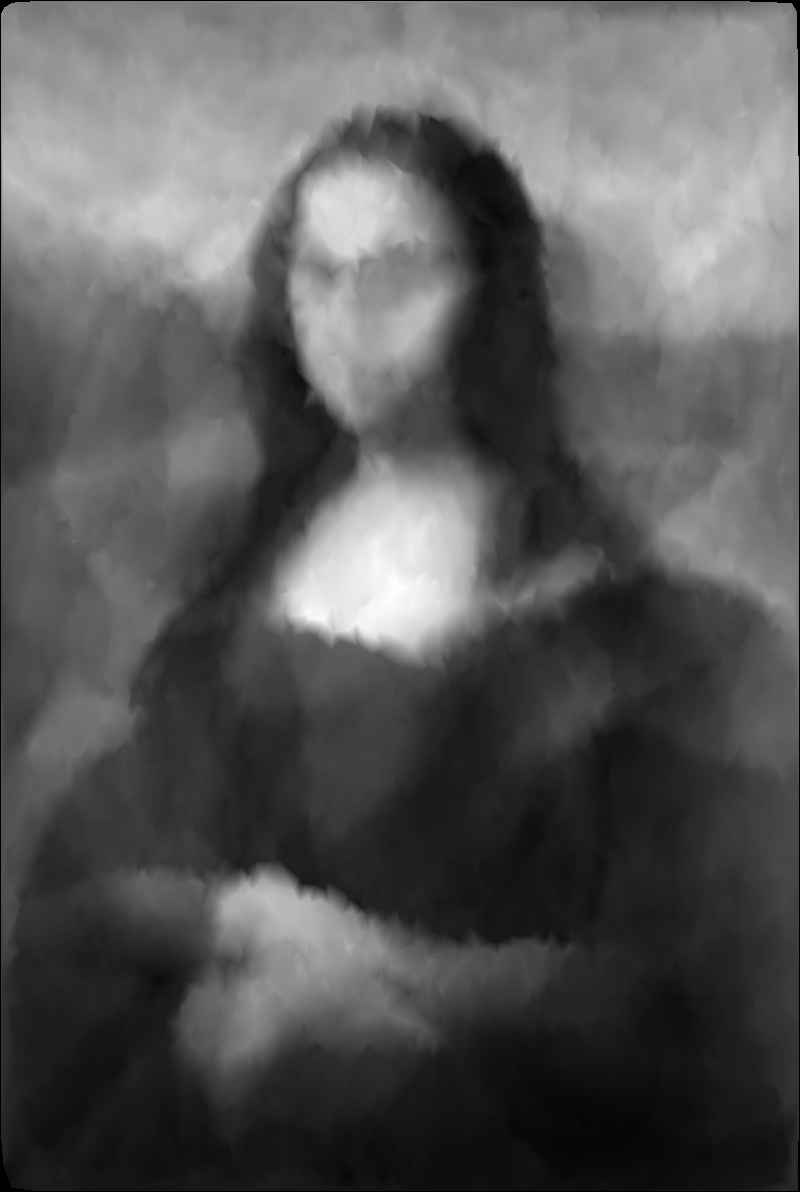}
			\includegraphics[width=\width\textwidth]{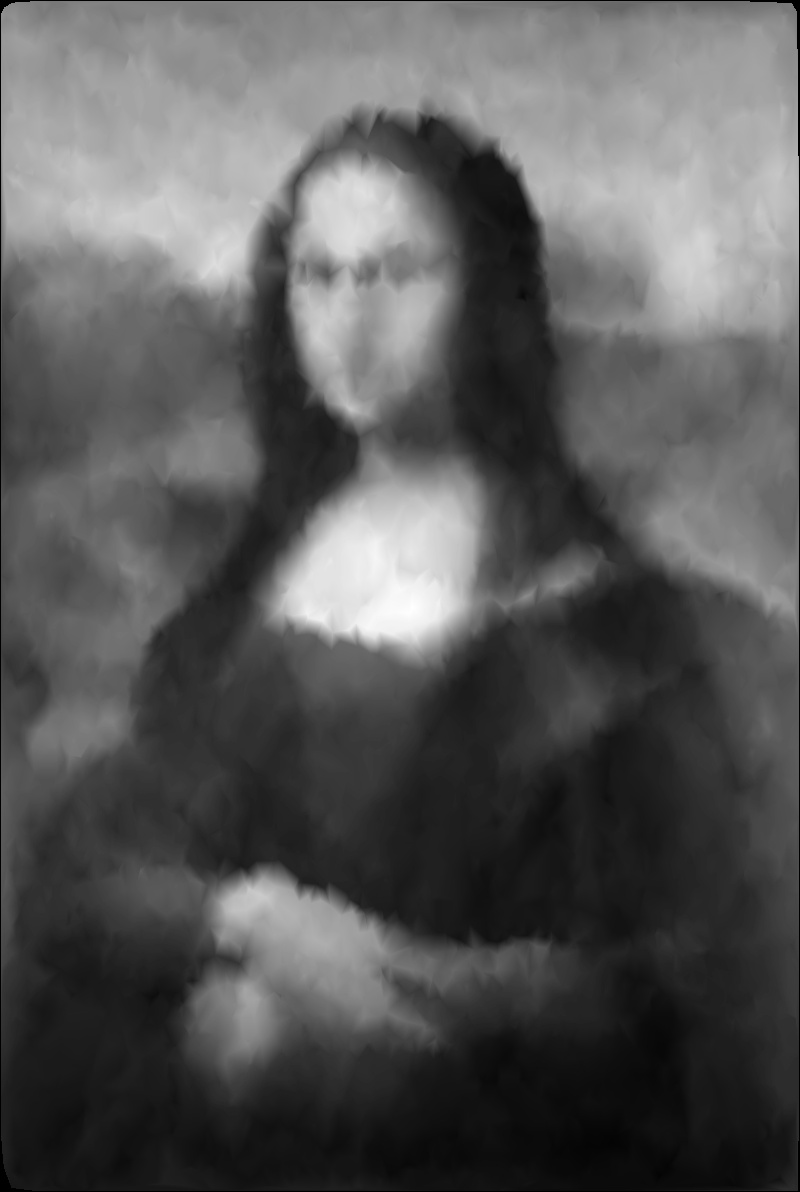}}
		
	\end{tabular}

	\protect
	\caption{(a) Spectral projections of the residual $\bar{m}_{t}$, $<\bar{\upsilon}_{i}^t, \bar{m}_{t}>^2$, at $t = 20000$ and $t = 600000$; (b) and (c) Fourier Transform of $\bar{m}_{t}$ at $t = 20000$ and $t = 600000$ respectively. The high frequency is observed to be dominant in (c). (d) a linear combination $\bar{f}_{t, k} \triangleq \sum_{i = 1}^{k}
		<\bar{\upsilon}_{i}^t, \bar{f}_{t}>
		\bar{\upsilon}_{i}^t
		$ of first $k = \{ 10, 100, 200, 500 \}$ eigenvectors at $t = 600000$. Each vector $\bar{f}_{t, k}$ was interpolated from training points $\{ X^i \}_{i = 1}^{N}$ to entire $[0, 1]^2$ via a linear interpolation.
	}
	\label{fig:NNSpectrum3}
\end{figure}

Further, a projection of the residual $\bar{m}_{t}$ onto \topp eigenvectors, shown in Figure \ref{fig:NNSpectrum1.0-c}, is decreasing along $t$, supporting Eq.~(\ref{eq:L2DynamicsM}). Particularly, we can see that at $t = 600000$ only 10\% of $\bar{m}_{t}$'s energy is located inside \topp 4000 eigenvectors, and thus at the optimization end 90\% of its energy is inside \btm eigenvectors. Moreover, in Figure \ref{fig:NNSpectrum3-a} we can observe that the projection of $\bar{m}_{t}$ along \btm 5000 eigenvectors almost does not change during the entire optimization. This may be caused by two main reasons - the slow convergence rate associated with \btm eigenvectors and a single-precision floating-point (float32) format used in our simulation. The latter can prevent the information flow along the \btm spectrum due to the numerical precision limit. No matter the case, we empirically observe that the information located in the \btm spectrum of $G_t$ was not learned, even for a relatively long optimization process (i.e. $600000$ iterations). Furthermore, since this spectrum part is also associated with high-frequency information \cite{Basri19arxiv}, $\bar{m}_{t}$ at $t = 600000$ comprises mostly the noise, which is also evident from Figures \ref{fig:NNSpectrum3-b}-\ref{fig:NNSpectrum3-c}.

\begin{figure}
	\centering
	
	\newcommand{\width}[0] {0.23}
	
	\begin{tabular}{ccc}

		\subfloat[\label{fig:NNSpectrum1.1-a}]{\includegraphics[width=\width\textwidth]{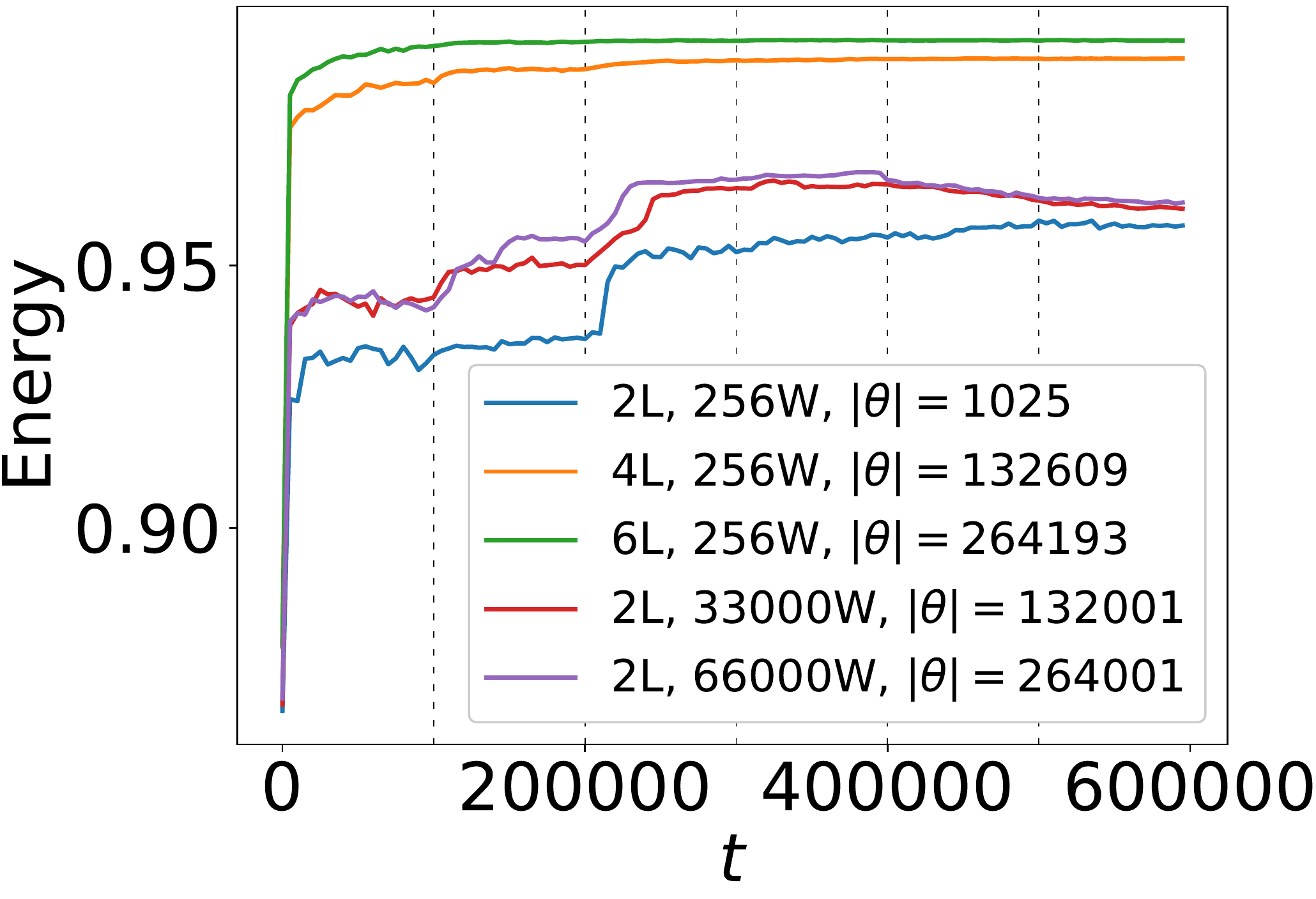}}
		
		&

		\subfloat[\label{fig:NNSpectrum1.1-b}]{\includegraphics[width=\width\textwidth]{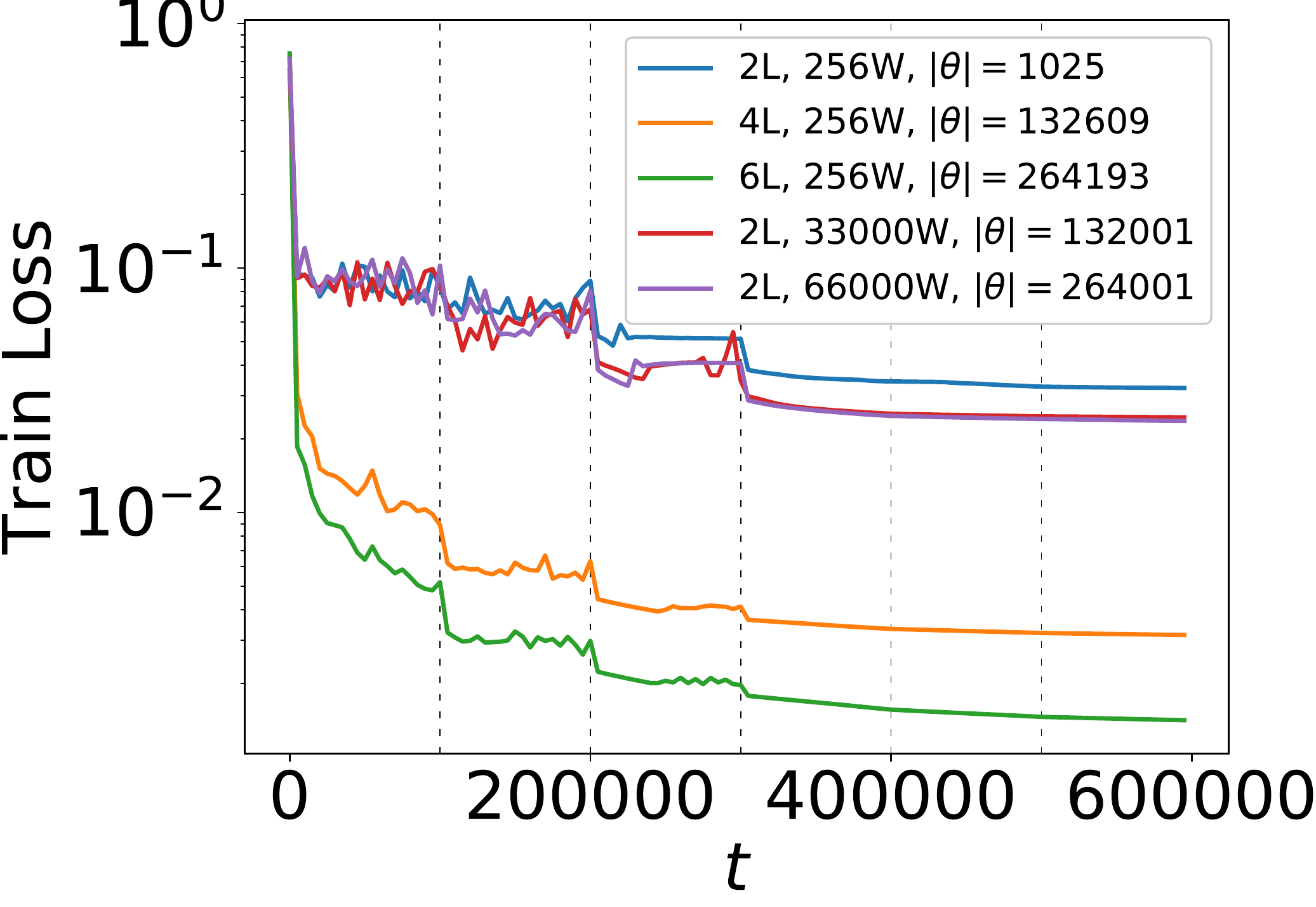}}
		
		&
		
		\subfloat[\label{fig:NNSpectrum1.1-c}]{\includegraphics[width=\width\textwidth]{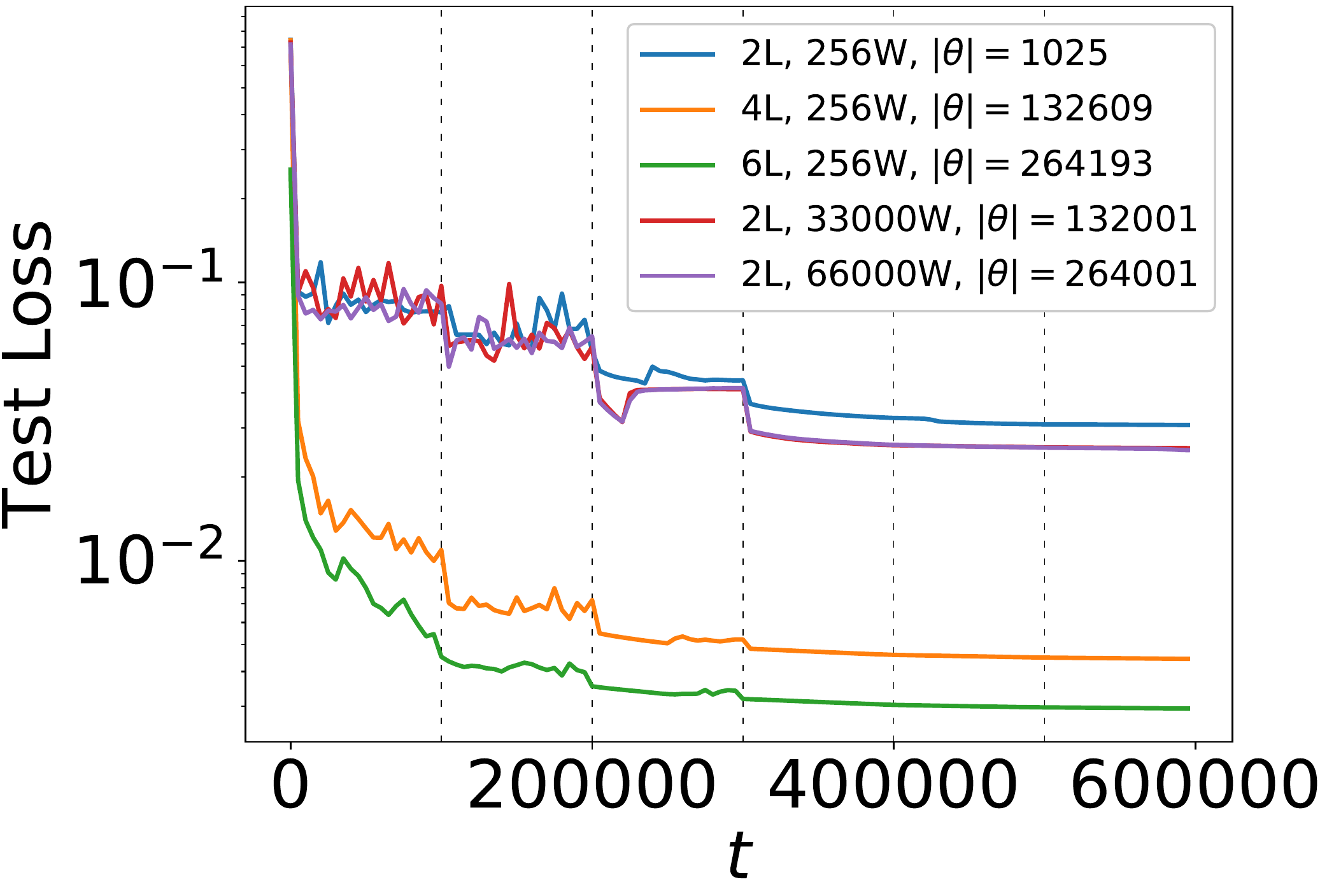}}

	\end{tabular}

	\begin{tabular}{c}
		\subfloat[\label{fig:NNSpectrum1.1-d}]{\includegraphics[width=\width\textwidth]{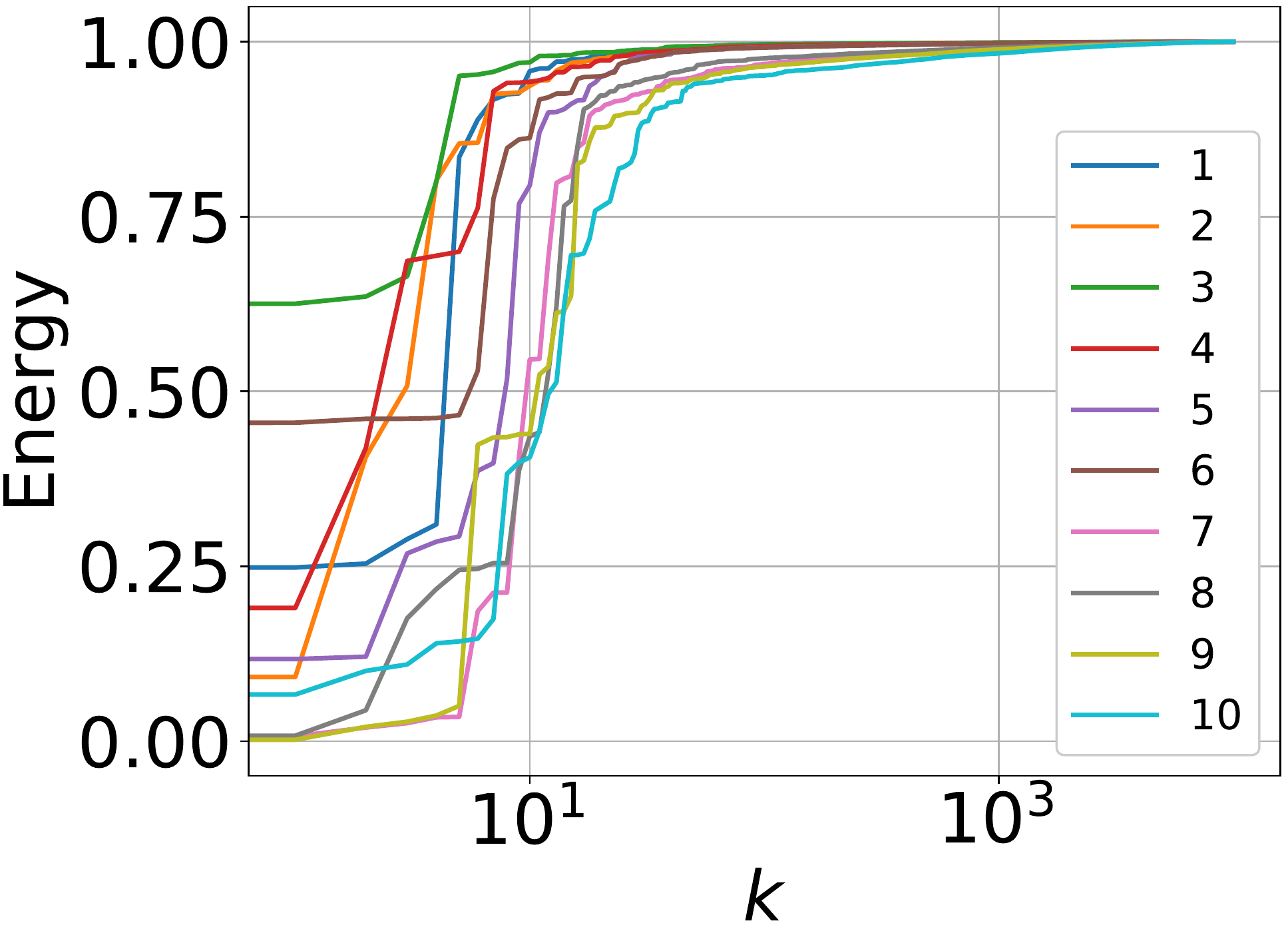}
			\includegraphics[width=\width\textwidth]{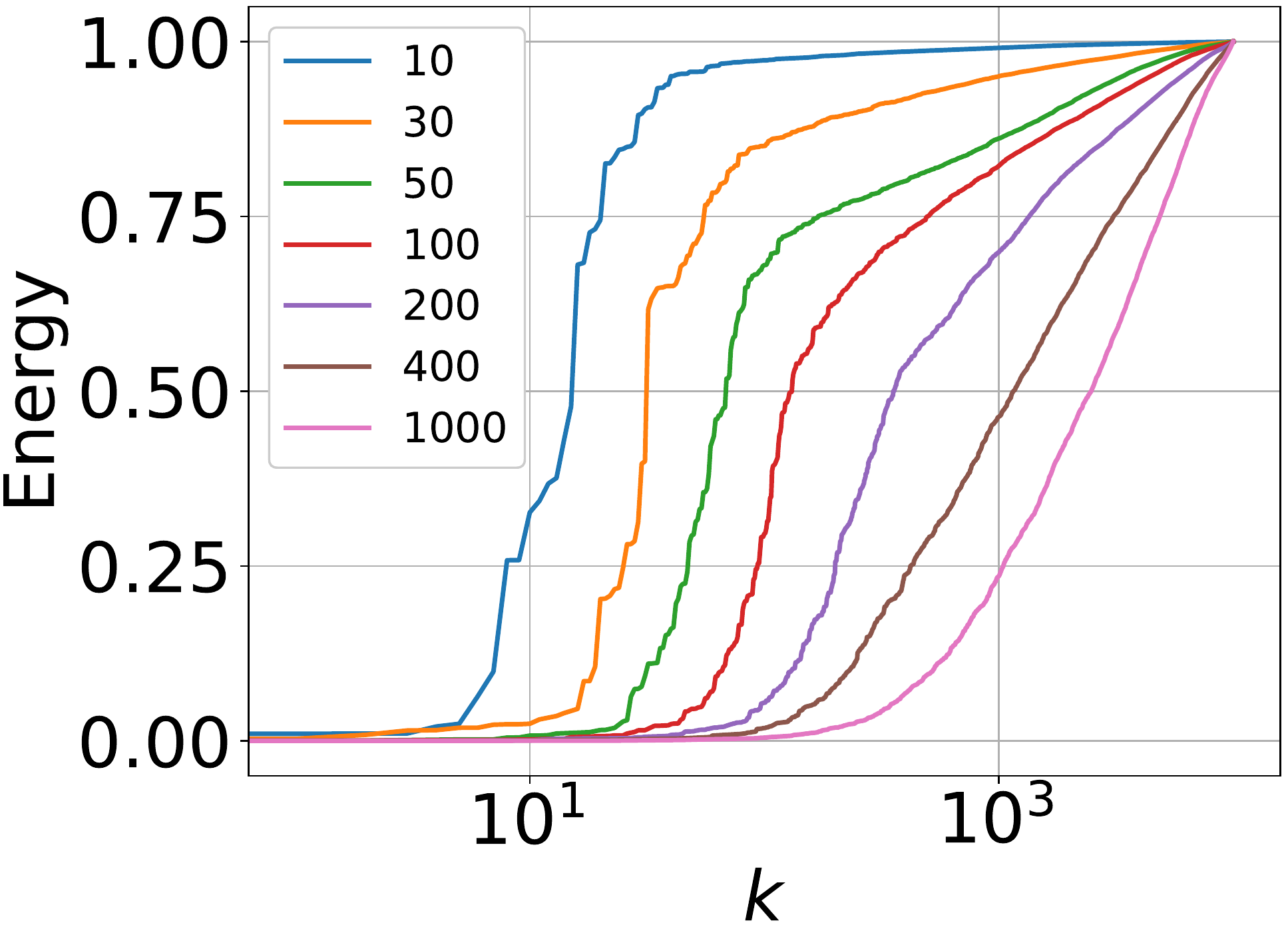}}	
	\end{tabular}

	\protect
	\caption{(a) For NNs with a different number of layers and of neurons, relative energy of the label vector $\bar{y}$ in \topp $400$ eigenvectors of $G_t$, $E_t(\bar{y}, 400)$, along the optimization time $t$; (b) training loss and (c) testing loss of these models. \textbf{L} and \textbf{W} stand for number of layers and number of neurons respectively.
	(d) For different $i$, relative energy of $\bar{\upsilon}_{i}^{600000}$ in spectrum of $G_{20000}$, $E_{20000}(\bar{\upsilon}_{i}^{600000}, k)$, as a function of $k$, with horizontal axes being log-scaled. As seen, 10 first \topp eigenvectors at final time $t = 600000$ are located also in the \topp spectrum of $G_{20000}$, hence the \topp Gramian spectrum was preserved along the optimization. Yet, \btm eigenvectors are significantly less stable.
	}
	\label{fig:NNSpectrum1.1}
\end{figure}

Moreover, we can also observe in Figure \ref{fig:NNSpectrum1.0-c} a special drop of $E_t(\bar{m}_{t}, k)$ at times of $\delta$ decrease. This can be explained by the fact that a lot of $\bar{m}_{t}$'s energy is located inside first several $\{\bar{\upsilon}_{i}^t\}$ (see $E_t(\bar{m}_{t}, 5)$ in Figure \ref{fig:NNSpectrum1.0-c}). When learning rate is decreased, the information flow speed $s_i^t \triangleq 1 - |1 - 
\frac{\delta_t}{N}
\lambda_i^t|$, discussed in Section \ref{sec:L2Loss}, is actually increasing for a few \topp eigenvectors (see Figure \ref{fig:EigvalsEv-c}). That is, terms $\frac{\delta_t}{N}
\lambda_i^t$, being very close to 2 before $\delta$'s decay, are getting close to 1 after, as seen in Figure \ref{fig:EigvalsEv-d}. In its turn this accelerates the information flow along these first $\{\bar{\upsilon}_{i}^t\}$, as described in Eq.~(\ref{eq:L2DynamicsF})-(\ref{eq:L2DynamicsM}), leading also to a special descend of $E_t(\bar{m}_{t}, k)$ and of the training loss in Figure \ref{fig:NNSpectrum1.1-b}. 


\paragraph{Eigenvectors}

\begin{figure}
	\centering\offinterlineskip
	\newcommand{\imagepath}[0] {Figures/FC_NN_6L/time_00595000}

	\includegraphics[width=0.166\linewidth]{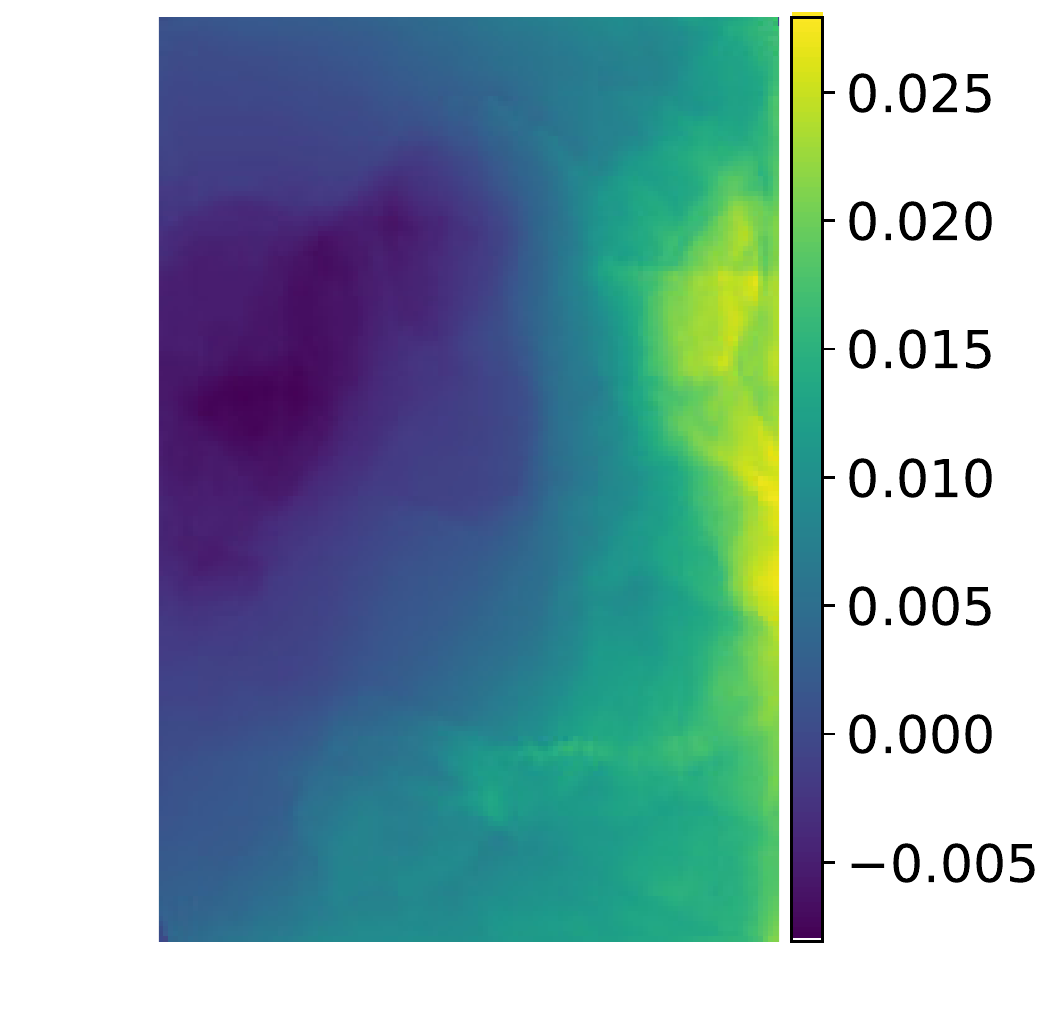}%
	\includegraphics[width=0.166\linewidth]{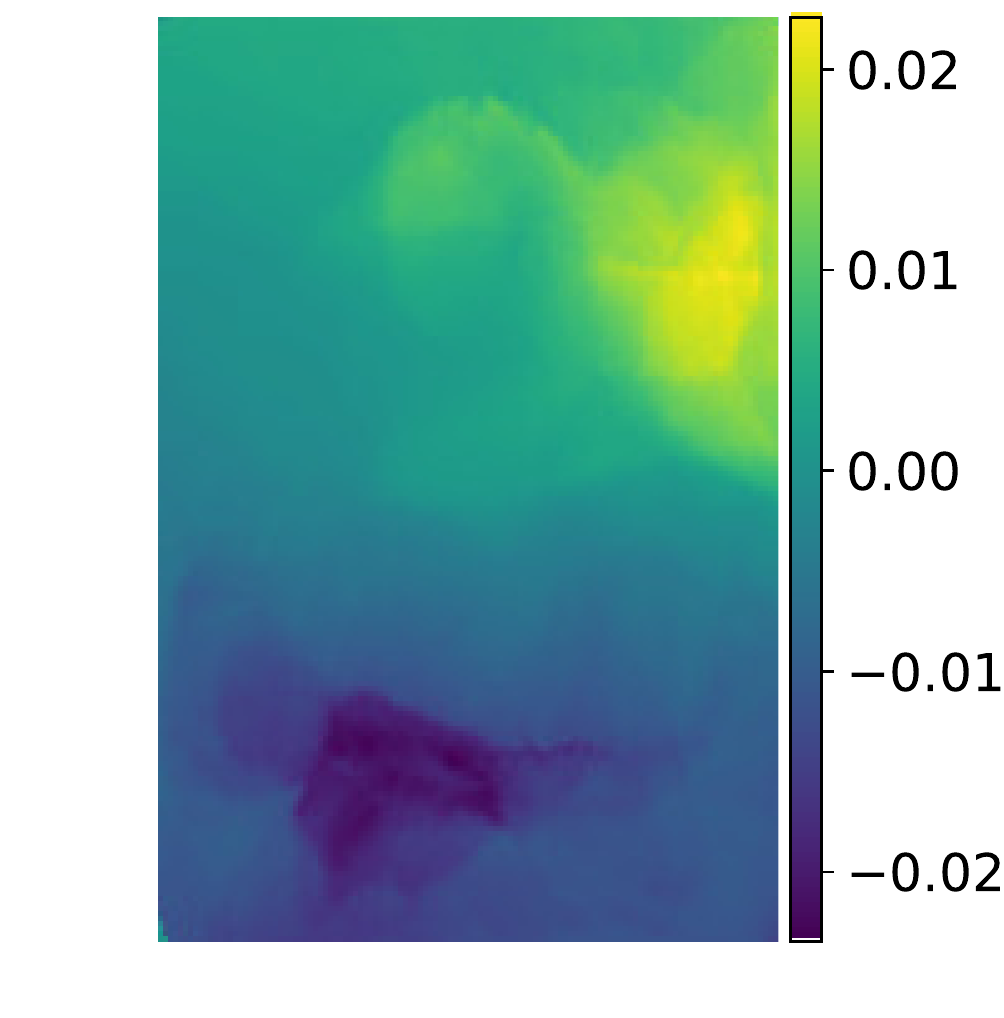}%
	\includegraphics[width=0.166\linewidth]{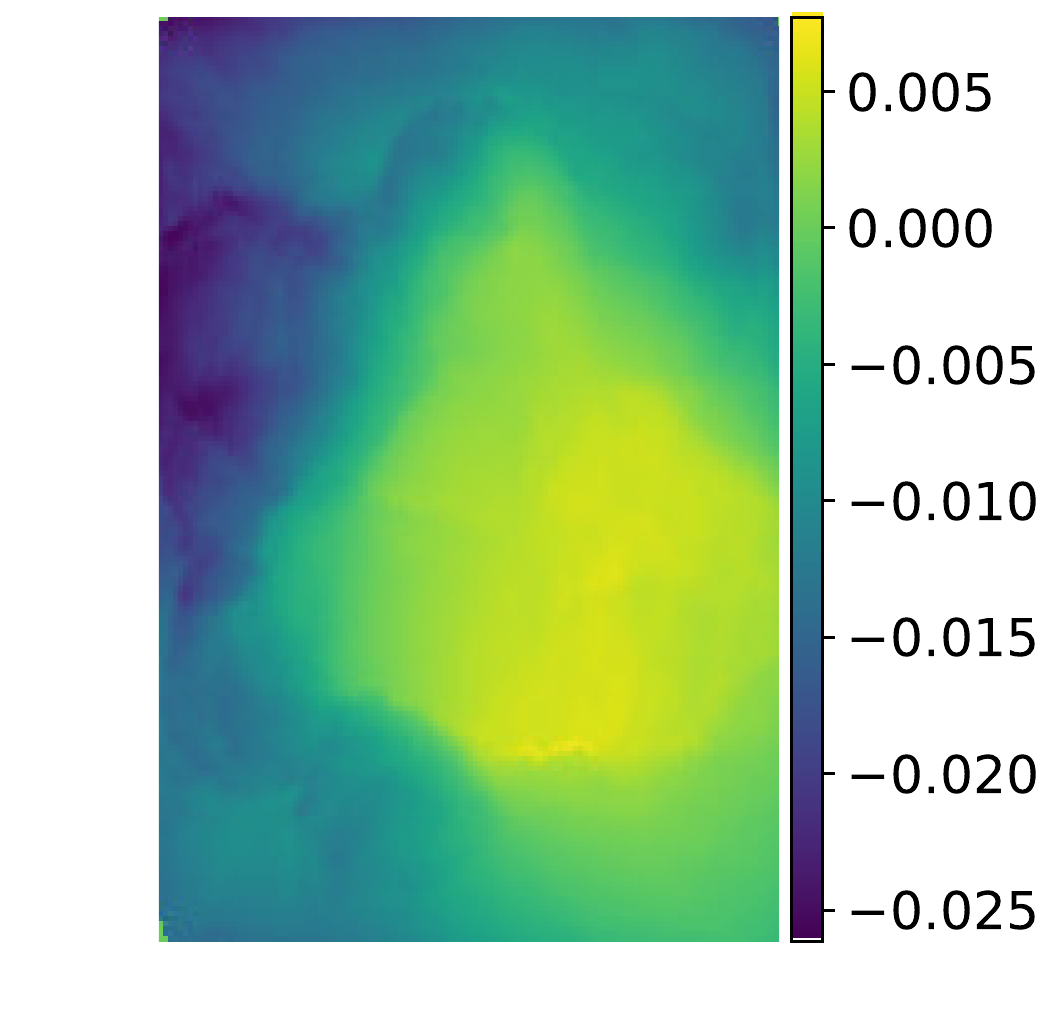}%
	\includegraphics[width=0.166\linewidth]{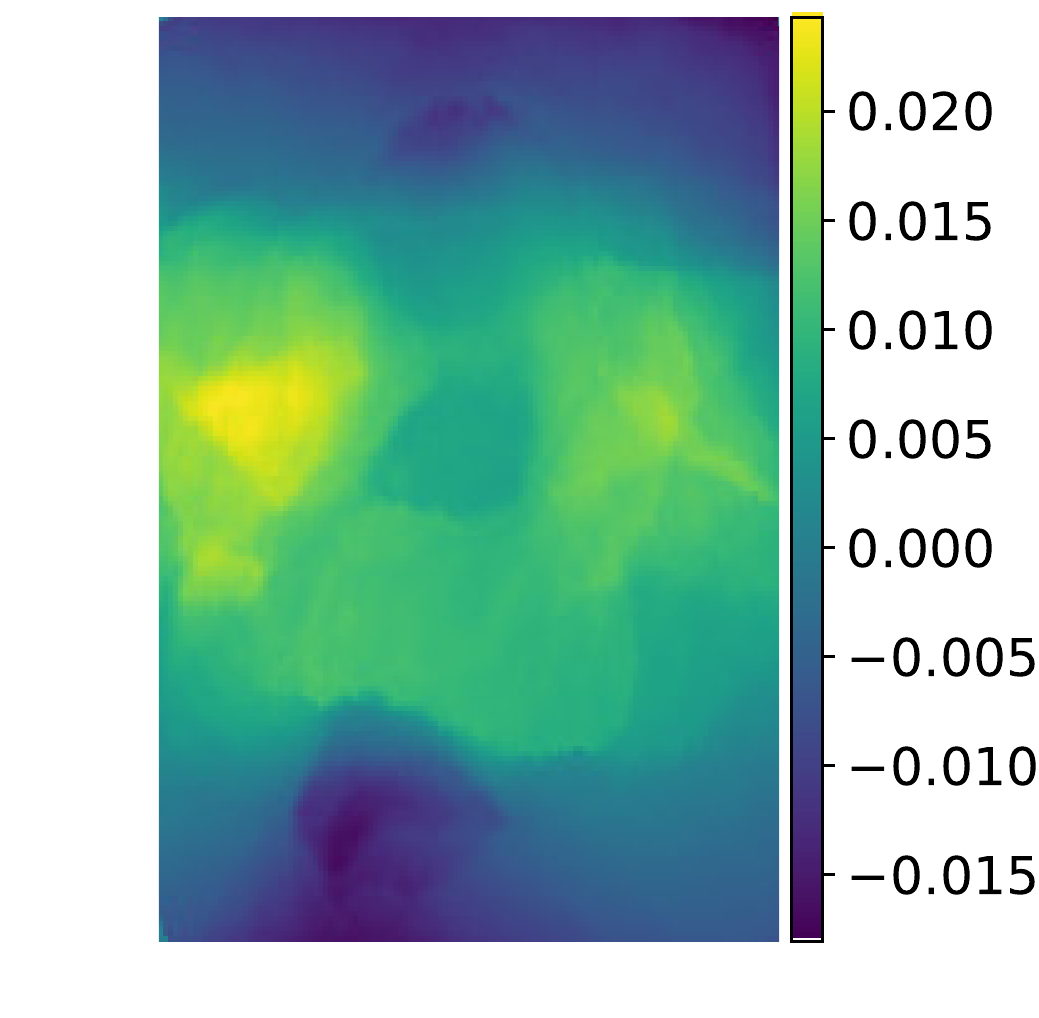}%
	\includegraphics[width=0.166\linewidth]{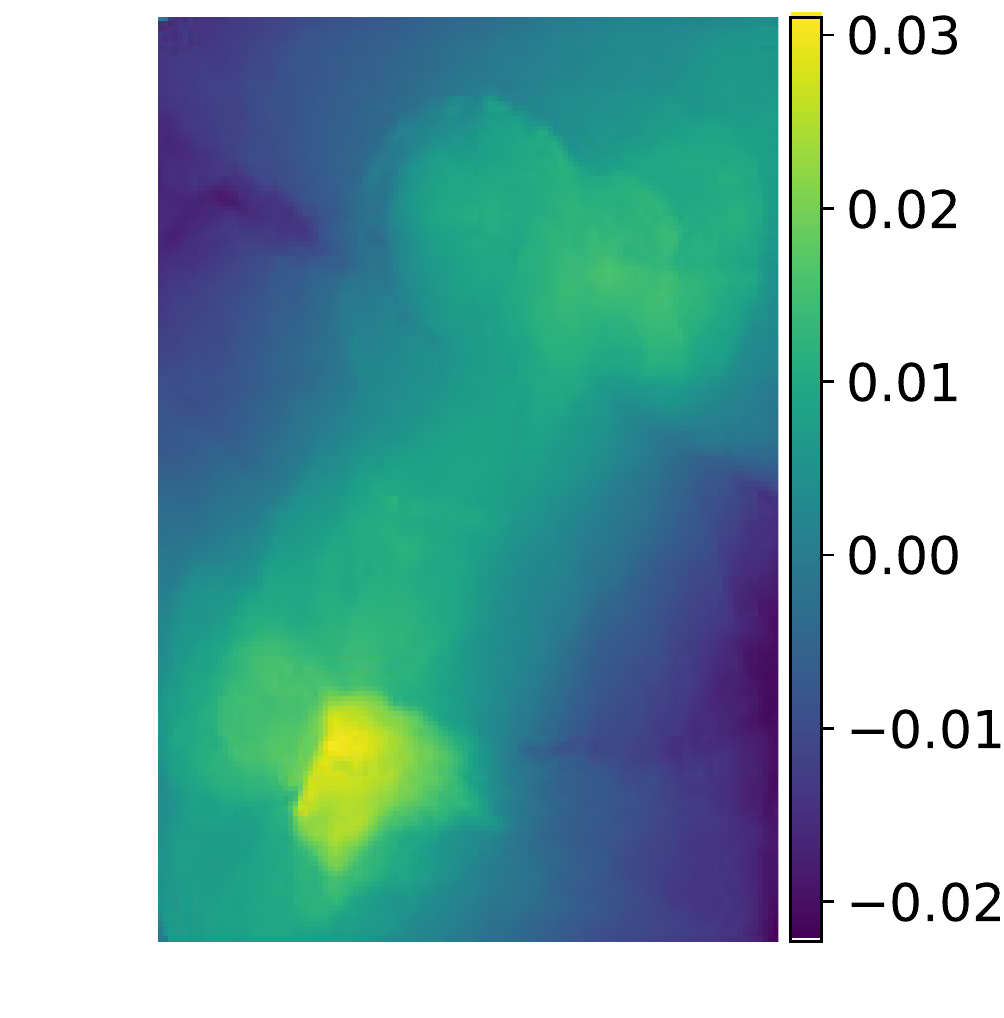}%
	\includegraphics[width=0.166\linewidth]{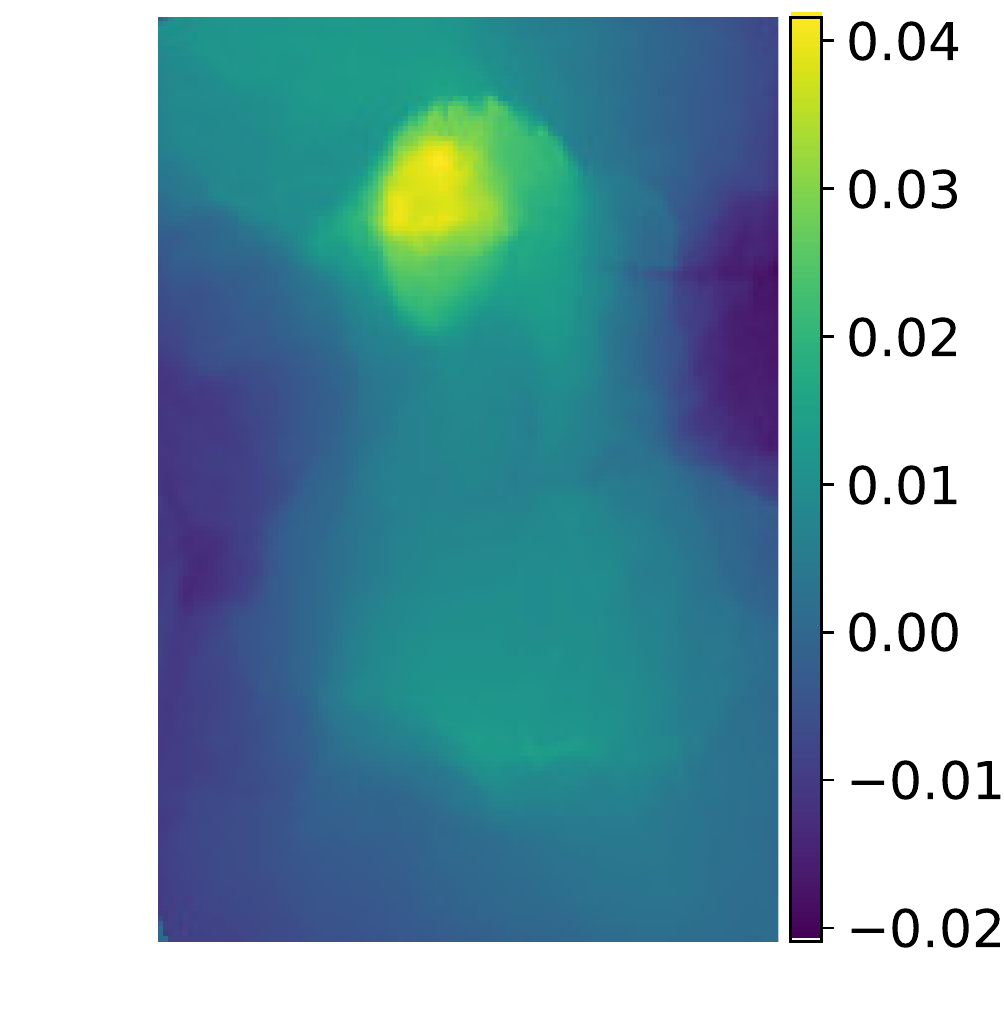}%
	\\
	\includegraphics[width=0.15\linewidth]{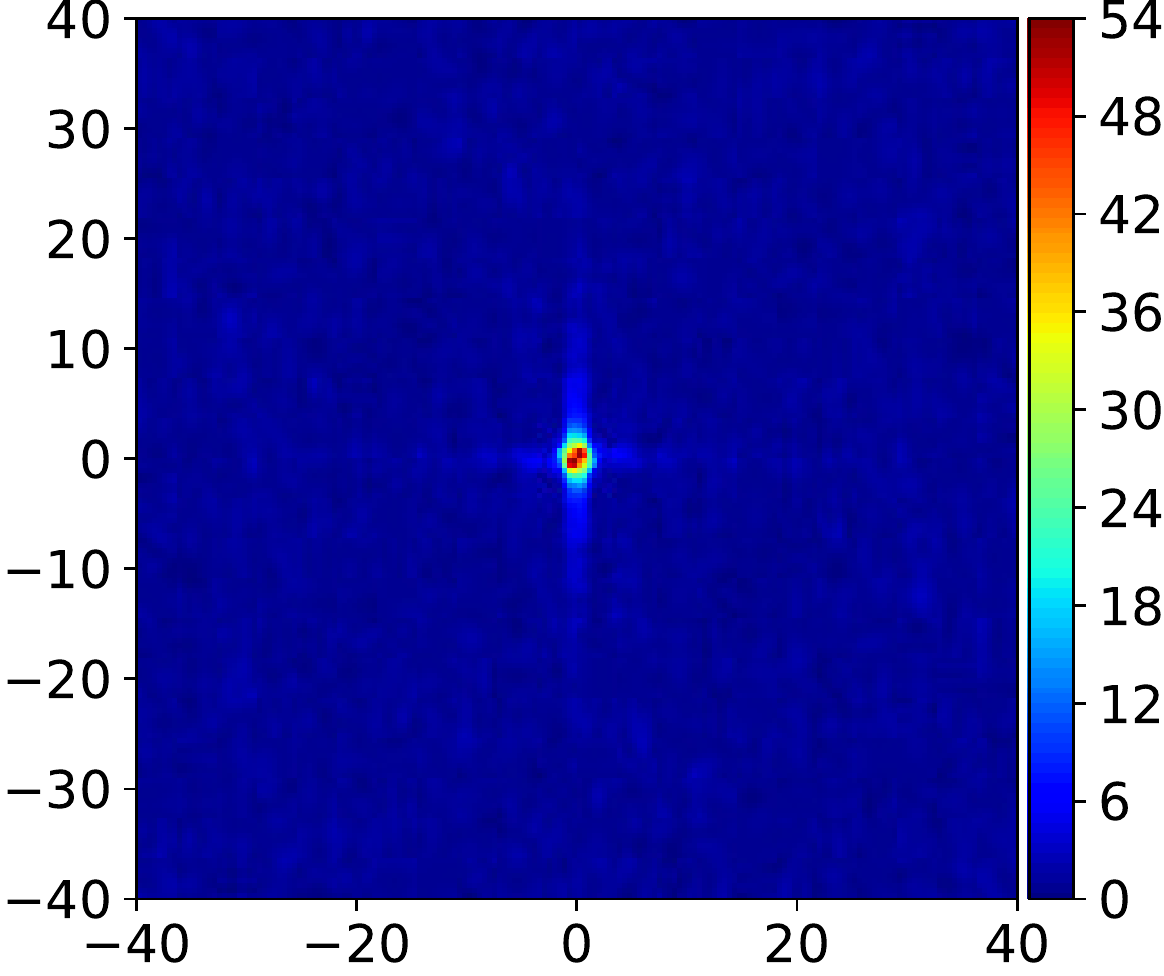}%
	\hspace{0.01\linewidth}
	\includegraphics[width=0.15\linewidth]{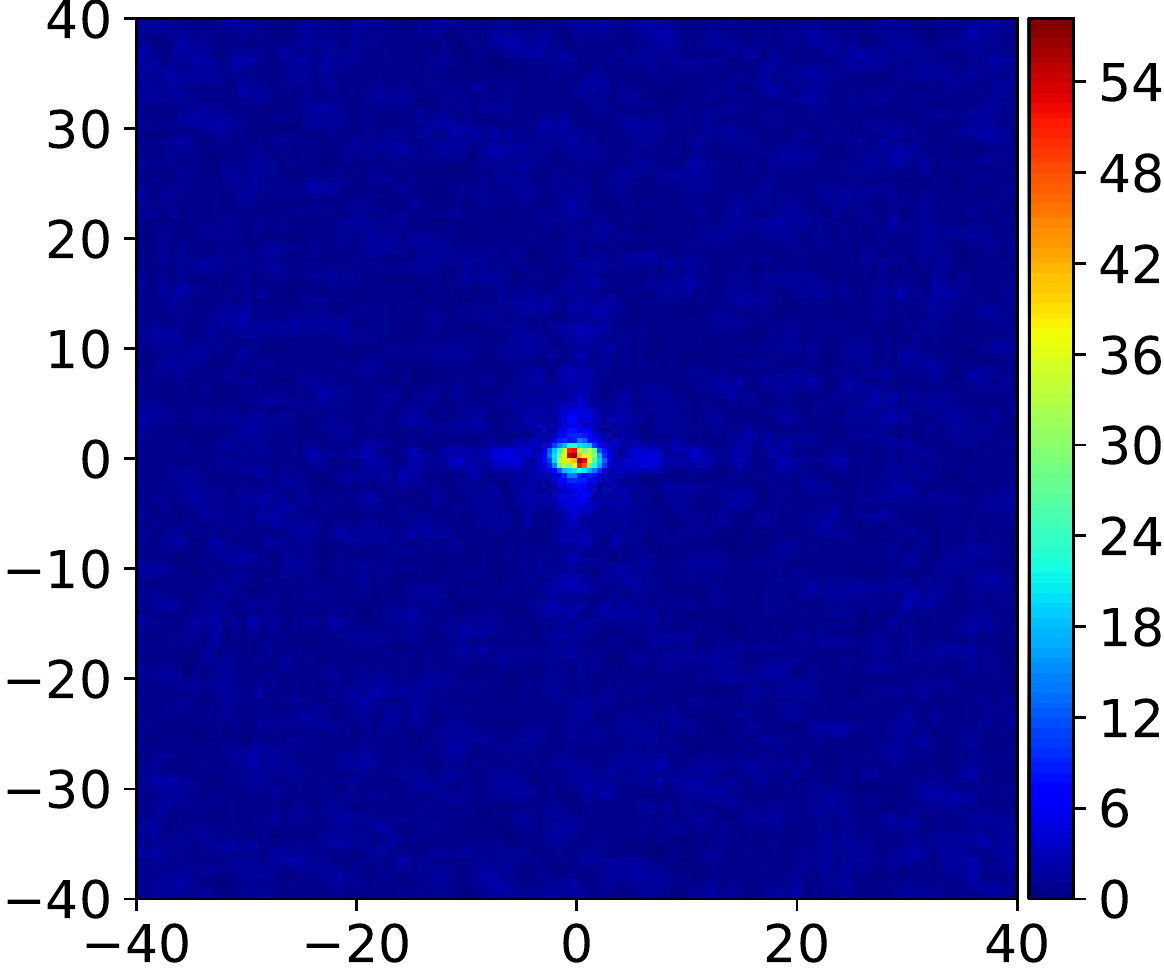}%
	\hspace{0.01\linewidth}
	\includegraphics[width=0.15\linewidth]{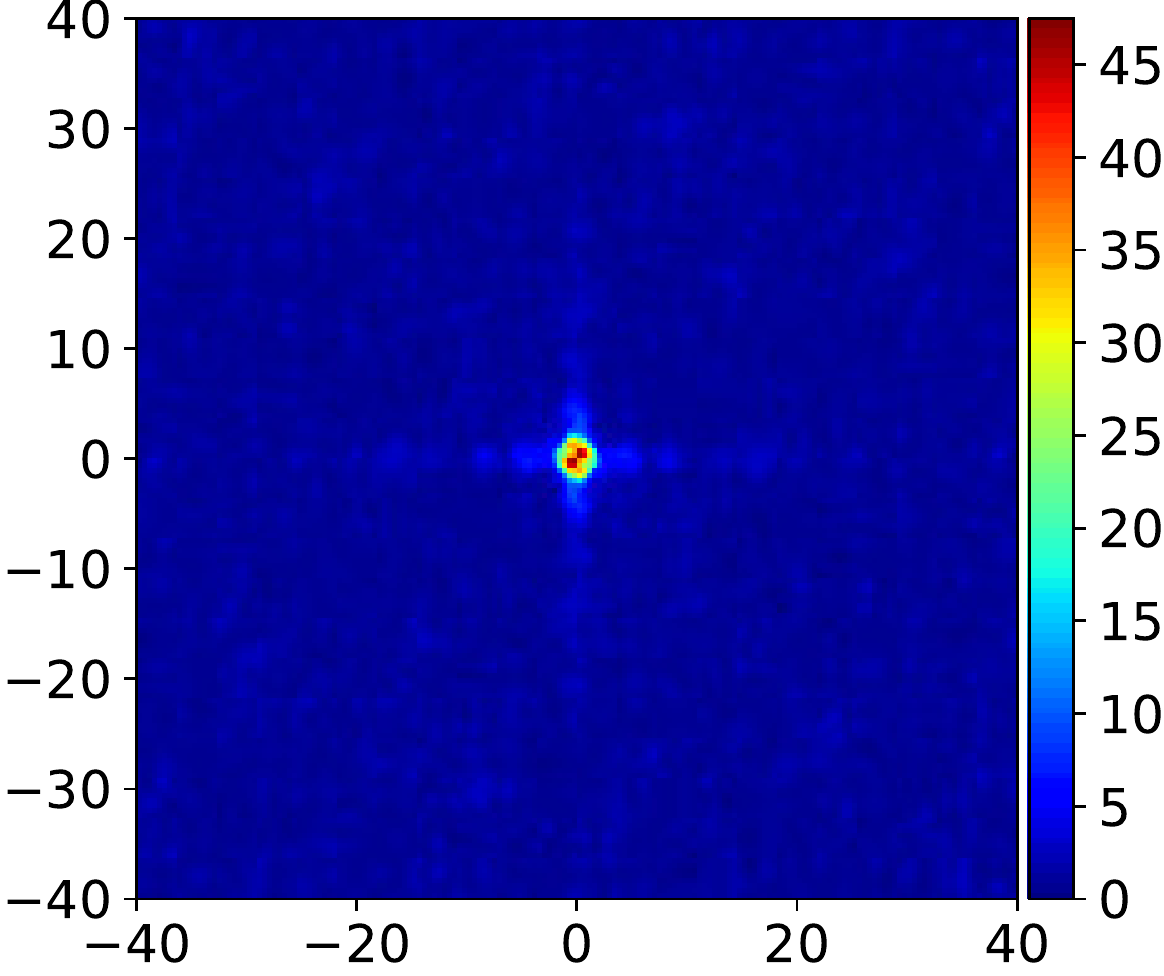}%
	\hspace{0.01\linewidth}
	\includegraphics[width=0.15\linewidth]{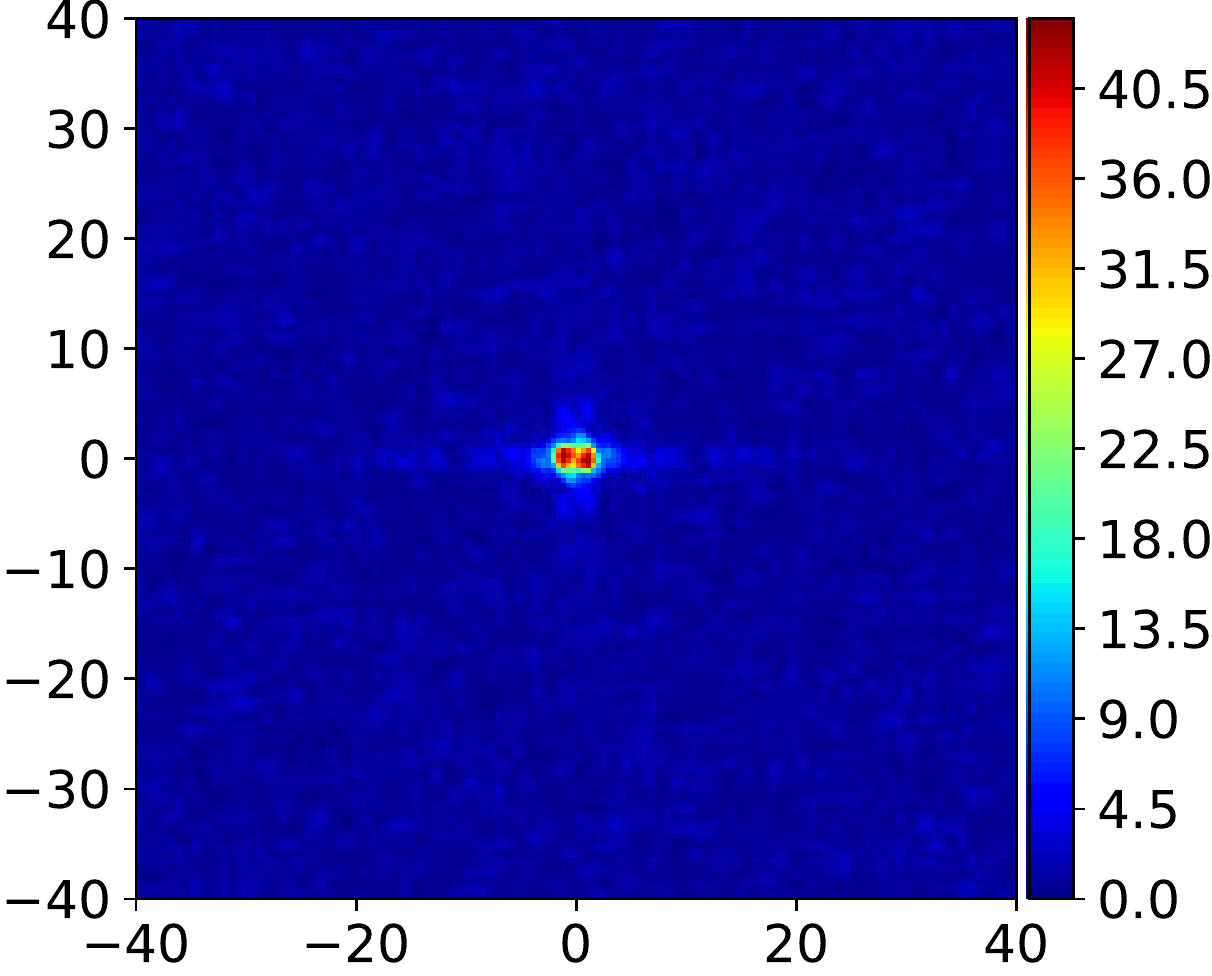}%
	\hspace{0.01\linewidth}
	\includegraphics[width=0.15\linewidth]{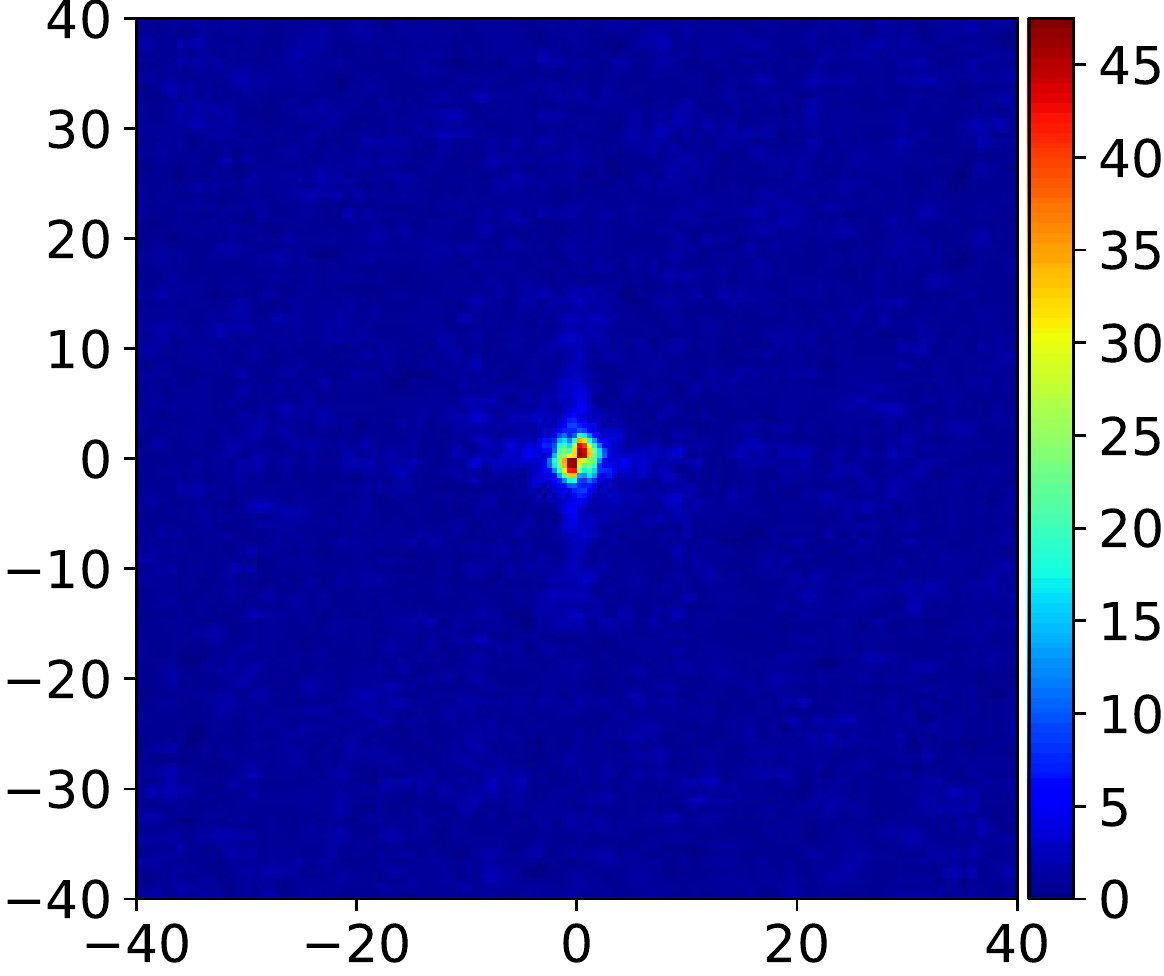}%
	\hspace{0.01\linewidth}
	\includegraphics[width=0.15\linewidth]{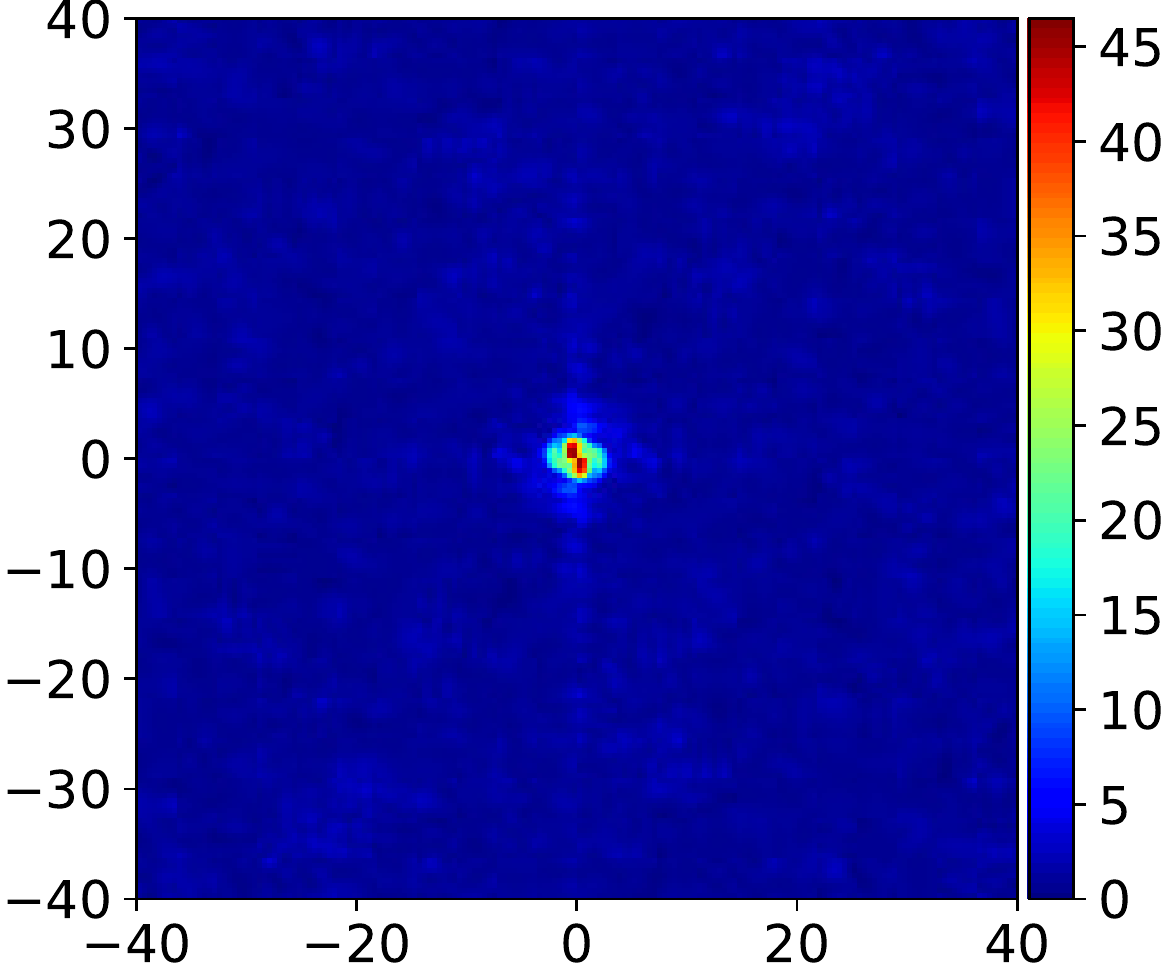}%
	\\
	\includegraphics[width=0.166\linewidth]{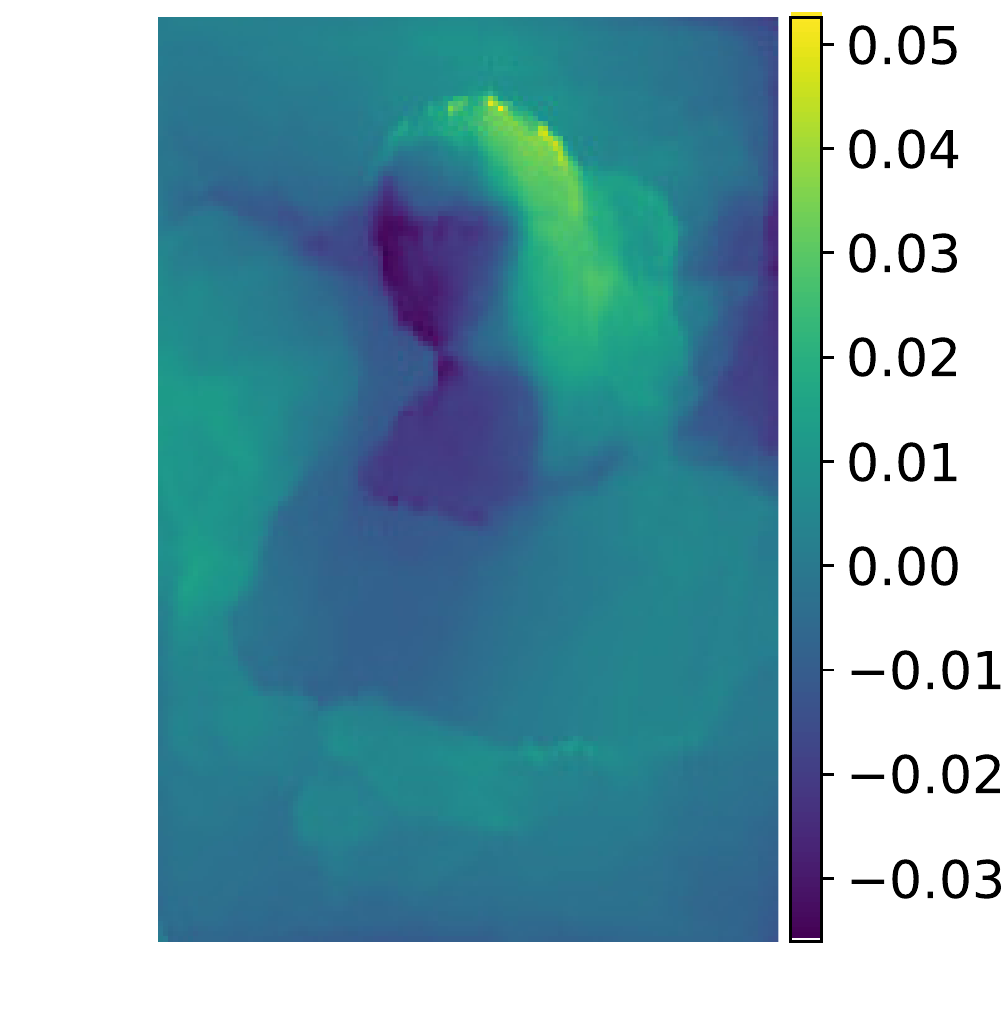}%
	\includegraphics[width=0.166\linewidth]{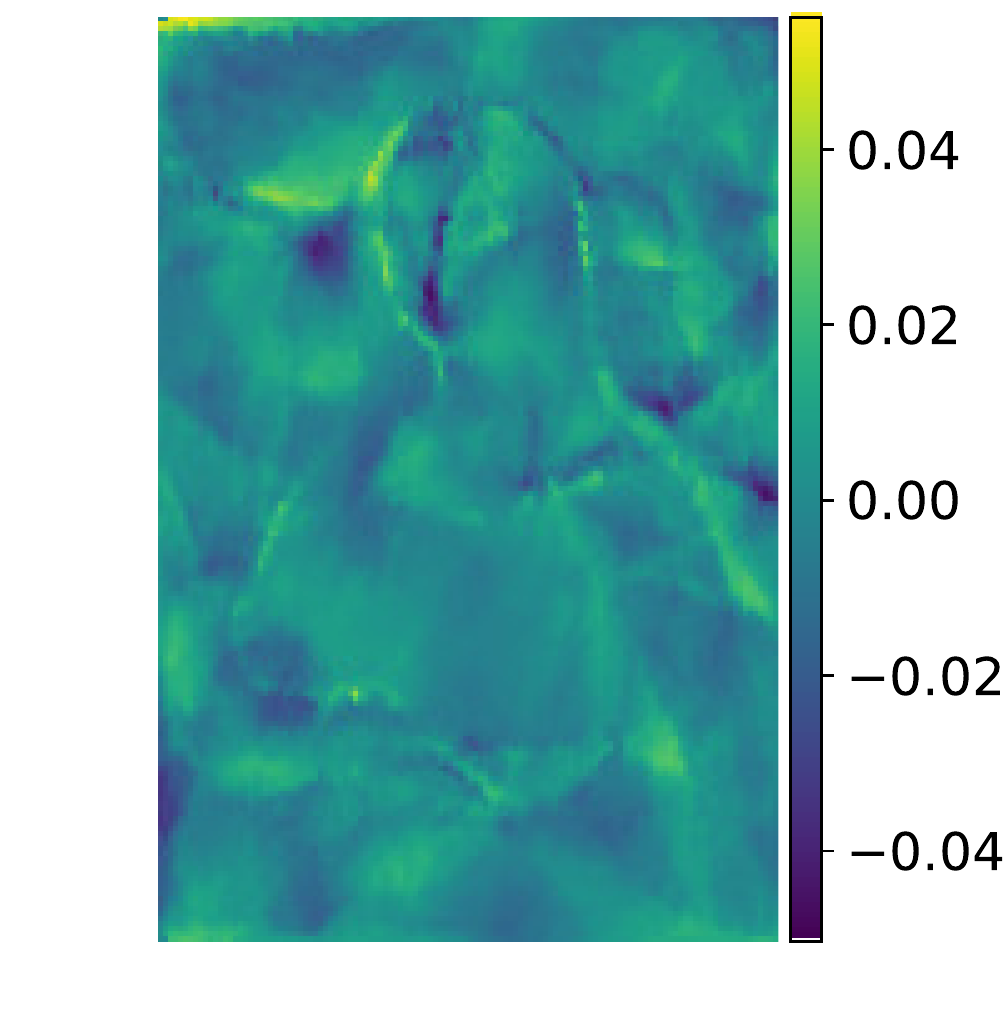}%
	\includegraphics[width=0.166\linewidth]{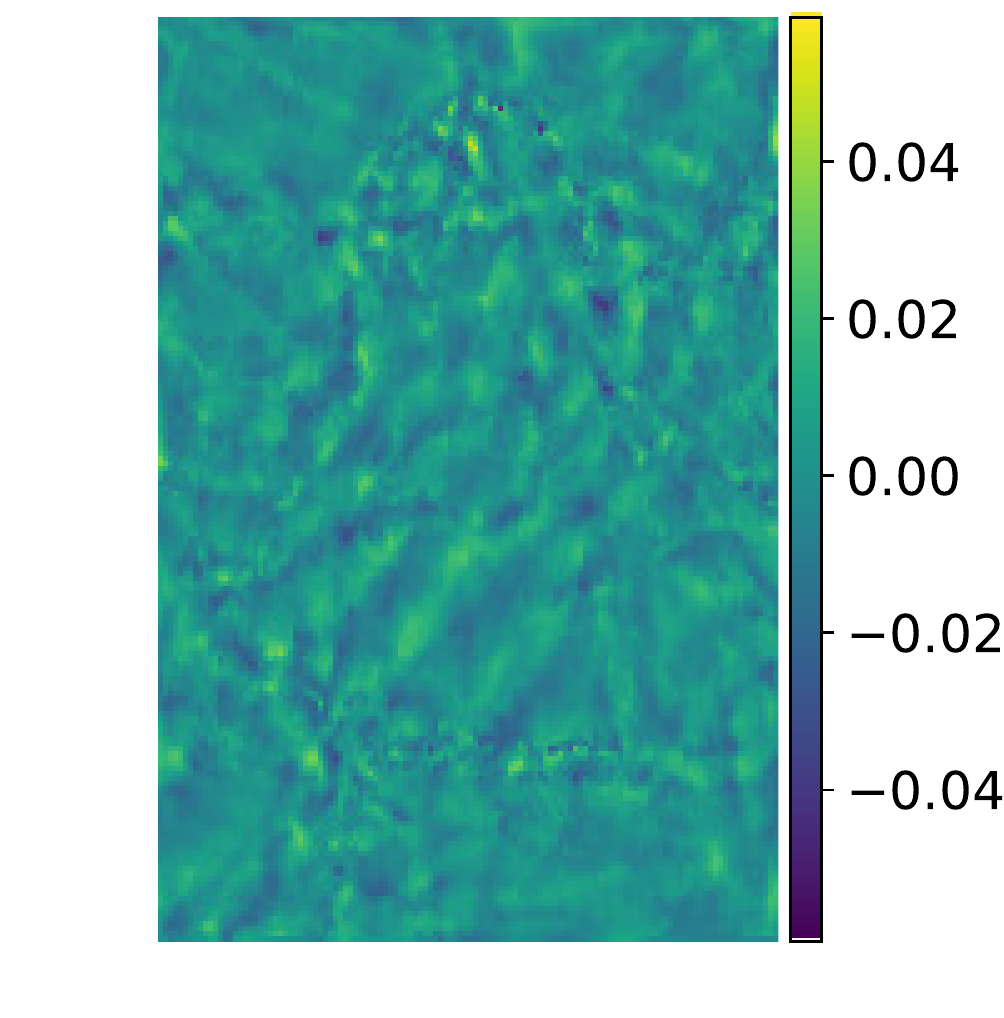}%
	\includegraphics[width=0.166\linewidth]{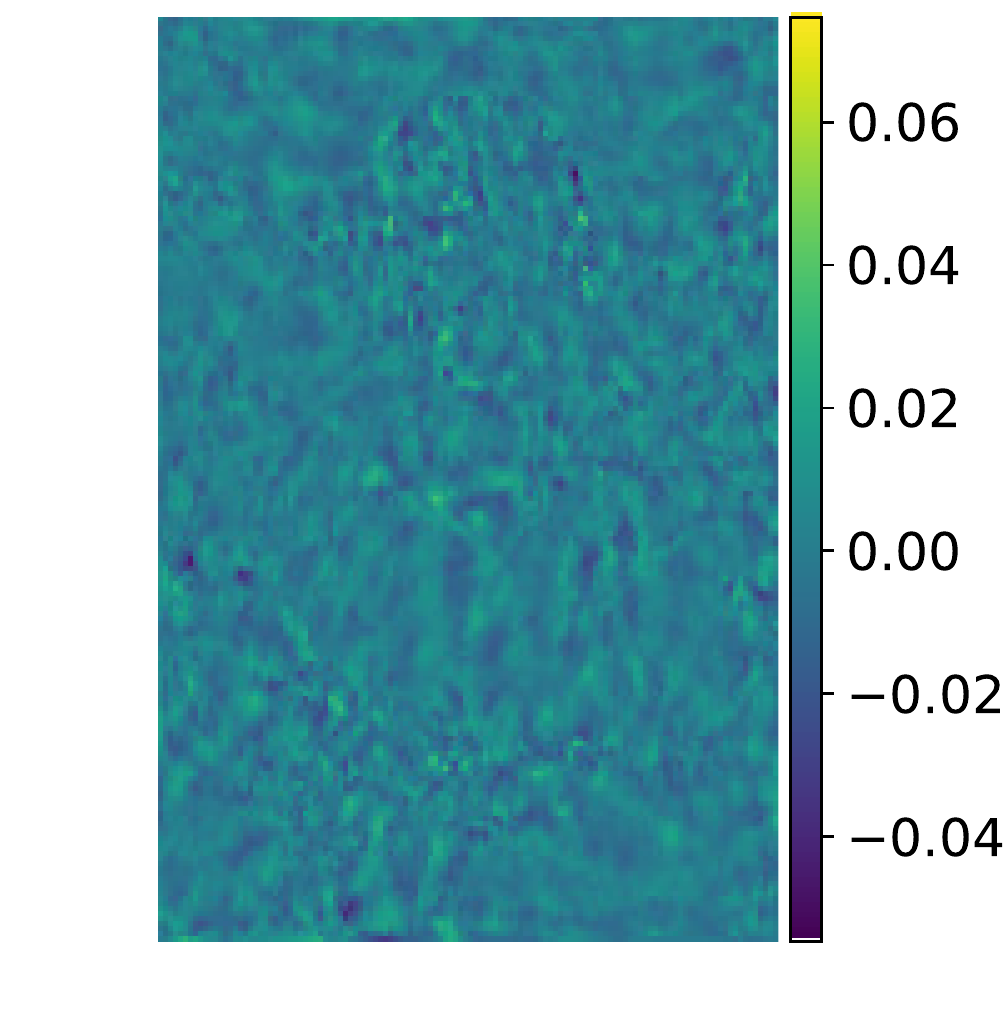}%
	\includegraphics[width=0.166\linewidth]{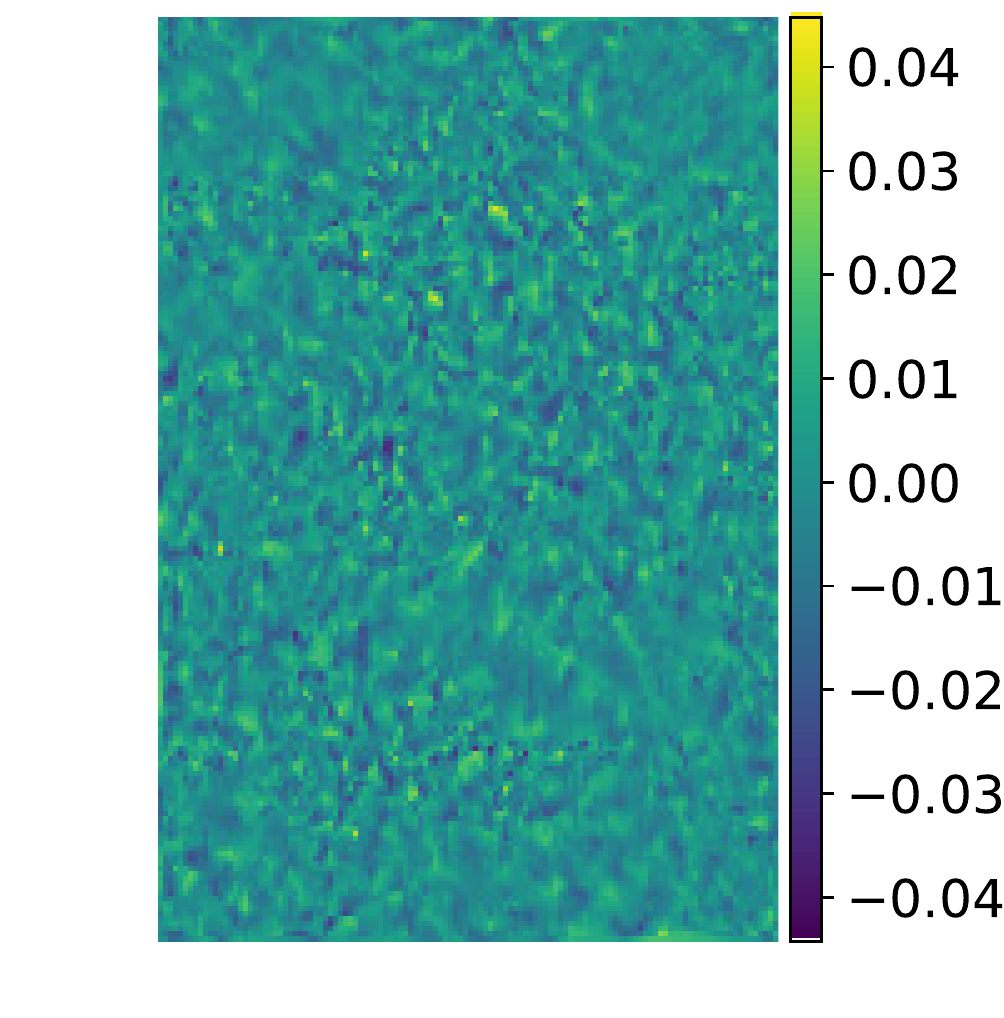}%
	\includegraphics[width=0.166\linewidth]{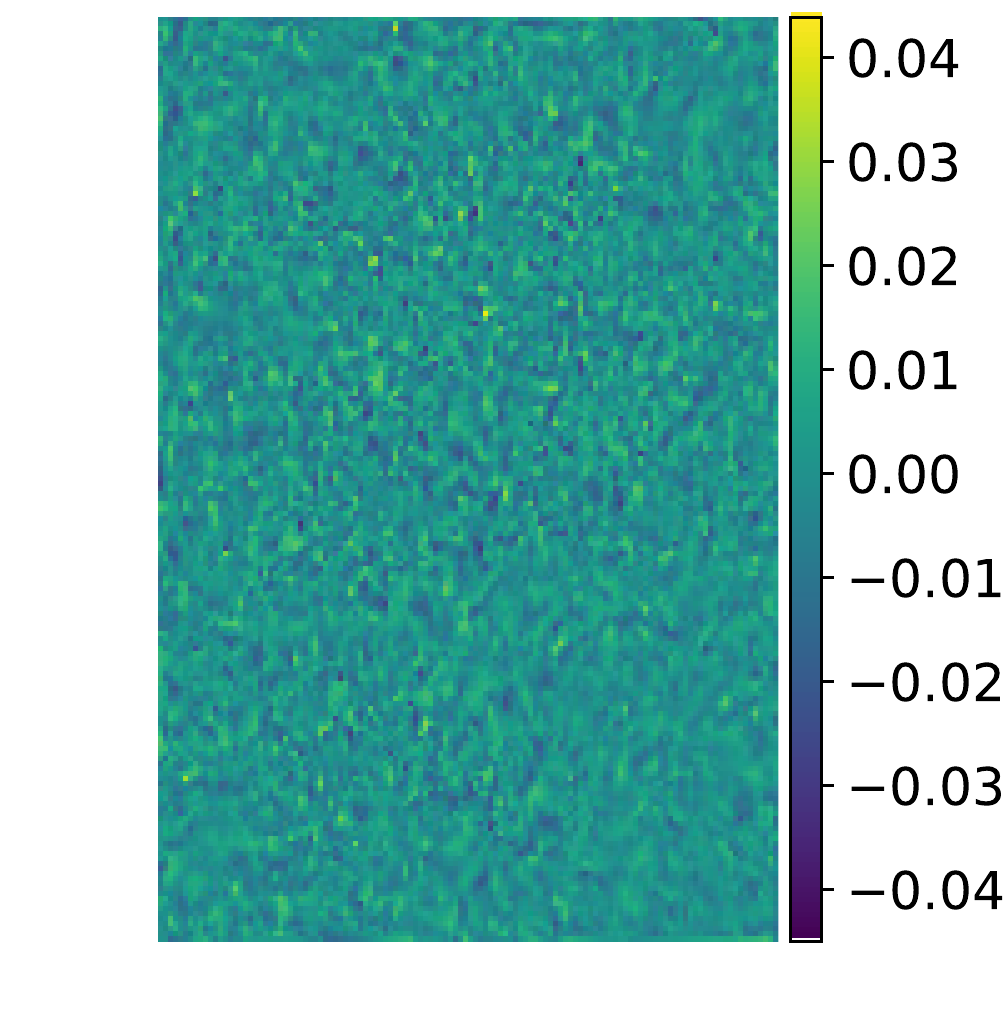}%
	\\
	\includegraphics[width=0.15\linewidth]{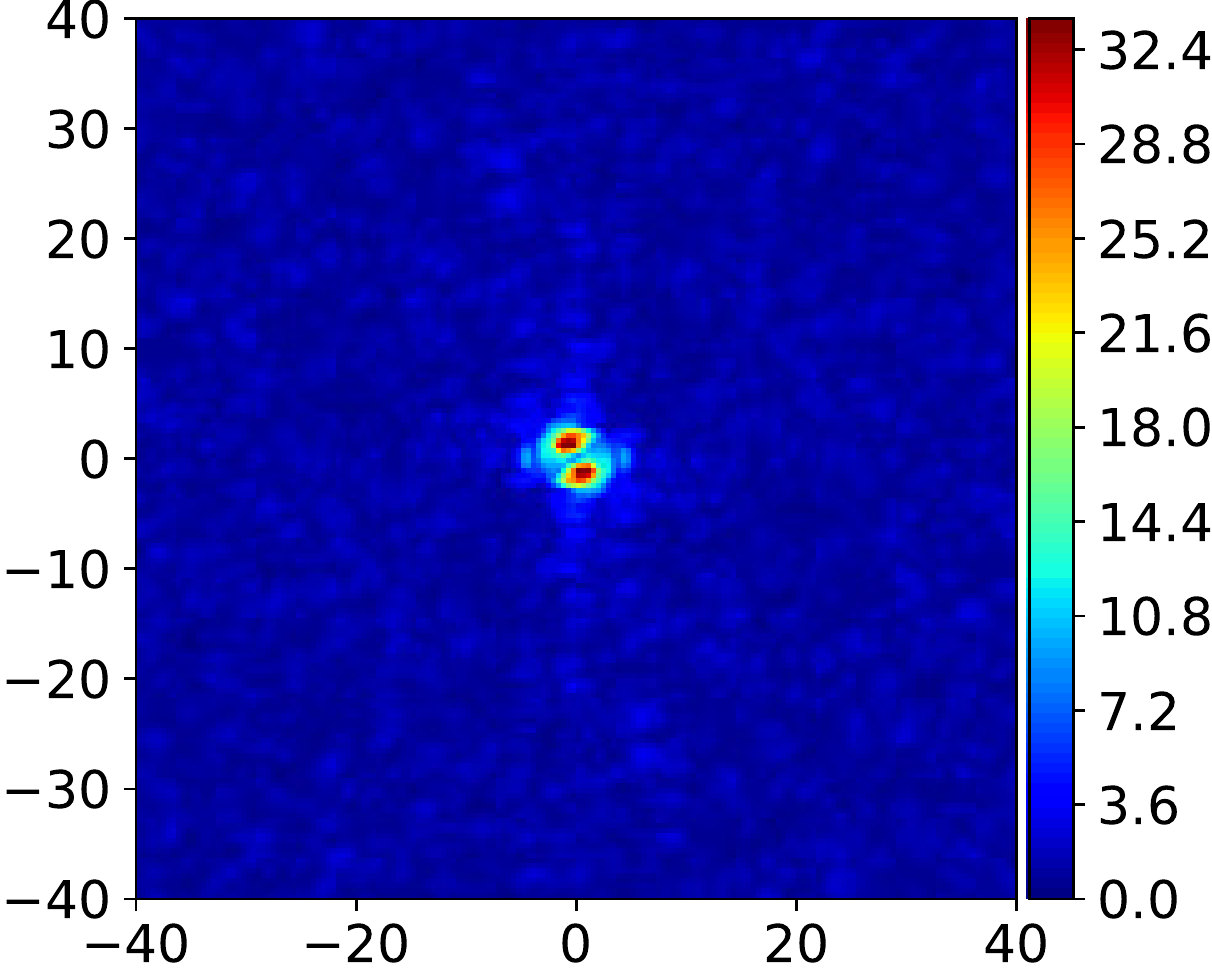}%
	\hspace{0.01\linewidth}
	\includegraphics[width=0.15\linewidth]{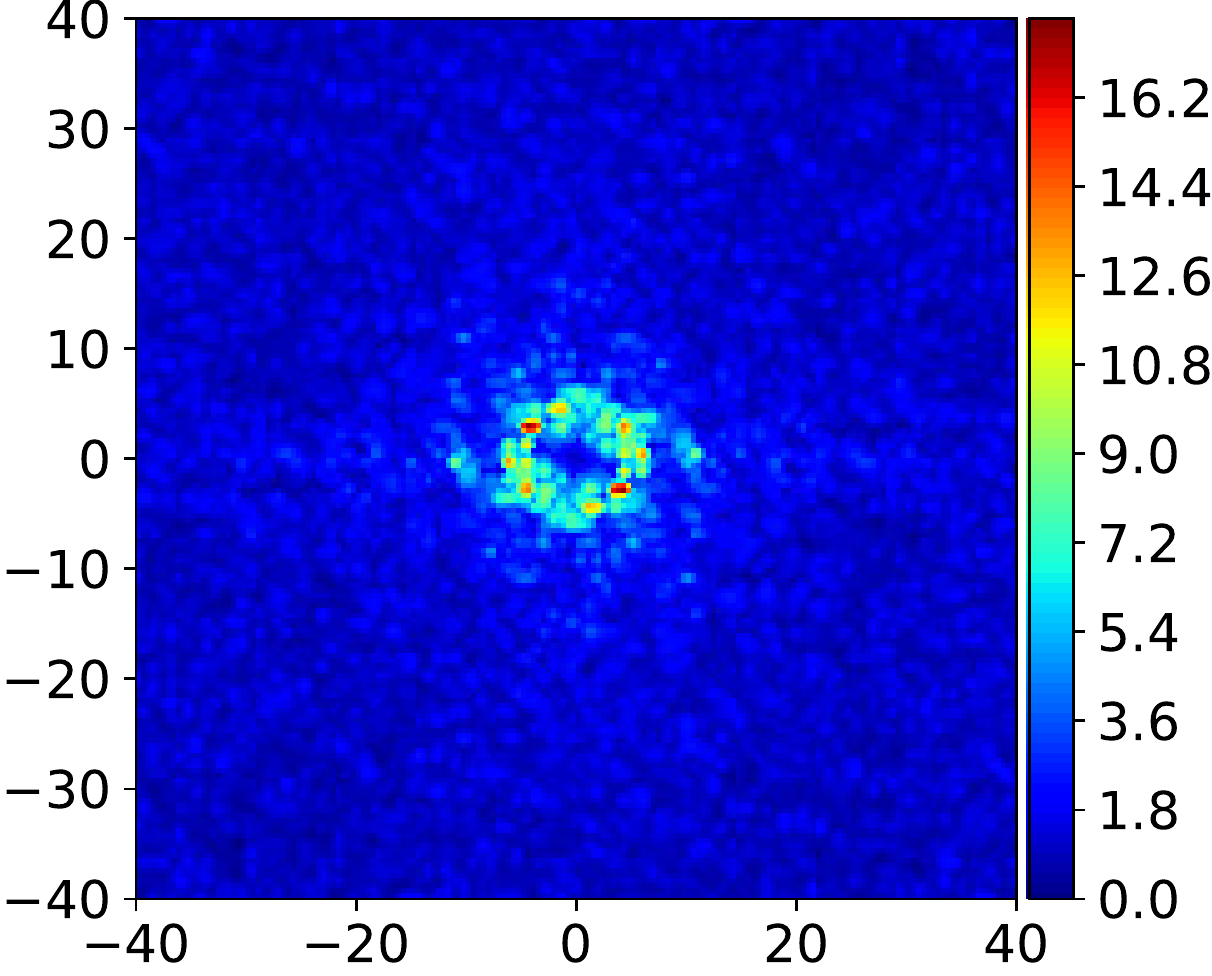}%
	\hspace{0.01\linewidth}
	\includegraphics[width=0.15\linewidth]{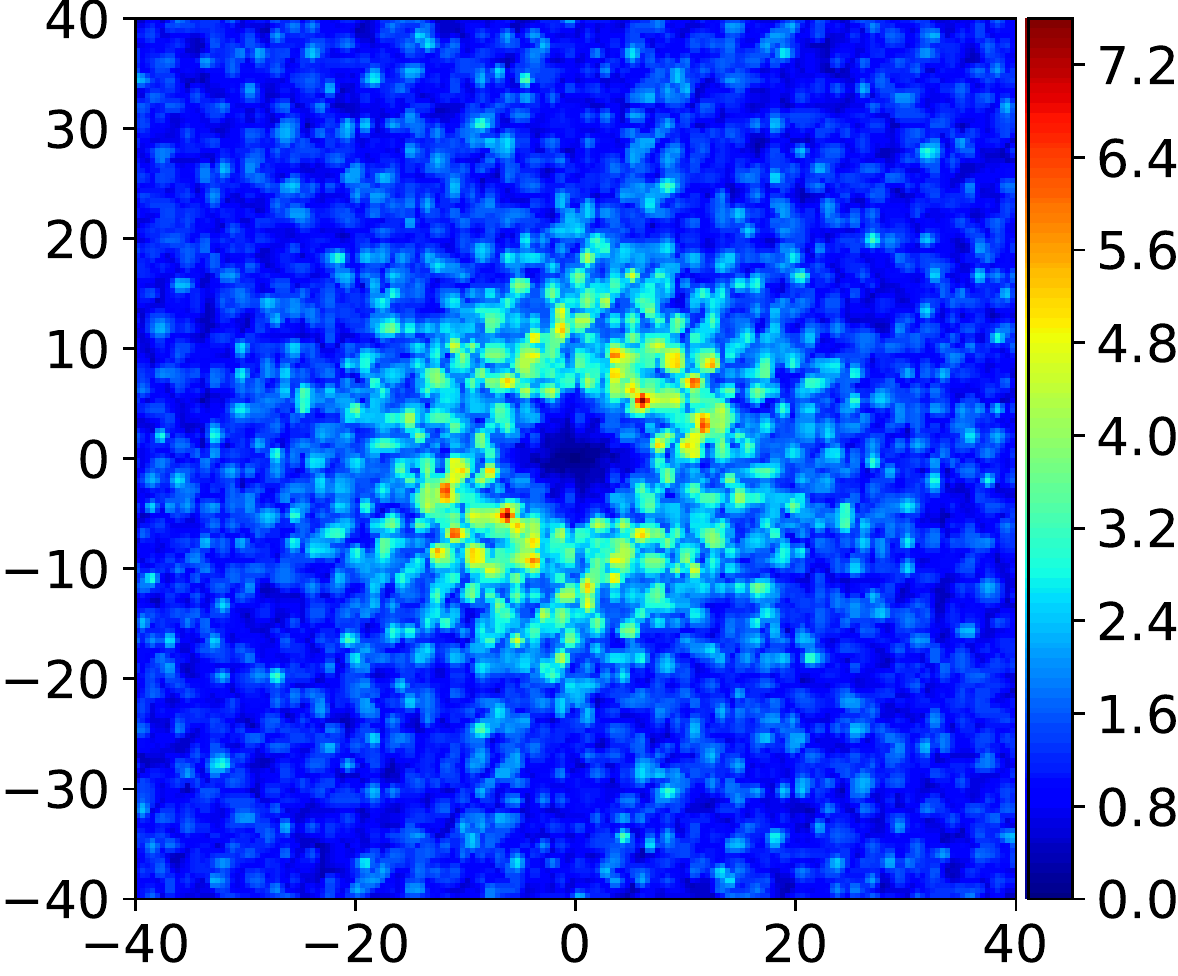}%
	\hspace{0.01\linewidth}
	\includegraphics[width=0.15\linewidth]{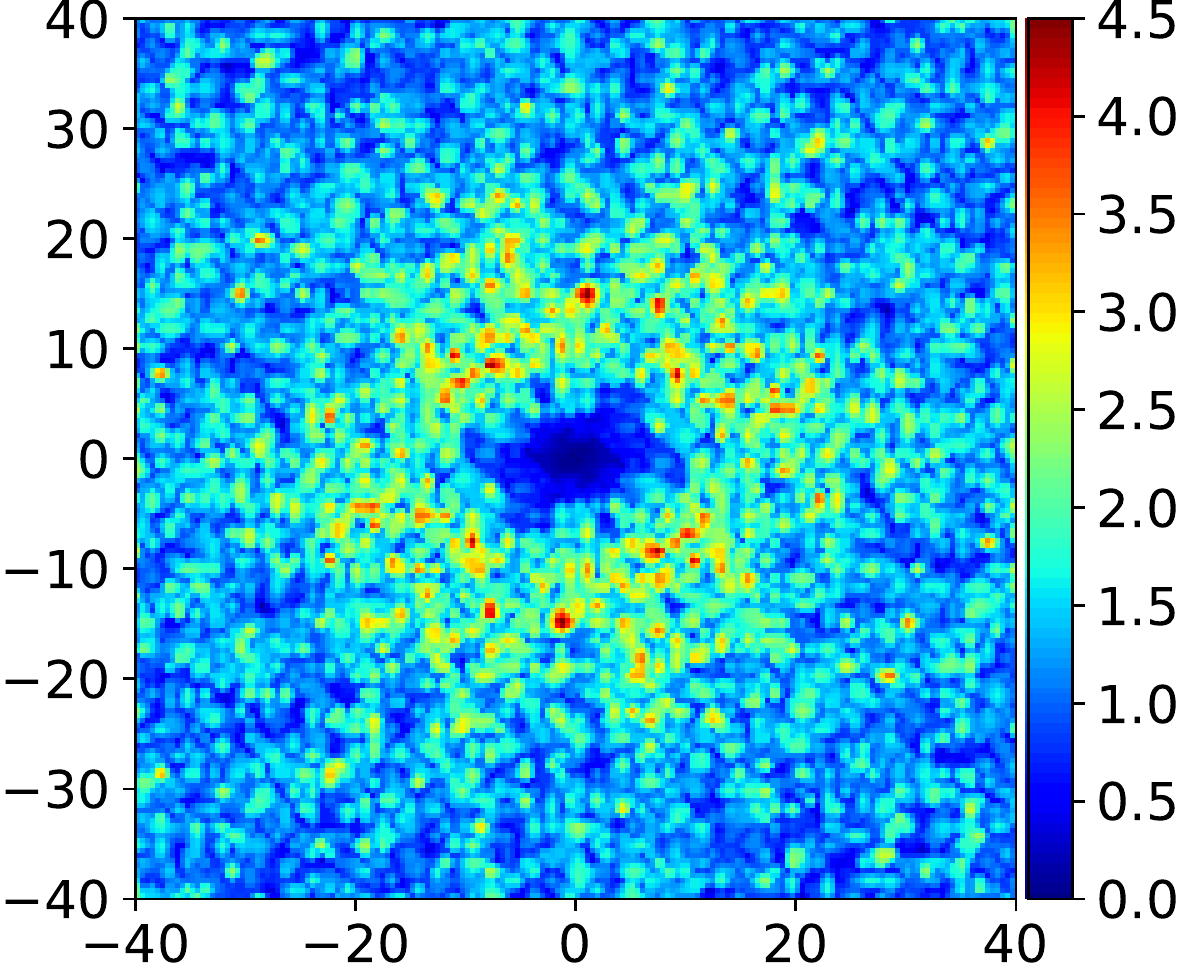}%
	\hspace{0.01\linewidth}
	\includegraphics[width=0.15\linewidth]{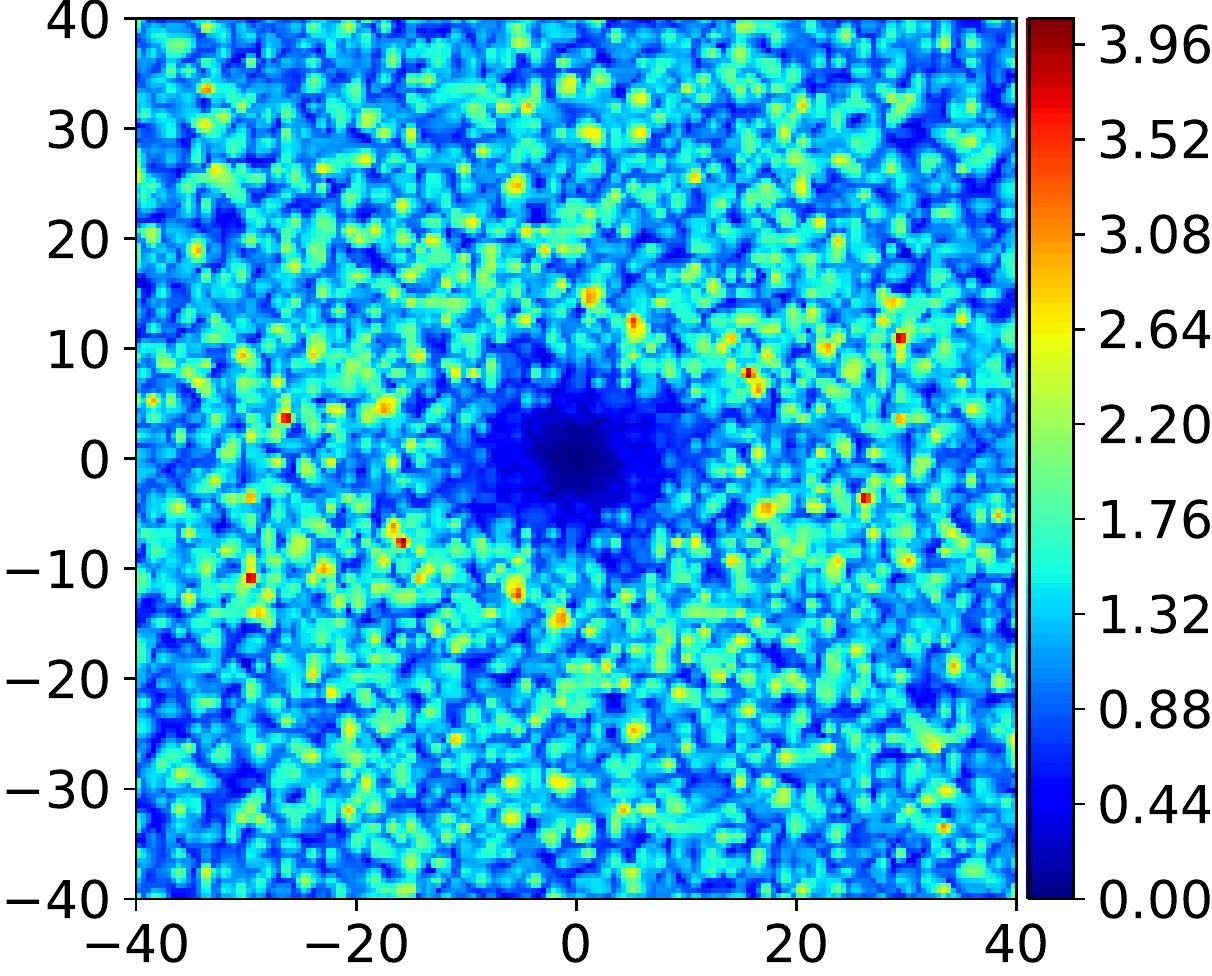}%
	\hspace{0.01\linewidth}
	\includegraphics[width=0.15\linewidth]{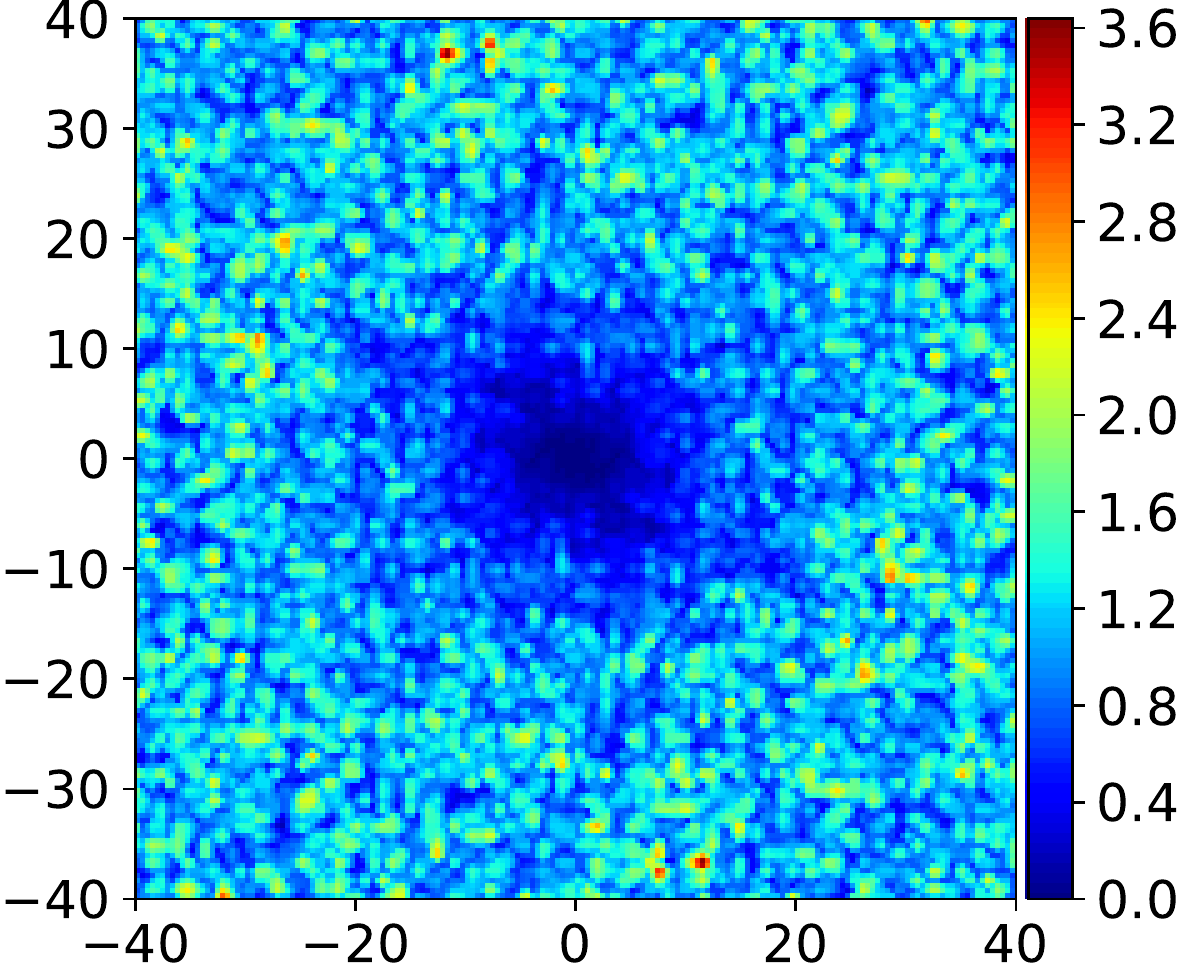}%

	\protect
	\caption{Eigenvectors of Gramian $G_{t}$ at $t = 600000$. First two rows: from left-to-right, 6 first eigenvectors and their Fourier Transforms (see the Appendix \ref{sec:AppF} for details). Last two rows: 10-th, 100-th, 500-th, 1000-th, 2000-th and 4000-th eigenvectors, and their Fourier Transforms. As observed, a frequency of signal inside of each eigenvector increases when moving from large to small eigenvalue.
	}
	\label{fig:NNSpectrum1}
\end{figure}

\begin{figure}
	\centering\offinterlineskip
	\newcommand{\imagepath}[0] {Figures/FC_NN_6L/time_00020000}
	\includegraphics[width=0.166\linewidth]{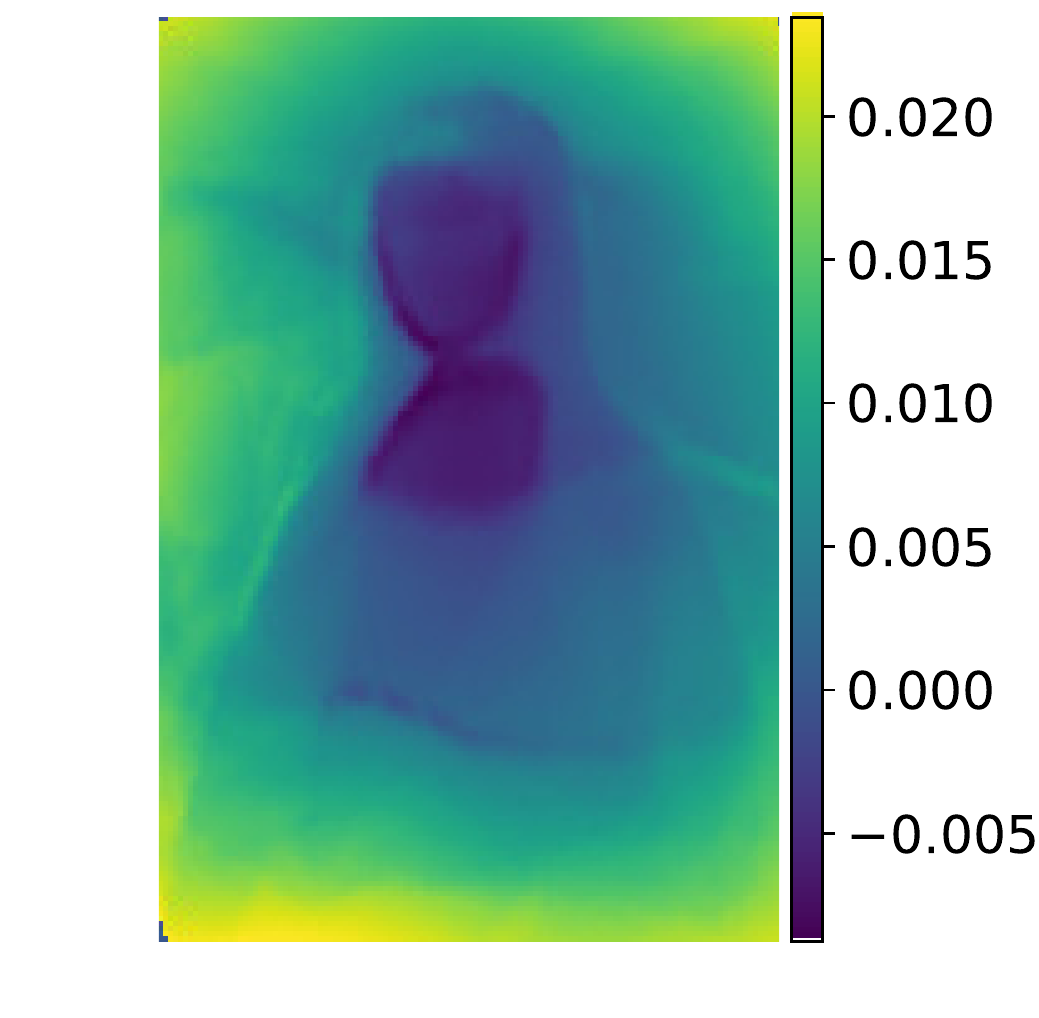}%
	\includegraphics[width=0.166\linewidth]{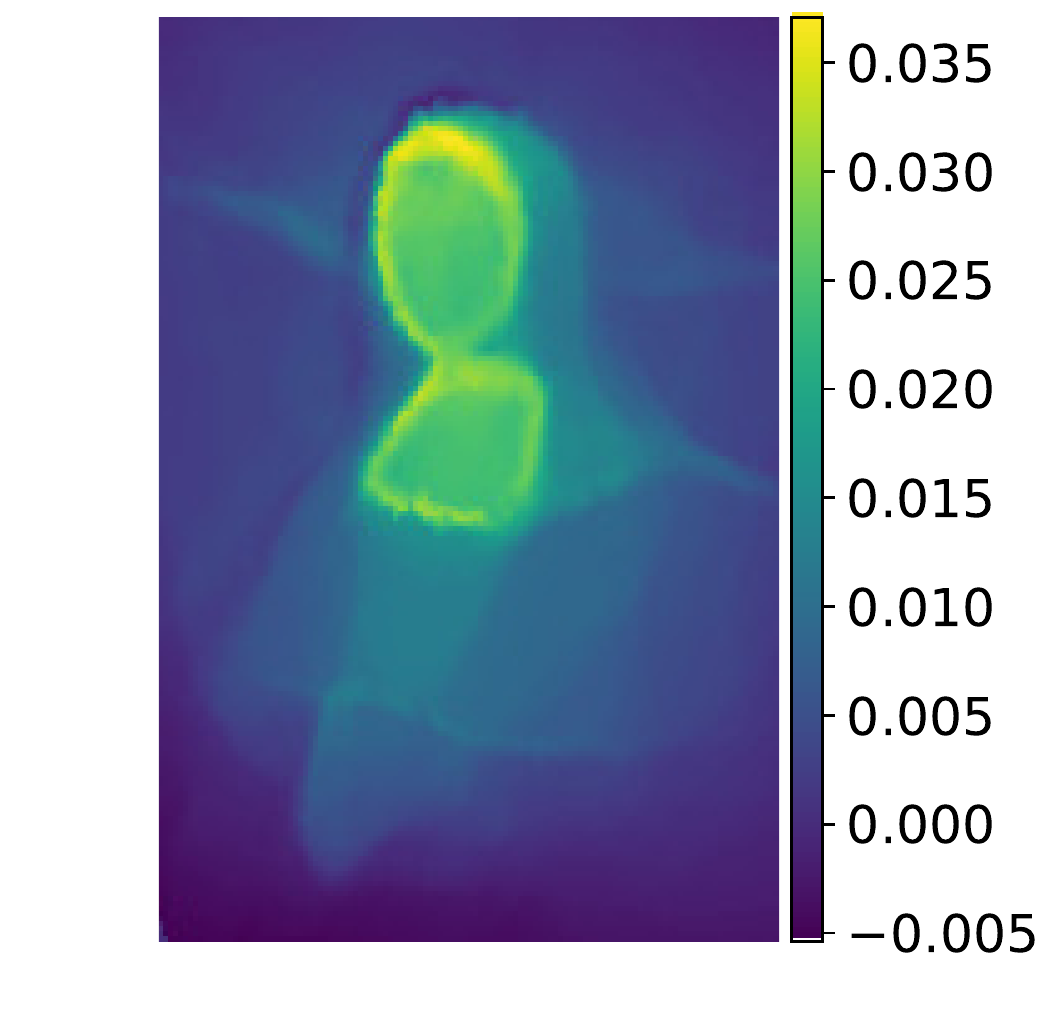}%
	\includegraphics[width=0.166\linewidth]{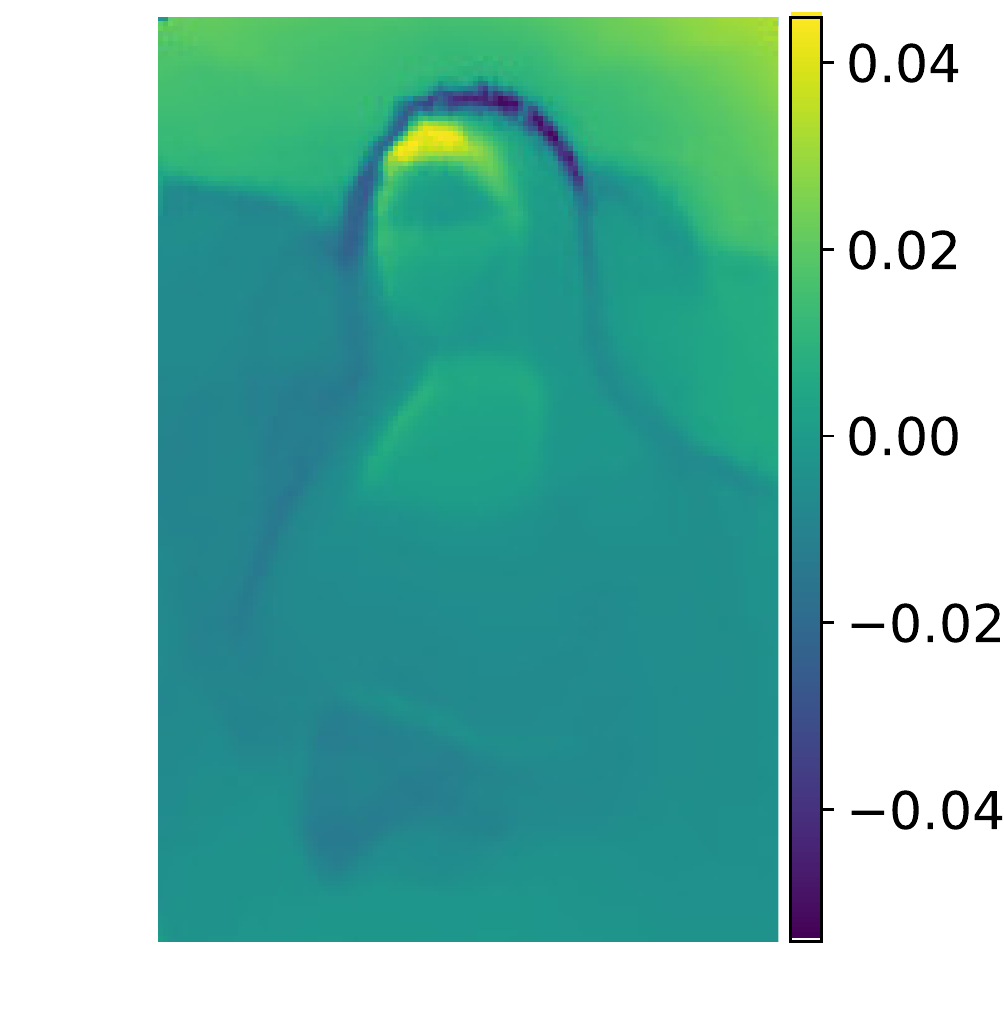}%
	\includegraphics[width=0.166\linewidth]{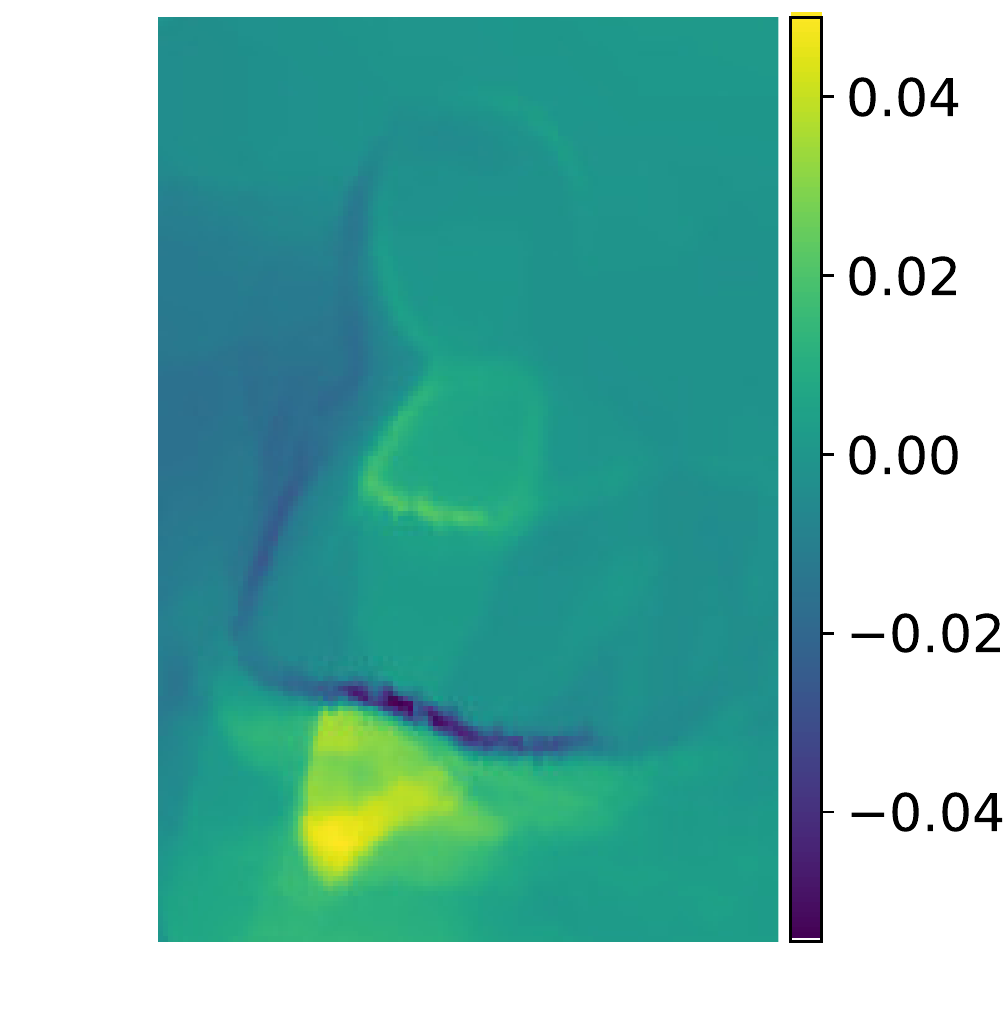}%
	\includegraphics[width=0.166\linewidth]{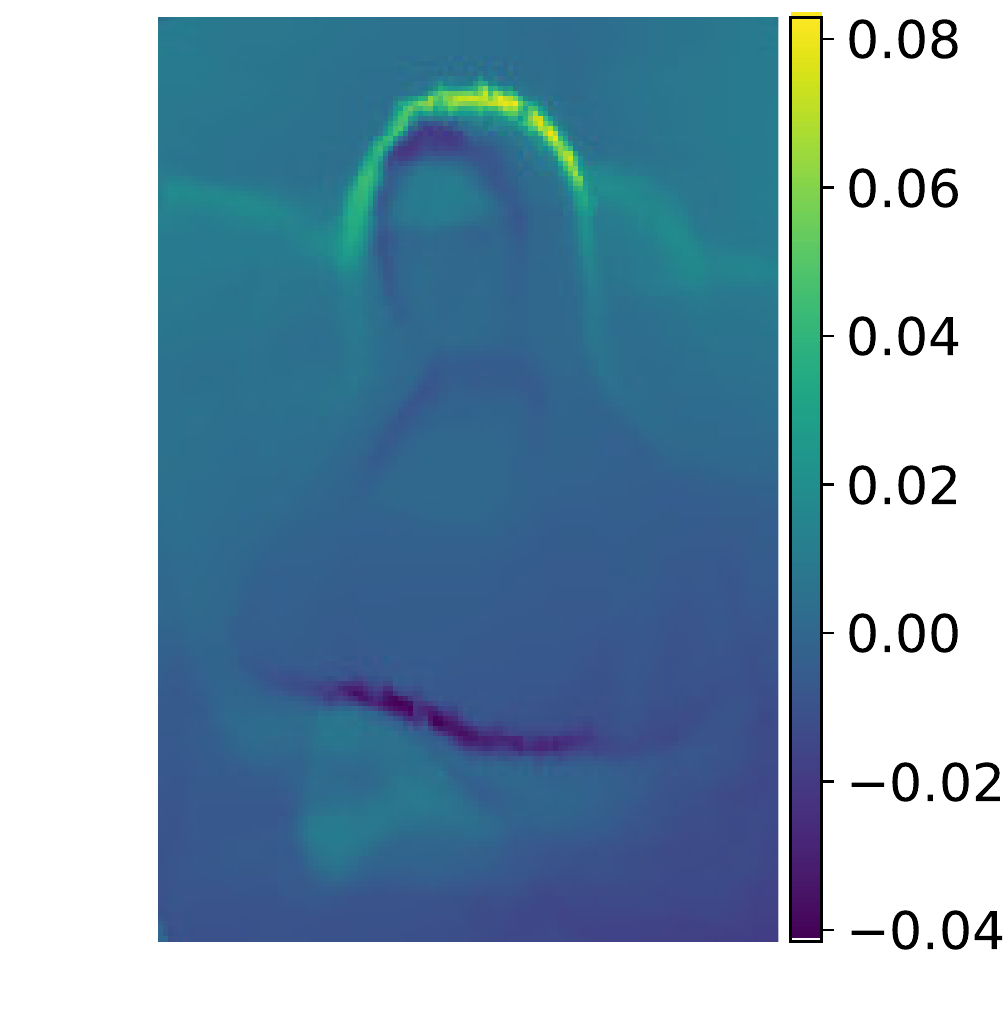}%
	\includegraphics[width=0.166\linewidth]{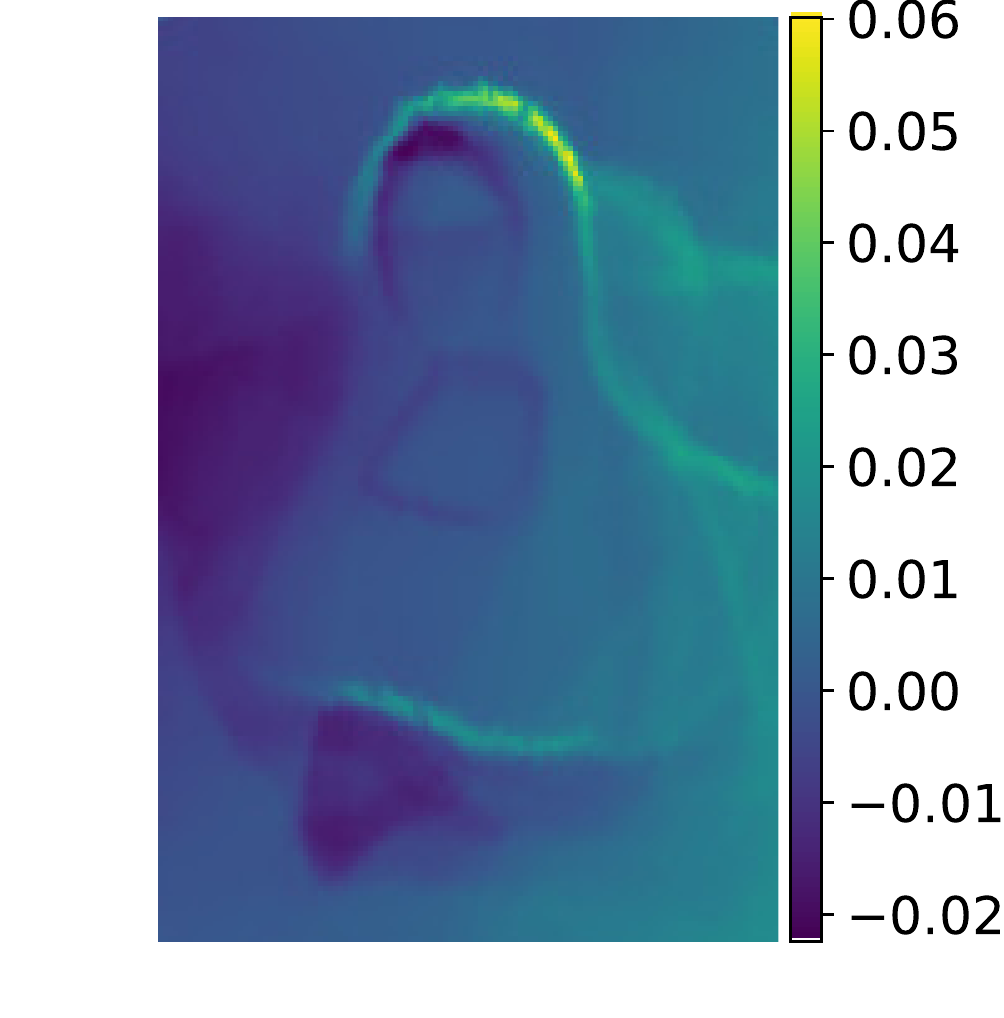}%
	\\

	\includegraphics[width=0.166\linewidth]{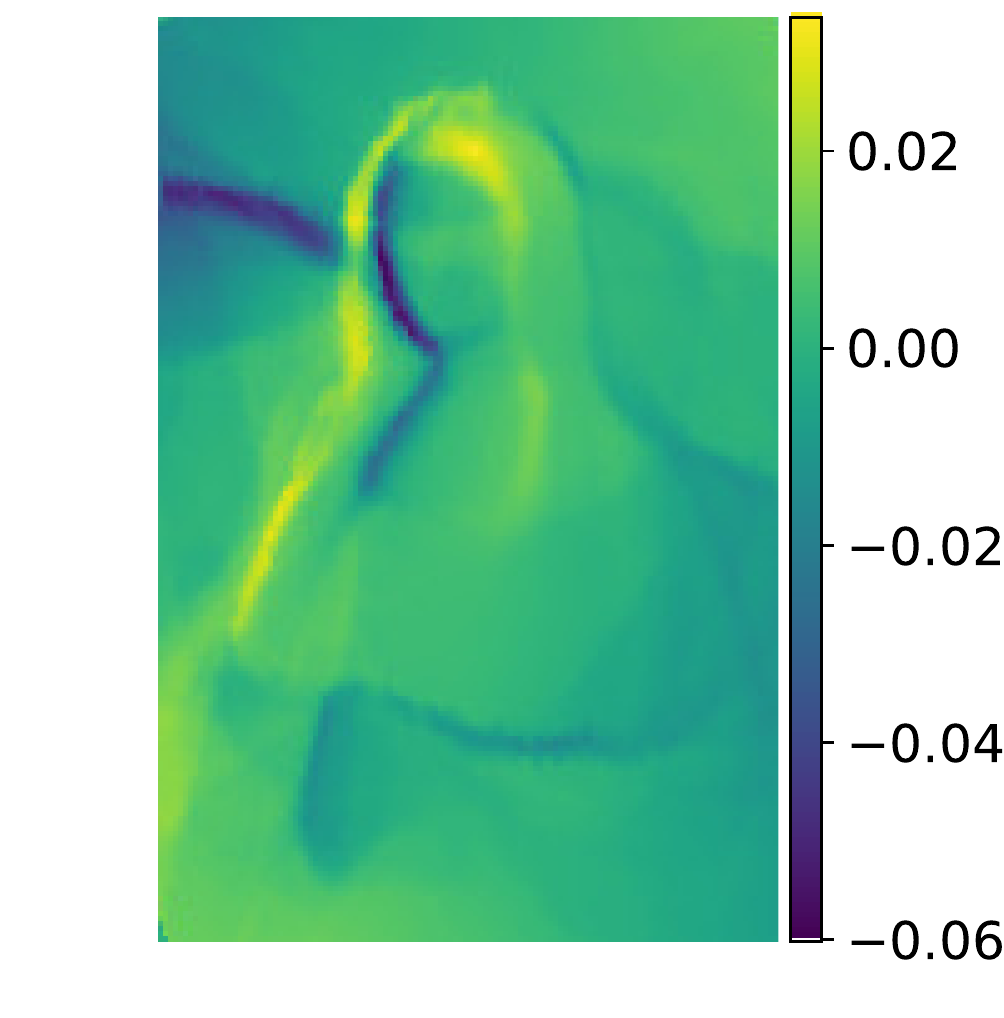}%
	\includegraphics[width=0.166\linewidth]{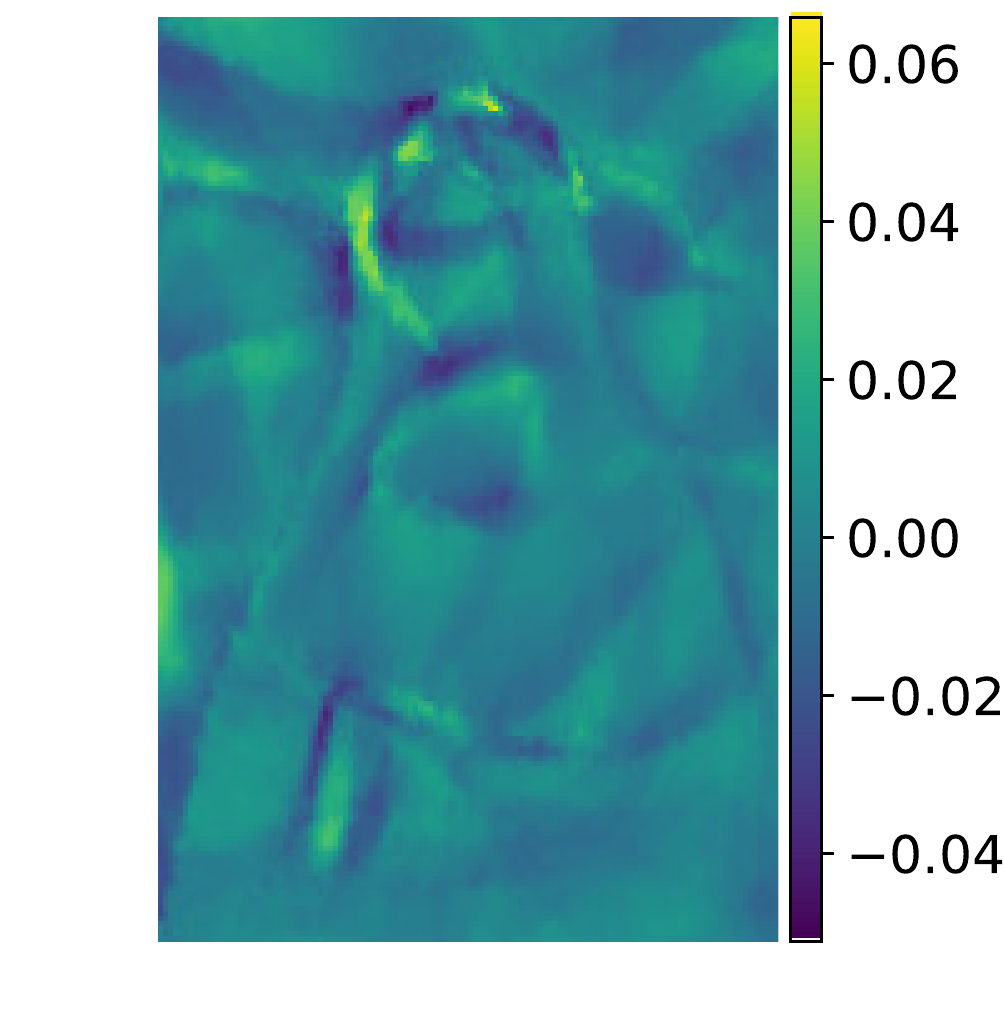}%
	\includegraphics[width=0.166\linewidth]{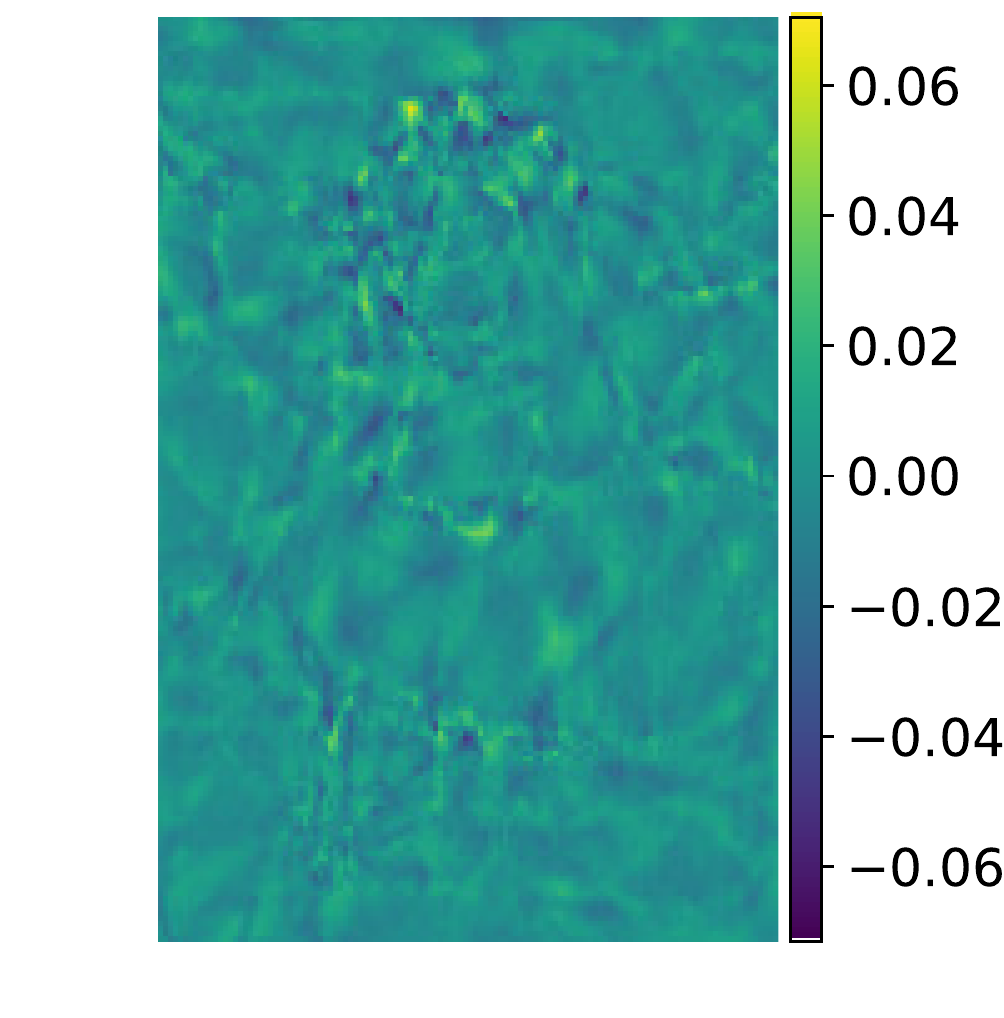}%
	\includegraphics[width=0.166\linewidth]{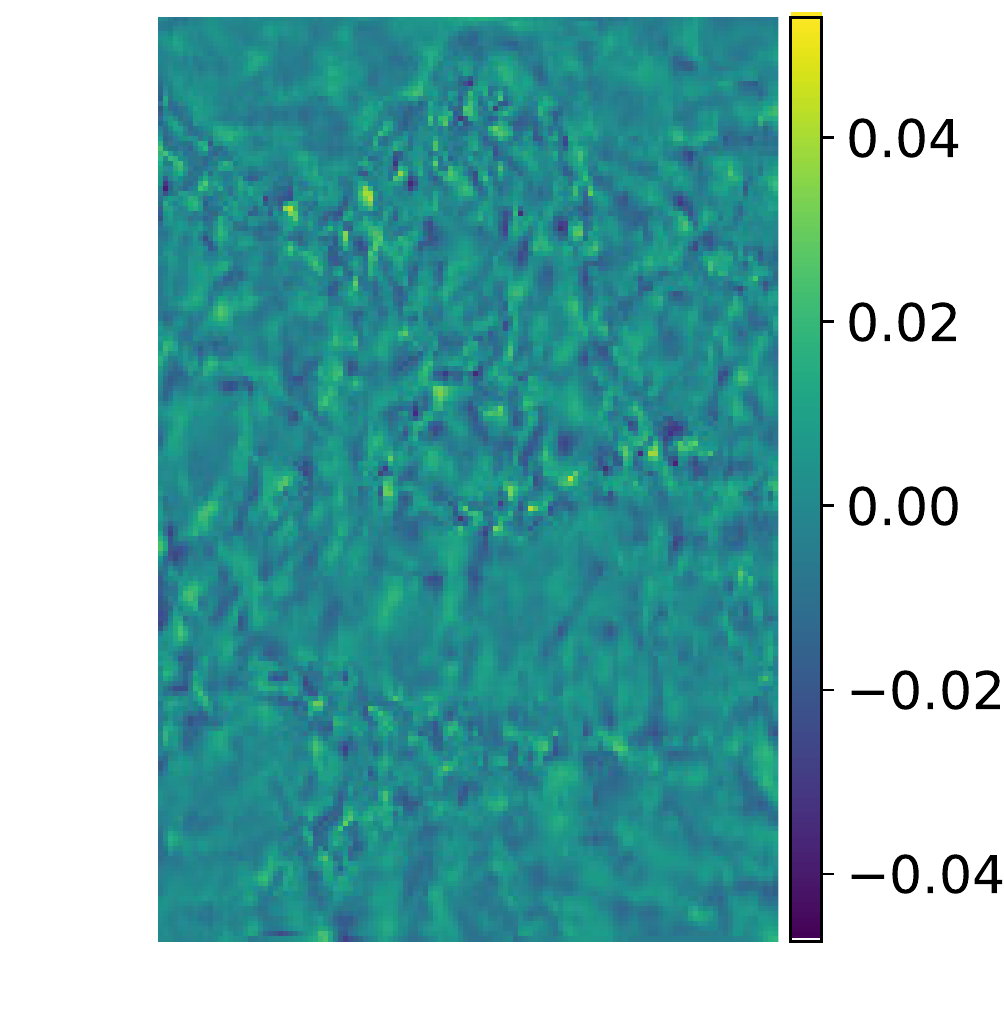}%
	\includegraphics[width=0.166\linewidth]{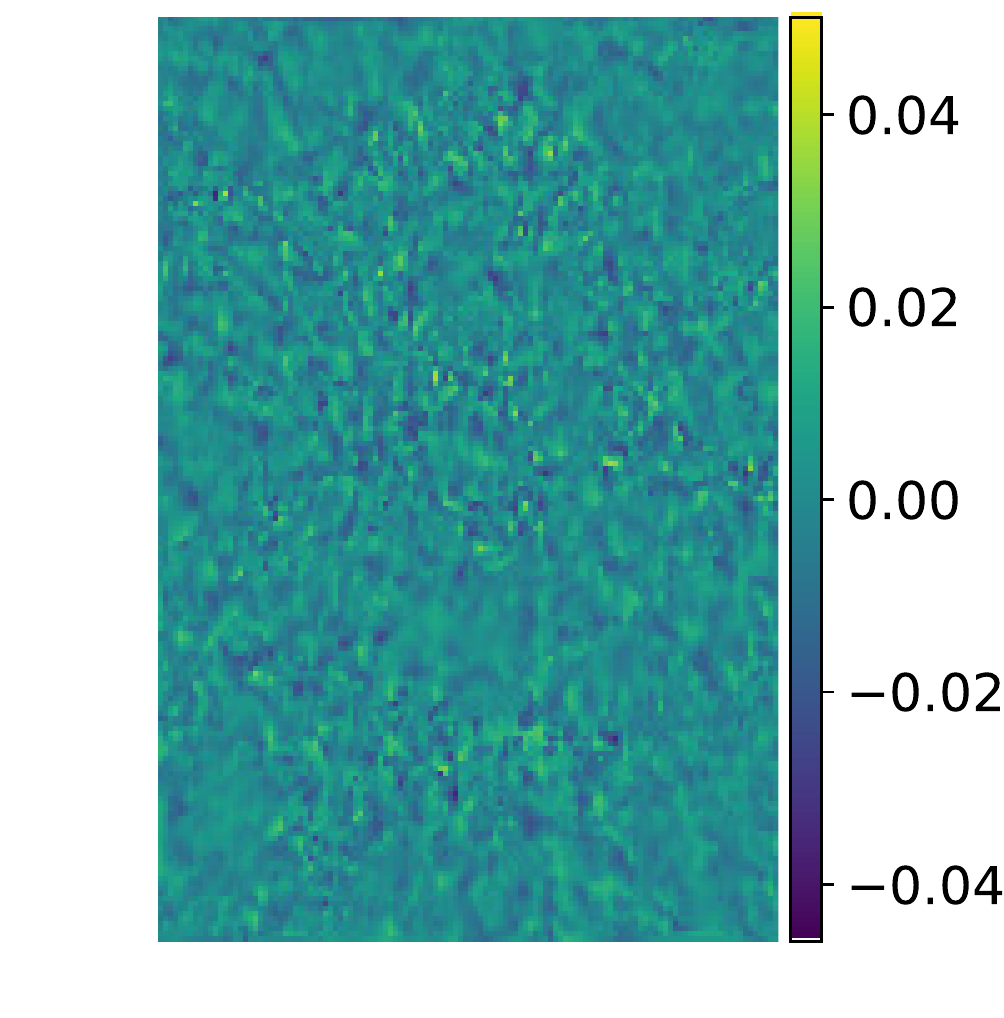}%
	\includegraphics[width=0.166\linewidth]{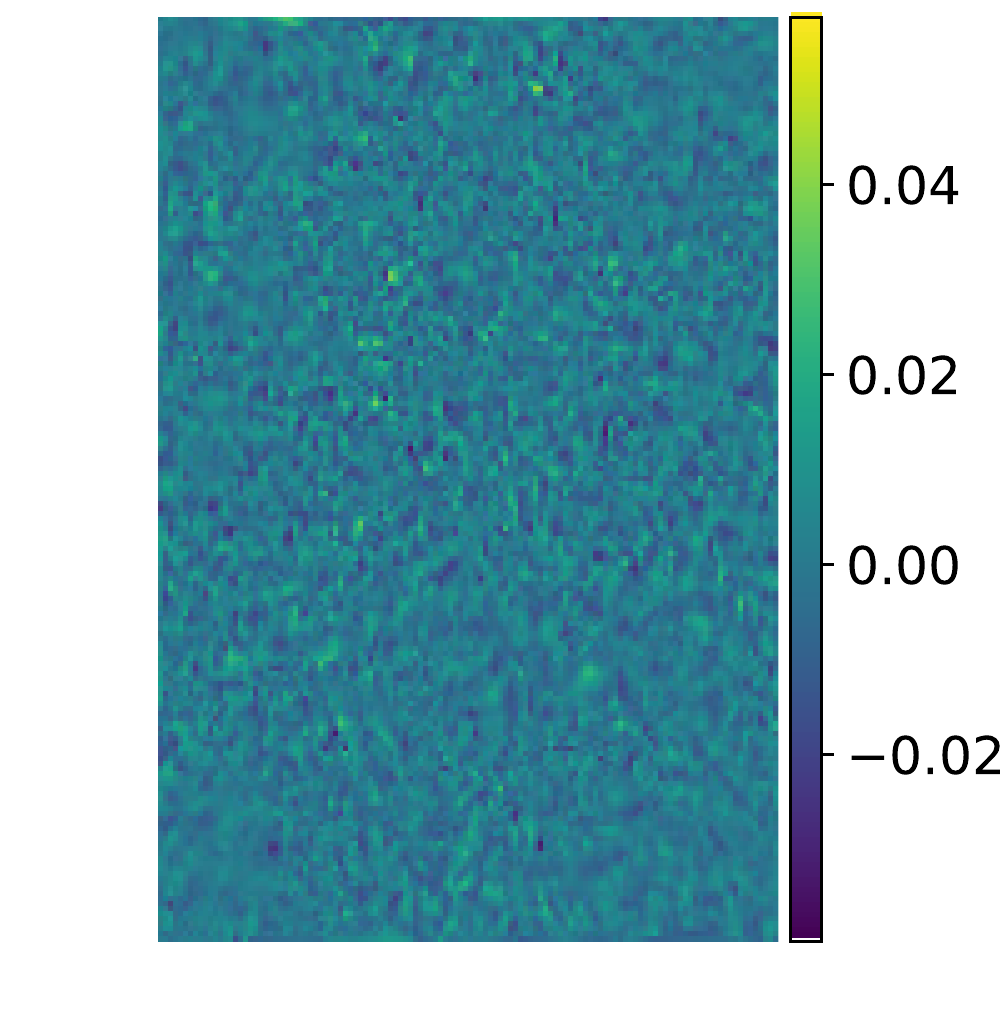}%

	\protect
	\caption{First line: from left-to-right, 6 first eigenvectors of Gramian $G_{t}$ at $t = 20000$. Second line: 10-th, 100-th, 500-th, 1000-th, 2000-th and 4000-th eigenvectors.
	}
	\label{fig:NNSpectrum4}
\end{figure}

We further explore $\{\bar{\upsilon}_{i}^t\}$ in a more illustrative manner, to produce a better intuition about their nature. In Figure \ref{fig:NNSpectrum3-d} a linear combination of several \topp eigenvectors at $t = 600000$ is presented, showing that with only 100 vectors we can accurately approximate the NN output. 

Furthermore, in Figure \ref{fig:NNSpectrum1} several eigenvectors are interpolated to entire $[0, 1]^2$. We can see that \topp $\{\bar{\upsilon}_{i}^t\}$ obtained visual similarity with various parts of Mona Lisa image and indeed can be seen as basis functions of $f_{\theta}(X)$ depicted in Figure \ref{fig:LearningDetails-b}. Likewise, we also demonstrate the Fourier Transform of each $\bar{\upsilon}_{i}^t$. As observed, the frequency of the contained information is higher for smaller eigenvalues, supporting conclusions of \cite{Basri19arxiv}. More eigenvectors are depicted in SM.

Likewise, in Figure \ref{fig:NNSpectrum4} same eigenvectors are displayed at $t = 20000$. At this time the visual similarity between each one of first eigenvectors and the target function in Figure \ref{fig:LearningDetails-a} is much stronger. This can be explained by the fact that the information about the target function within $G_t$ is spread from first few towards higher number of \topp eigenvectors after each learning rate drop, as was described above. Hence, before the first drop at $t = 100000$ this information is mostly gathered within first few $\{\bar{\upsilon}_{i}^t\}$ (see also $E_t(\bar{y}, 10)$ in Figure \ref{fig:NNSpectrum1.0-a}).

\paragraph{Alignment and NN Depth / Width}

Here we further study how a width and a depth of NN affect the alignment between $G_t$ and the ground truth signal $\bar{y}$. To this purpose,
we performed the optimization under the identical setup, yet with NNs containing various numbers of layers and neurons. In Figure \ref{fig:NNSpectrum1.1-a} we can see that in deeper NN \topp eigenvectors of $G_t$ aligned more towards $\bar{y}$ - the relative energy $E_t(\bar{y}, 400)$ is higher for a larger depth. This implies that more layers, and the higher level of non-linearity produced by them, yield a better alignment between $G_t$ and $\bar{y}$. In its turn this allows NN to better approximate a given target function, as shown in Figures \ref{fig:NNSpectrum1.1-b}-\ref{fig:NNSpectrum1.1-c}, making it more expressive for a given task. Moreover, in evaluated 2-layer NNs, with an increase of neurons and parameters the alignment rises only marginally.

\paragraph{Spectrum Preservation}

Next, we examine how stable are eigenvectors of $G_t$ along $t$. For this we explore the relative energy of $G_{600000}$'s eigenvectors, final eigenvectors of the optimization, within spectrum of $G_{20000}$. Note that we compare spectrums at $t = 600000$ and $t = 20000$ to skip first several thousands of iterations since during this bootstrap period the change of $G_t$ is highly notable.

In Figure \ref{fig:NNSpectrum1.1-d} we depict $E_{20000}(\bar{\upsilon}_{i}^{600000}, k)$ as a function of $k$, for various $\{ \bar{\upsilon}_{i}^{600000} \}$. As observed, 10 first \topp eigenvectors of $G_{600000}$ are also located in the \topp spectrum of $G_{20000}$ - the function $E_{20000}(\bar{\upsilon}_{i}^{600000}, k)$ is almost 1 for even relatively small $k$. Hence, the \topp Gramian spectrum was preserved, roughly, along the performed optimization. 
Further, eigenvectors of smaller eigenvalues (i.e. with higher indexes $i$) are significantly less stable, with large amount of their energy widely spread inside \btm eigenvectors of $G_{20000}$. Moreover, we can see a clear trend that with higher $i$ the associated eigenvector is less preserved.

\paragraph{Scope of Analysis}

The above empirical analysis was repeated under numerous different settings and can be found in SM \cite{Kopitkov19dm_Supplementary}. We evaluated various FC architectures, with and without shortcuts between the layers and including various activation functions. Likewise, optimizers GD, stochastic GD and Adam were tested on problems of regression (L2 loss) and density estimation (noise contrastive estimation \cite{Gutmann10aistats}). Additionally, various high-dimensional real-world datasets were tested, including MNIST and CIFAR100. \textbf{All} experiments exhibit the same alignment nature of kernel towards the learned target function.

\section{Discussion and Conclusions}
\label{sec:Concl}

In this paper we empirically revealed that during GD  \topp eigenfunctions of \emph{gradient similarity} kernel change to align with the target function $y(X)$ learned by NN $f_{\theta}(X)$, and hence can be considered as a NN memory tuned during the optimization to better represent $y(X)$. This alignment is significantly higher for deeper NNs, whereas a NN width has only a minor effect on it.
Moreover, the same \topp eigenfunctions represent
a \emph{neural spectrum} - the $f_{\theta}(X)$ is a linear combination of these eigenfunctions during the optimization. As well, we showed various trends of the kernel dynamics as a result of the learning rate decay, accounting for which we argue may lead to a further progress in DL theory. The considered herein optimization scenarios include various \supp and \unsupp losses over various high-dimensional datasets, optimized via several different optimizers. Several variants of FC architecture were evaluated. The results are consistent with our previous experiments in \cite{Kopitkov19arxiv}.

The above revealed behavior leads to several implications. First, our empirical study suggests that the high approximation power of deep models is produced by the above alignment capability of the \emph{gradient similarity}, since the learning along its \topp eigenfunctions is considerably faster. Furthermore, it also implies that the family of functions that a NN can approximate (in reasonable time) is limited to functions within the \topp spectrum of the kernel. Recently, it was proved in \cite{Arora19arxiv,Basri19arxiv,Oymak19arxiv}. Thus, it leads to the next main question - how the NN architecture and optimization hyper-parameters affect this spectrum, and what is their optimal configuration for learning a given function $y(X)$. Moreover, NN dynamics behavior beyond first-order Taylor expansion is still unexplored. We shall leave it for a future exciting research.

\section{Acknowledgments}

The authors thank Daniel Soudry and Dar Gilboa for discussions on dynamics of a Neural Tangent Kernel (NTK).
This work was supported in part by the Israel Ministry of Science \& Technology (MOST) and Intel Corporation. We gratefully acknowledge the support of NVIDIA Corporation with the donation of the Titan Xp GPU, which, among other GPUs, was used for this research.

\appendix

\section{Appendix: Relation between spectrums of $g_{t}(X, X')$ and its Gramian $G_t$}
\label{sec:AppSp}

Consider $N$ dataset points $\dtX = \{ X^i \in \RR^d \}_{i = 1}^{N}$ sampled from an arbitrary probability density function (pdf) $P(X)$. Further, consider a kernel $g_{t}(X, X')$ and the corresponding Gramian $G_t$ defined on $\dtX$, with $G_{t}(i,j) = g_{t}(X^i, X^j)$. Eigenvalues $\{ \tilde{\lambda}_k \}_k$, sorted in decreasing order, and eigenvectors $\{ \tilde{\upsilon}_k(\cdot) \}_k$ of $g_{t}(\cdot, \cdot)$ w.r.t. $P(X)$ are defined as solutions of:
\begin{equation}
\tilde{\lambda}_k
\cdot
\tilde{\upsilon}_k(X)
=
\int
g_{t}(X, X')
\cdot
\tilde{\upsilon}_k(X')
\cdot
P(X)
dX'
.
\label{eq:KernelSpctrm}
\end{equation}

The integral in Eq.~(\ref{eq:KernelSpctrm}) can be approximated via a sampled approximation:
\begin{equation}
\int
g_{t}(X, X')
\cdot
\tilde{\upsilon}_k(X')
\cdot
P(X)
dX'
\approx
\frac{1}{N}
\sum_{i = 1}^{N}
g_{t}(X, X^i)
\cdot
\tilde{\upsilon}_k(X^i)
,
\label{eq:IntgrlApprox}
\end{equation}
with the LHS of the above expression converging to the RHS as $N \rightarrow \infty$ due to the law of large numbers.

Further, denote by $\bar{\upsilon}_k$ a $N \times 1$ vector whose $i$-th entry is $\tilde{\upsilon}_k(X^i)$. Combining Eq.~(\ref{eq:KernelSpctrm}) and Eq.~(\ref{eq:IntgrlApprox}), $\bar{\upsilon}_k$ can be written as:
\begin{equation}
\tilde{\lambda}_k
\cdot
\bar{\upsilon}_k
=
\frac{1}{N}
G_t
\cdot
\bar{\upsilon}_k
,
\label{eq:GSpctrm}
\end{equation}
where we can see $\bar{\upsilon}_k$ to be eigenvector of $G_t$. Therefore, eigenvectors $\{ \bar{\upsilon}_k \}_k$ of $G_t$ can be considered as unbiased estimations of eigenfunctions $\{ \tilde{\upsilon}_k(\cdot) \}_k$ at points in $\dtX$. Note that the above sampled approximations are expected to be less accurate for larger indexes $k$ since the corresponding $\tilde{\upsilon}_k(\cdot)$ will contain more high-frequency oscillations.

Furthermore, from Eq.~(\ref{eq:GSpctrm}) it is clear that each $\bar{\upsilon}_k$ is associated with the eigenvalue $\lambda_k = N \cdot \tilde{\lambda}_k$ of $G_t$. Hence, eigenvalues $\{ \lambda_k \}_k$ of $G_t$ can be considered as unbiased estimations of eigenfunctions $\{ \tilde{\lambda}_k \}_k$, up to a multiplier $N$.

Likewise, $\tilde{\upsilon}_k(X)$ at an arbitrary point $X$ can be estimated in a similar way, by combining Eq.~(\ref{eq:KernelSpctrm}) and Eq.~(\ref{eq:IntgrlApprox}):
\begin{equation}
\tilde{\lambda}_k \cdot \tilde{\upsilon}_k(X)
\approx
\frac{1}{N}
\sum_{i = 1}^{N}
g_{t}(X, X^i)
\cdot
\tilde{\upsilon}_k(X^i)
\quad
\Longrightarrow
\quad
\lambda_k \cdot \tilde{\upsilon}_k(X)
\approx
g_{t}(X, \dtX)
\cdot
\bar{\upsilon}_k
,
\label{eq:TestingSpectrum}
\end{equation}
where $g_{t}(X, \dtX)$ is a row vector with $g_{t}(X, \dtX)_{(i)} = g_{t}(X, X^i)$. The above approximation is used in the Appendix \ref{sec:AppC_test} to derive NN dynamics at testing points.

\section{Appendix: Relation between FIM and Hessian of the Loss}
\label{sec:AppA}

Hessian of a typical loss in Eq.~(\ref{eq:GeneralLoss}) can be written as:
\begin{equation}
H_t
\triangleq
\frac{\partial^2 L(\theta_t, D)}{\partial \theta^2}
=
\frac{1}{N}
A_t
D_t
A_t^T
+
\frac{1}{N}
\sum_{i = 1}^{N}
\ell'
\left[
X^i,
Y^i,
f_{\theta_t}(X^i)
\right]
\cdot
\mathcal{H}_{t}(X^i)
,
\label{eq:GeneralHessian}
\end{equation}
where $A_t$ is Jacobian matrix defined in Section \ref{sec:FIMSec}, $D_t$ is a diagonal matrix with $D_{t}(i,i) = 
\frac{\partial^2 \ell
	\left[
	X^i,
	Y^i,
	f_{\theta_{t}}(X^i)
	\right]}{\partial f_{\theta}^2}
$ and $\mathcal{H}_{t}(X) \triangleq \frac{\partial^2 f_{\theta_t}(X)}{\partial \theta^2}$ is the model Hessian.

Further, in case of L2 loss we will have  $D_{t} = I$ and 
\begin{equation}
H_t
=
\frac{1}{N}
F_t
+
\frac{1}{N}
\sum_{i = 1}^{N}
\ell'
\left[
X^i,
Y^i,
f_{\theta_t}(X^i)
\right]
\cdot
\mathcal{H}_{t}(X^i)
.
\label{eq:L2Hessian}
\end{equation}

Finally, considering final stages of the optimization, the residual $\ell'
\left[
X^i,
Y^i,
f_{\theta_t}(X^i)
\right] = f_{\theta_t}(X^i) - Y^i$ is approximately zero and hence the second term of Eq.~(\ref{eq:L2Hessian}) RHS can be neglected. Therefore, for L2 loss we will have $H_t \approx \frac{1}{N}
F_t
$.

Beyond L2 loss, a connection between FIM and the loss Hessian was also observed for the cross-entropy loss in \cite{Gur18arxiv}. Authors empirically observed that the loss gradient $\nabla_{\theta} L(\theta_t, D)$ converges very fast into a tiny subspace spanned by a few \topp eigenvectors of $H_t$. This suggests that \topp eigenvectors of $H_t$ and $F_t$ are tightly aligned and are spanning the same subspace of $\RR^{|\theta|}$ also for cross-entropy case, as follows. Denote $A_t$'s SVD as triplets $\{\sqrt{\lambda_i^t}, \bar{\omega}_{i}^t, \bar{\upsilon}_{i}^t \}_{i = 1}^{N'}$ of ordered singular values, left and right singular vectors respectively, where $N'$ is a number of non-zero singular values. Then, $\nabla_{\theta} L(\theta_t, D)$ can be written as:
\begin{equation}
\nabla_{\theta} L(\theta_t, D)
=
\frac{1}{N}
A_t
\cdot
\bar{m}_{t}
=
\frac{1}{N}
\left[
\sum_{i = 1}^{N'}
\sqrt{\lambda_i^t}
\cdot
\bar{\omega}_{i}^t
\cdot
(\bar{\upsilon}_{i}^t)^T
\right]
\cdot
\bar{m}_{t}
=
\frac{1}{N}
\sum_{i = 1}^{N'}
\sqrt{\lambda_i^t}
<\bar{\upsilon}_{i}^t, \bar{m}_{t}>
\bar{\omega}_{i}^t
.
\label{eq:LossGrrr}
\end{equation}
Due to typical extremely fast decay of $\lambda_i^t$ w.r.t. $i$, described along this paper, $\nabla_{\theta} L(\theta_t, D)$ in the above expression can be roughly seen as a linear combination of only $\{ \bar{\omega}_{i}^t \}$ associated with several \topp $\{ \lambda_i^t \}$. Noting that these are also the \topp eigenvectors of $F_t$, we see that $\nabla_{\theta} L(\theta_t, D)$ is located in top-spectrum of $F_t$. Further, taking into account the empirical observation from \cite{Gur18arxiv}, we can conclude from above that \topp eigenvectors of $F_t$ and $H_t$ are tightly aligned.

%

\section{Appendix: Movement of $\theta$ along FIM Eigenvector causes Movement of NN Output along Gramian Eigenvector}
\label{sec:AppB}

To understand the relation between FIM $F_t$ and Gramian $G_t$ more intuitively, here we show their dual connection in terms of how the movement along FIM eigenvector $\bar{\omega}_{i}^t$ in $\theta$-space affects the movement in the function space. Specifically, consider $\bar{f}_{t}$ to be a vector of NN outputs at training points at optimization time $t$, similarly to the formulation in Section \ref{sec:Nottn}. Further, consider a movement of the model in $\theta$-space from current $\theta_t$ to a new location $\theta_{t'} = \theta_t + \sqrt{\lambda_i^t} \cdot \bar{\omega}_{i}^t$ in direction $\bar{\omega}_{i}^t$ where $\sqrt{\lambda_i^t}$ is used as a step size. Then the $\bar{f}_{t'}$ at the new location can be approximated via first-order Taylor as:
\begin{equation}
\bar{f}_{t'}
=
\bar{f}_{t}
+
\sqrt{\lambda_i^t}
\cdot
A_t^T
\cdot
\bar{\omega}_{i}^t
,
\label{eq:FMovement}
\end{equation}
where $A_t$ is Jacobian matrix defined in Section \ref{sec:FIMSec}. Moreover, considering the singular value decomposition (SVD) of $A_t$, we can see that $\bar{f}_{t'}
-
\bar{f}_{t}
=
\lambda_i^t
\cdot
\bar{\upsilon}_{i}^t
$. That is, walking in the direction $\bar{\omega}_{i}^t$ in $\theta$-space changes NN outputs only along $\bar{\upsilon}_{i}^t$, according to first-order dynamics.

\section{Appendix: Dynamics of L2 Loss for a Fixed Gramian, at Training Points}
\label{sec:AppC}

Consider Eq.~(\ref{eq:FVecDff}) with a fixed Gramian $G$ whose eigenvalues and eigenvectors are $\{ \lambda_{i} \}_{i = 1}^{N}$ and $\{ \bar{\upsilon}_{i} \}_{i = 1}^{N}$ respectively. Define $N'$ to be a number of non-zero eigenvalues. Likewise, consider the residual vector $\bar{m}_{t} = \bar{f}_{t} - \bar{y}$ whose first-order dynamics can be written as:
\begin{multline}
d \bar{m}_{t}
\triangleq
\bar{m}_{t + 1}
-
\bar{m}_{t}
=
\bar{f}_{t + 1}
-
\bar{f}_{t}
=
d \bar{f}_{t}
=
-
\frac{\delta}{N}
\cdot
G
\cdot
\bar{m}_{t}
\quad
\Longrightarrow
\\
\Longrightarrow
\quad
\bar{m}_{t + 1}
=
\left[
I
-
\frac{\delta}{N}
\cdot
G
\right]
\cdot
\bar{m}_{t}
\quad
\Longrightarrow
\quad
\bar{m}_{t}
=
\sum_{i = 1}^{N'}
\left[
1 - 
\frac{\delta}{N}
\lambda_i
\right]^t
<\bar{\upsilon}_{i}, \bar{m}_{0}>
\bar{\upsilon}_{i}
+
\bar{m}_{0}^z
,
\label{eq:MDff}
\end{multline}
where $\bar{m}_{0}^z$ is a projection of $\bar{m}_{0}$ to null-space of $G$, with $G \cdot \bar{m}_{0}^z = \bar{0}$.

Further, noting that:
\begin{equation}
\sum_{j = 0}^{t - 1}
\bar{m}_{j}
=
\sum_{i = 1}^{N'}
\frac{1 - 
\left[
1 - 
\frac{\delta}{N}
\lambda_i
\right]^t
}
{\frac{\delta}{N}
	\lambda_i
	}
<\bar{\upsilon}_{i}, \bar{m}_{0}>
\bar{\upsilon}_{i}
+
t \bar{m}_{0}^z
,
\label{eq:MSum}
\end{equation}
the $\bar{f}_{t}$ can be then rewritten as:
\begin{multline}
\bar{f}_{t}
=
\bar{f}_{0}
+
\sum_{j = 0}^{t - 1}
d \bar{f}_{j}
=
\bar{f}_{0}
-
\frac{\delta}{N}
G
\cdot
\sum_{j = 0}^{t - 1}
\bar{m}_{j}
=
\bar{f}_{0}
-
\frac{\delta}{N}
G
\cdot
\sum_{i = 1}^{N'}
\frac{1 - 
	\left[
	1 - 
	\frac{\delta}{N}
	\lambda_i
	\right]^t
}
{\frac{\delta}{N}
	\lambda_i
}
<\bar{\upsilon}_{i}, \bar{m}_{0}>
\bar{\upsilon}_{i}
=
\\
=
\bar{f}_{0}
-
\sum_{i = 1}^{N'}
\left[
1 - 
	\left[
	1 - 
	\frac{\delta}{N}
	\lambda_i
	\right]^t
\right]
<\bar{\upsilon}_{i}, \bar{m}_{0}>
\bar{\upsilon}_{i}
.
\label{eq:FDynamics}
\end{multline}

\section{Appendix: Dynamics of L2 Loss for a Fixed Gramian, at Testing Points}
\label{sec:AppC_test}

From Eq.~(\ref{eq:FDff}) we can also derive dynamics of NN output at an arbitrary testing point $X'$:

\begin{equation}
d f_{\theta_{t}}(X') 
=
f_{\theta_{t + 1}}(X')
-
f_{\theta_{t}}(X')
=
-
\frac{\delta}{N}
g(X', \dtX)
\cdot
\bar{m}_{t}
,
\label{eq:FDfffff}
\end{equation}
where $g(X', \dtX) \triangleq \nabla_{\theta} 
f_{\theta_{t}}(X')^T
\cdot
A_t
$ is a row vector with $g(X', \dtX)_{(j)} = g(X', X^j)$. Moreover, similarly to Eq.~(\ref{eq:FDynamics}) we get:
\begin{multline}
f_{\theta_{t}}(X')
=
f_{\theta_{0}}(X')
+
\sum_{j = 0}^{t - 1}
d f_{\theta_{j}}(X')
=
f_{\theta_{0}}(X')
-
\frac{\delta}{N}
g(X', \dtX)
\cdot
\sum_{j = 0}^{t - 1}
\bar{m}_{j}
=\\
=
f_{\theta_{0}}(X')
-
\frac{\delta}{N}
g(X', \dtX)
\cdot
\left[
\sum_{i = 1}^{N'}
\frac{1 - 
	\left[
	1 - 
	\frac{\delta}{N}
	\lambda_i
	\right]^t
}
{\frac{\delta}{N}
	\lambda_i
}
<\bar{\upsilon}_{i}, \bar{m}_{0}>
\bar{\upsilon}_{i}
+
t \bar{m}_{0}^z
\right]
.
\label{eq:FDynamicsTest}
\end{multline}

In case $G$ is invertible (i.e. $\lambda_{min} > 0$), the above expression can also be written as $f_{\theta_{t}}(X')
=
f_{\theta_{0}}(X')
-
g(X', \dtX)
\cdot
G^{-1}
\cdot
\left[
I
-
\left[
I
-
\frac{\delta}{N}
\cdot
G
\right]^t
\right]
\cdot
\bar{m}_{0}
$; a very similar expression was previously derived in \cite{Lee19arxiv}. Likewise, considering the stability condition $\delta < \frac{2N}{\lambda_{max}}$, which is required for a proper optimization convergence $\lim\limits_{t \rightarrow \infty} \left[
1 - 
\frac{\delta}{N}
\lambda_i
\right]^t = 0$, at time $t = \infty$ we will have $f_{\theta_{\infty}}(X')
=
f_{\theta_{0}}(X')
-
g(X', \dtX)
\cdot
G^{-1}
\cdot
\bar{m}_{0}
$.

Furthermore, for a singular $G$ Eq.~(\ref{eq:FDynamicsTest}) can be simplified via two methods, using a gradient at $X'$ or eigenfunctions of the kernel $g(\cdot, \cdot)$.

\paragraph{Simplification via Gradient}

Observe that for $G = A_t^T \cdot A_t$ to be time-invariant it is necessary for gradients $\{ \nabla_{\theta} 
f_{\theta_{t}}(X^i) \}_{i = 1}^{N}$ at training points either to be constant along the optimization or rotating together
via some time-variant rotation matrix $R_t$, $\nabla_{\theta} 
f_{\theta_{t}}(X^i) = R_t \cdot \nabla_{\theta} 
f_{\theta_{0}}(X^i)$ and $A_t = R_t \cdot A_0$. Such rotational behavior will lead to the required time-independence of $G = A_0^T \cdot R_t^T \cdot R_t \cdot A_0 = A_0^T \cdot A_0$. Similarly, for $g(X', \dtX)$ to be time-invariant the gradient $\nabla_{\theta} 
f_{\theta_{t}}(X')$ at the testing point must rotate with the same rotation $R_t$, $\nabla_{\theta} 
f_{\theta_{t}}(X') = R_t \cdot \nabla_{\theta} 
f_{\theta_{0}}(X')$.

Assuming the above gradient rotation, the row vector $g(X', \dtX)$ can be written as:
\begin{equation}
g(X', \dtX)
=
\nabla_{\theta} 
f_{\theta_{t}}(X')^T
\cdot
A_t
=
\nabla_{\theta} 
f_{\theta_{0}}(X')^T
\cdot
R_t^T
\cdot
R_t
\cdot
A_0
=
\nabla_{\theta} 
f_{\theta_{0}}(X')^T
\cdot
A_0
.
\label{eq:GSCross}
\end{equation}

Next, consider $A_0$'s SVD as triplets $\{\sqrt{\lambda_i}, \bar{\omega}_i, \bar{\upsilon}_{i} \}_{i = 1}^{N'}$ of ordered singular values, left and right singular vectors respectively, and denote $\nabla_{\theta} 
f_{\theta_{0}}(X') = \sum_{i = 1}^{N'}
a_i
\cdot
\sqrt{\lambda_i}
\cdot
\bar{\omega}_i
$ for $a_i \triangleq \frac{<\bar{\omega}_i, \nabla_{\theta} 
	f_{\theta_{0}}(X')>}{\sqrt{\lambda_i}}$. Using SVD properties of $A_0$, we get an identity $g(X', \dtX) = \sum_{i = 1}^{N'}
a_i
\cdot
\lambda_i
\cdot
\bar{\upsilon}_{i}^T$, and we can rewrite $f_{\theta_{t}}(X')$ from Eq.~(\ref{eq:FDynamicsTest}) as (note that $\bar{m}_{0}^z$ is reduced since it is orthogonal to $\{ \bar{\upsilon}_{i} : \lambda_i \neq 0 \}$):
\begin{multline}
f_{\theta_{t}}(X')
=
f_{\theta_{0}}(X')
-
\sum_{i = 1}^{N'}
\left[
1 - 
	\left[
	1 - 
	\frac{\delta}{N}
	\lambda_i
	\right]^t
\right]
a_i
<\bar{\upsilon}_{i}, \bar{m}_{0}>
=\\
=
f_{\theta_{0}}(X')
-
\sum_{i = 1}^{N'}
\left[
1 - 
\left[
1 - 
\frac{\delta}{N}
\lambda_i
\right]^t
\right]
\frac{1}{\sqrt{\lambda_i}}
<\bar{\upsilon}_{i}, \bar{m}_{0}>
<\bar{\omega}_i, \nabla_{\theta} 
f_{\theta_{0}}(X')>
.
\label{eq:FDynamicsTest2}
\end{multline}

Likewise, under the stability condition $\delta < \frac{2N}{\lambda_{max}}$, $f_{\theta_{t}}(X')$ at time $t = \infty$ can be expressed as:
\begin{equation}
f_{\theta_{\infty}}(X')
=
f_{\theta_{0}}(X')
-
\sum_{i = 1}^{N'}
\frac{1}{\sqrt{\lambda_i}}
<\bar{\upsilon}_{i}, \bar{m}_{0}>
<\bar{\omega}_i, \nabla_{\theta} 
f_{\theta_{0}}(X')>
.
\label{eq:FDynamicsTestInfinite}
\end{equation}

\paragraph{Simplification via Kernel Eigenfunctions}

According to Eq.~(\ref{eq:TestingSpectrum}), a product $g(X', \dtX) \cdot \bar{\upsilon}_{i}$ can be approximated by $\lambda_i \cdot \tilde{\upsilon}_{i}(X')$, with $\tilde{\upsilon}_{i}(\cdot)$ being an eigenfunction of $g(\cdot, \cdot)$. Using this approximation, Eq.~(\ref{eq:FDynamicsTest}) is reduced to:
\begin{multline}
f_{\theta_{t}}(X')
\approx
f_{\theta_{0}}(X')
-
\frac{\delta}{N}
\cdot
\Bigg[
\sum_{i = 1}^{N'}
\frac{1 - 
	\left[
	1 - 
	\frac{\delta}{N}
	\lambda_i
	\right]^t
}
{\frac{\delta}{N}
}
<\bar{\upsilon}_{i}, \bar{m}_{0}>
\tilde{\upsilon}_{i}(X')
+\\
+
t
\cdot
\sum_{i: \lambda_i = 0}
\lambda_i
<\bar{\upsilon}_{i}, \bar{m}_{0}>
\tilde{\upsilon}_{i}(X')
\Bigg]
=
f_{\theta_{0}}(X')
-
\sum_{i = 1}^{N'}
\left[
1 - 
	\left[
	1 - 
	\frac{\delta}{N}
	\lambda_i
	\right]^t
\right]
<\bar{\upsilon}_{i}, \bar{m}_{0}>
\tilde{\upsilon}_{i}(X')
,
\label{eq:FDynamicsTest3}
\end{multline}
which at time $t = \infty$ will converge to:
\begin{equation}
f_{\theta_{\infty}}(X')
=
f_{\theta_{0}}(X')
-
\sum_{i = 1}^{N'}
<\bar{\upsilon}_{i}, \bar{m}_{0}>
\tilde{\upsilon}_{i}(X')
.
\label{eq:FDynamicsTestInfinite3}
\end{equation}

\paragraph{Intuition}

Eq.~(\ref{eq:FDynamicsTest2}) and Eq.~(\ref{eq:FDynamicsTest3}) describe first-order dynamics of NN output at a testing point. The intuition behind these expressions can be summarized as following. First, for standard NN initialization
$f_{\theta_{0}}(X')$ is typically very close to be zero and can be neglected, leading to $\bar{m}_{0} \approx - \bar{y}$. Like in Eq.~(\ref{eq:FDynamics}), the inner-product term $<\bar{\upsilon}_{i}, \bar{m}_{0}>$, independent of testing point $X'$, defines which part of the signal contained in $\bar{m}_{0}$ is learned along each spectral direction. In general, $\left[
1 - 
\frac{\delta}{N}
\lambda_i
\right]^t$ converges faster for large eigenvalues. Also, due to large $\lambda_i$ being typically associated with $\bar{\upsilon}_{i}$ that contains a low-frequency signal, this leads to fast learning of low-frequency information and slow (sometimes infinitely slow) learning of high-frequency information. Further, the inner-product term $<\bar{\omega}_i, \nabla_{\theta} 
f_{\theta_{0}}(X')>$ in Eq.~(\ref{eq:FDynamicsTest2}) or the eigenfunction $\tilde{\upsilon}_{i}(X')$ in Eq.~(\ref{eq:FDynamicsTest3}), that are functions of $X'$, determine amount of information along $i$-th spectral direction that is transferred into $f_{\theta_{t}}(X')$, basically describing the generalization behind Eq.~(\ref{eq:FVecDff}) for a fixed Gramian $G$. Note that the convergence rate of $f_{\theta_{t}}(X')$ towards $f_{\theta_{\infty}}(X')$ is governed by how close terms $1 - 
\frac{\delta}{N}
\lambda_i
$ in Eq.~(\ref{eq:FDynamicsTest2}) and Eq.~(\ref{eq:FDynamicsTest3}) are to zero, similarly to the convergence rate of a system in Eq.~(\ref{eq:FDynamics}). Hence, we expect $f_{\theta_{t}}$ to converge to its final state at both training and testing points with a similar speed.

\section{Appendix: First-order Change of $G_t$}
\label{sec:AppD}

Here we describe the first-order Taylor approximation of a change in $G_t$ between sequential iterations of GD optimization. We theorize that the thorough analysis of below expressions will lead to the mathematical explanation required to understand evolution of $G_t$ as also to better understanding of NN dynamics.

First, change of the Jacobian $A_t$, defined in Section \ref{sec:FIMSec}, can be described as:
\begin{equation}
dA_{t} \triangleq A_{t + 1} - A_{t}
\approx
 - 
 \frac{\delta}{N}
 \cdot
W_t
,
\label{eq:JacobDff}
\end{equation}
where $W_t$ is $|\theta| \times N$ matrix with $i$-th column being $\mathcal{H}_{t}(X^i)
\cdot
A_{t}
\cdot
\bar{m}_{t}
$, with $\mathcal{H}_{t}(X) \triangleq \frac{\partial^2 f_{\theta_t}(X)}{\partial \theta^2}$ being the model Hessian.

Hence, the change between $G_{t + 1} = A_{t+1}^T \cdot A_{t+1}$ and $G_{t} = A_{t}^T \cdot A_{t}$ can be written as:
\begin{equation}
dG_{t} \triangleq G_{t + 1} - G_{t}
\approx
- 
\frac{\delta}{N}
\cdot
\left[
A_t^T \cdot W_t
+
W_t^T \cdot A_t
\right]
+
\frac{\delta^2}{N^2}
\cdot
W_t^T \cdot W_t
.
\label{eq:GramianDff}
\end{equation}
The last term can be neglected due to $\frac{\delta^2}{N^2}$ being significantly smaller than $\frac{\delta}{N}$, which leads to:
\begin{equation}
dG_{t} \triangleq G_{t + 1} - G_{t}
\approx
- 
\frac{\delta}{N}
\cdot
\left[
Q_t
+
Q_t^T
\right]
,
\label{eq:GramianDff2}
\end{equation}
where $Q_t$ is $N \times N$ matrix whose $i$-th column is $A_{t}^T \cdot 
\mathcal{H}_{t}(X^i)
\cdot
A_{t}
\cdot
\bar{m}_{t}
$.

Recently, similar expressions were reported by \cite{Dyer19arxiv} (specifically, see Eq.~(100-102)) and by \cite{Huang19arxiv}.

\section{Appendix: Computation Details of Fourier Transform}
\label{sec:AppF}

Here we provide more details on how Fourier Transform was calculated in our experiments. Consider a function $\varphi(X)$ and $N$ dataset points $\dtX = \{ X^k \in \RR^d \}_{k = 1}^{N}$ sampled from an arbitrary pdf $P(X)$. Further, consider a $N \times 1$ vector $\bar{\varphi}$ with entries $\bar{\varphi}(k) = \varphi(X^k)$. Given $\bar{\varphi}$, we compute Fourier Transform $\hat{\varphi}(\xi)$ of a function $\varphi(X)$ at $\xi \in \RR^d$ as following:
\begin{multline}
\hat{\varphi}(\xi) =
\int
\varphi(X)
\cdot
\exp
\left[
-2 \pi i
\cdot
< \xi, X >
\right]
\cdot
P(X)
dX
\approx
\\
\approx
\frac{1}{N}
\sum_{k = 1}^{N}
\varphi(X^k)
\cdot
\exp
\left[
-2 \pi i
\cdot
< \xi, X^k >
\right]
=
\frac{1}{N}
\bar{\varphi}^T
\bar{\varepsilon}
,
\label{eq:FTDetails}
\end{multline}
where $\bar{\varepsilon}$ is a $N \times 1$ vector with entries $\bar{\varepsilon}(k) = \exp
\left[
-2 \pi i
\cdot
< \xi, X^k >
\right]
$. Note that the above definition of Fourier Transform w.r.t. pdf $P(X)$ is identical to the common formulation without a term $P(X)$ inside, since in our experiments data distribution is $P(X) = 1$ (see "Setup" in Section \ref{sec:Expr}).

In all our experiments we compute $\hat{\varphi}(\xi)$ for $\xi$ taking values in $[- 40, 40]^2$. Further, we present a frequency component $| \hat{\varphi}(\xi) |$ as an image.

To perform the above computation, we require sampled values $\bar{\varphi}$ of the analyzed function $\varphi(X)$. In case this function is the eigenfunction of \emph{gradient similarity} kernel, the eigenvector of $G_t$ approximates this eigenfunction' values at the training points, as is shown in the Appendix \ref{sec:AppSp}. Hence, in this case the eigenvector of $G_t$ serves as a vector $\bar{\varphi}$ in Eq.~(\ref{eq:FTDetails}). Likewise, the above calculation using the residual vector $\bar{m}_{t}$ can be considered as a Fourier Transform of a function $r(X) \triangleq f_{\theta_t}(X) - y(X)$.

\end{document}